%% file: main.tex
\documentclass[letterpaper, 10 pt, conference]{ieeeconf}

\IEEEoverridecommandlockouts
\usepackage[T1]{fontenc}
\usepackage{microtype}
\usepackage{amsmath,amssymb}
\usepackage{graphicx}
\usepackage{booktabs}
\usepackage{multirow}
\usepackage{siunitx}
\usepackage{adjustbox}
\usepackage[dvipsnames]{xcolor}
\usepackage{colortbl}
\usepackage{caption}
\usepackage{subcaption}
\usepackage{pgfplots}
\pgfplotsset{compat=1.18}
\usepgfplotslibrary{groupplots}
\usepackage[ruled,linesnumbered]{algorithm2e}
\usepackage{placeins}
\usepackage{flafter}
\usepackage{float}
\usepackage{stfloats}
\usepackage{url}
\usepackage{cite}

\makeatletter
\let\NAT@parse\undefined
\makeatother

\usepackage[
  colorlinks=true,
  linkcolor=MidnightBlue,
  citecolor=MidnightBlue,
  urlcolor=RoyalBlue,
  bookmarksopen=true,
  bookmarksnumbered=true
]{hyperref}

\usepackage{cleveref}
\crefname{section}{Sec.}{Secs.}
\crefname{table}{Tab.}{Tabs.}
\crefname{figure}{Fig.}{Figs.}

\newcommand{\projectpage}{\href{https://scenes.aeva.com}{\texttt{scenes.aeva.com}}}
  
\definecolor{colInfra}{HTML}{5A9178}
\definecolor{colVehicle}{HTML}{5778A3}
\definecolor{colVRU}{HTML}{B56576}
\definecolor{colOther}{HTML}{9A8C78}

\makeatletter
\DeclareRobustCommand\onedot{\futurelet\@let@token\@onedot}
\def\@onedot{\ifx\@let@token.\else.\null\fi\xspace}

\makeatother

\renewcommand{\keywords}[1]{%
  \par\smallskip
  \begingroup
  \def\and{, }%
  \noindent\textit{Index Terms}---#1\par
  \endgroup
}

\title{\LARGE \bf AevaScenes: An FMCW LiDAR Dataset and Benchmark\\[-0.15em]
for Long-Range Perception\vspace{-0.7em}}
\author{Gautham Narayan Narasimhan$^{1}$,
        Heethesh Vhavle$^{1}$,
        Kumar Bhargav Viswanatha$^{1}$,\\
        James Reuther$^{1}$,
        and Deva Ramanan$^{2}$%
\thanks{$^{1}$Aeva Inc., Mountain View, CA, USA.
$^{2}$Carnegie Mellon University, Pittsburgh, PA, USA.}%
}

\IEEEaftertitletext{%
\vspace{-1.5em}%
\begin{minipage}{\textwidth}
\centering
\includegraphics[width=\textwidth]{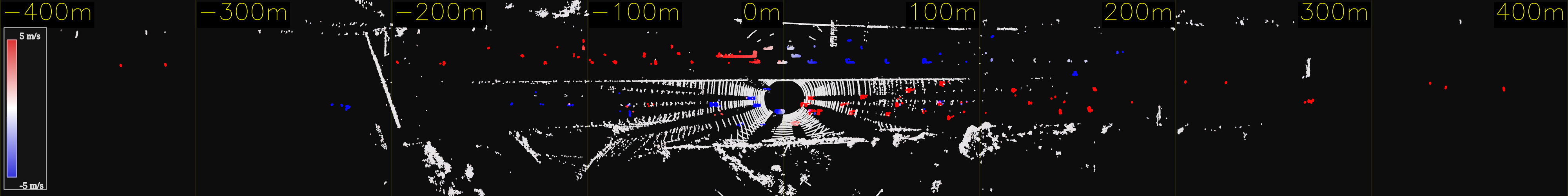}\\[0.15em]
\includegraphics[width=\textwidth]{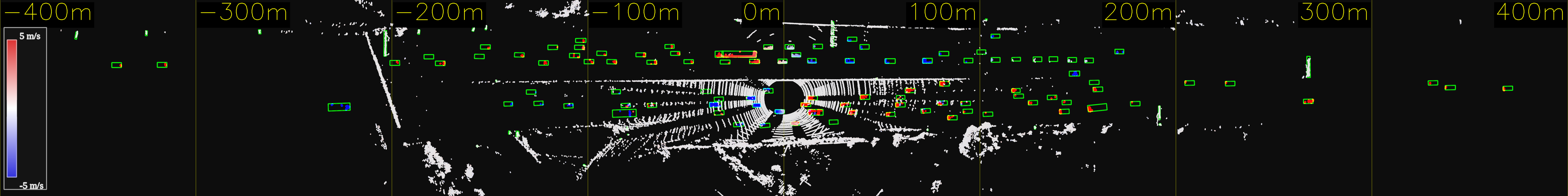}\\[-0.2em]
\includegraphics[width=\textwidth]{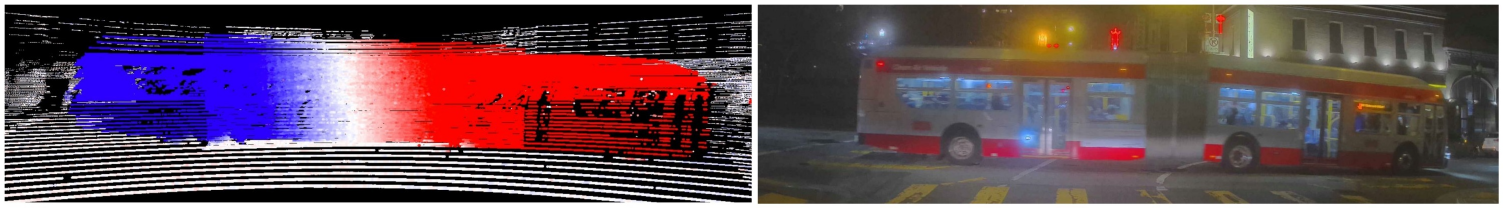}
\captionof{figure}{({\bf Top})
We color each point in a FMCW LiDAR frame by its radial velocity $d= \hat{\mathbf{r}}^T\mathbf{v}$ after ego-motion compensation, where $\hat{\mathbf{r}}$ is the unit line-of-sight (LOS) vector to that point and $\mathbf{v}$ is the point's true 3D velocity. In this scene, the ego-vehicle is moving to the right and a highway barrier separates the opposing direction of travel. Vehicles moving away from the ego-vehicle have positive \textcolor{red}{$d$} (red), while vehicles moving toward the ego-vehicle have negative \textcolor{blue}{$d$} (blue).
({\bf Middle}) The same scene with annotated 3D bounding boxes.
({\bf Bottom}) We show a bus moving to the right. Its front is moving away from the ego-vehicle and so has positive \textcolor{red}{$d$}, while its rear is moving toward the ego-vehicle, and so has negative \textcolor{blue}{$d$}. Importantly, points in the center have zero $d$  (white) because their direction of travel $\mathbf{v}$ is orthogonal to the LOS $\hat{\mathbf{r}}$. This "French flag" visual illustrates the {\em zero-Doppler blindspot}.}
\label{fig:teaser}
\end{minipage}\\[0.8em]
}

\begin{document}

\bstctlcite{IEEEetal}

\maketitle
\thispagestyle{empty}
\pagestyle{empty}

\input{sections/abstract}
\input{sections/introduction}
\input{sections/related_work}
\input{sections/dataset}
\input{sections/benchmark}
\input{sections/baseline_methods}
\input{sections/experiments}
\input{sections/conclusion}

\IEEEtriggeratref{28}
\bibliographystyle{IEEEtran}
\bibliography{main}

\clearpage
\input{sections/supplementary}

\end{document}

%% file: sections/abstract.tex
\begin{abstract}

FMCW LiDAR measures per-point radial Doppler velocity alongside range, providing a motion cue unavailable in conventional time-of-flight sensors. Exploiting this signal at long range remains understudied. We present an FMCW LiDAR dataset of 575 sequences (57.5K frames) with over 8 million annotated 3D boxes across 16 detection classes and per-point labels across 24 semantic classes, captured by six commercial FMCW LiDAR sensors and six paired 4K cameras across eight Bay Area cities, including 237 nighttime sequences, with annotations extending to 400\,m. We define a benchmark with three tasks: 3D object detection, scene flow estimation, and semantic segmentation. Detection and scene flow are evaluated across three range bins to 400\,m, with a public evaluation server. We explore the impact of Doppler measurements on flagship recognition tasks, and find significant improvements up to 2X in detection AP of far-away vehicles and pedestrians, particularly in low-latency single-frame settings. We similarly find scene flow accuracy is significantly improved with Doppler measurements across all ranges. Our dataset and benchmark have been publicly released at \projectpage.

\keywords{FMCW LiDAR \and Object Detection \and Scene Flow \and Semantic Segmentation \and Long Range Perception}
\end{abstract}

%% file: sections/introduction.tex
\section{Introduction}
\label{sec:introduction}

Safe autonomous driving at highway speeds demands reliable 3D perception well beyond current dataset ranges. Typical highway speeds of 60\,mph require obstacle detection at 400\,m to ensure enough reaction time ($\sim$15 seconds) for safe planning and deceleration. Yet widely-used LiDAR benchmarks, KITTI~\cite{geiger2012kitti}, nuScenes~\cite{caesar2020nuscenes}, Waymo Open~\cite{sun2020waymo}, and Argoverse~2~\cite{wilson2023argoverse2}, annotate only to 80--214\,m. Beyond $\sim$150\,m, per-object point counts drop to single digits, making it difficult to annotate, much less detect, objects at range. 
In theory, multi-frame aggregation could enable annotation/detection with fewer points, but this adds blur for moving objects, noise from ego-motion registration, and sensing latency: five frames (500\,ms) at 60\,mph cost $\sim$13\,m of travel.

Frequency-modulated continuous-wave (FMCW) LiDAR potentially addresses this gap by measuring radial point velocity at the sensor level. It measures both distance and radial velocity by transmitting a laser whose frequency changes over time (a chirp) and then correlating the returned light with a copy of the transmitted light. From the resulting interference signal, the sensor extracts two frequencies that correspond to {\bf range} (distance) and {\bf Doppler shift} (radial velocity). This is analogous to radar, but implemented with coherent laser light instead of radio frequencies.

\begin{figure}[t]
\centering
\captionsetup[sub]{skip=2pt}
\begin{subfigure}[t]{0.42\linewidth}
    \includegraphics[width=\linewidth,height=0.477\linewidth,trim=0 110 630 103,clip]{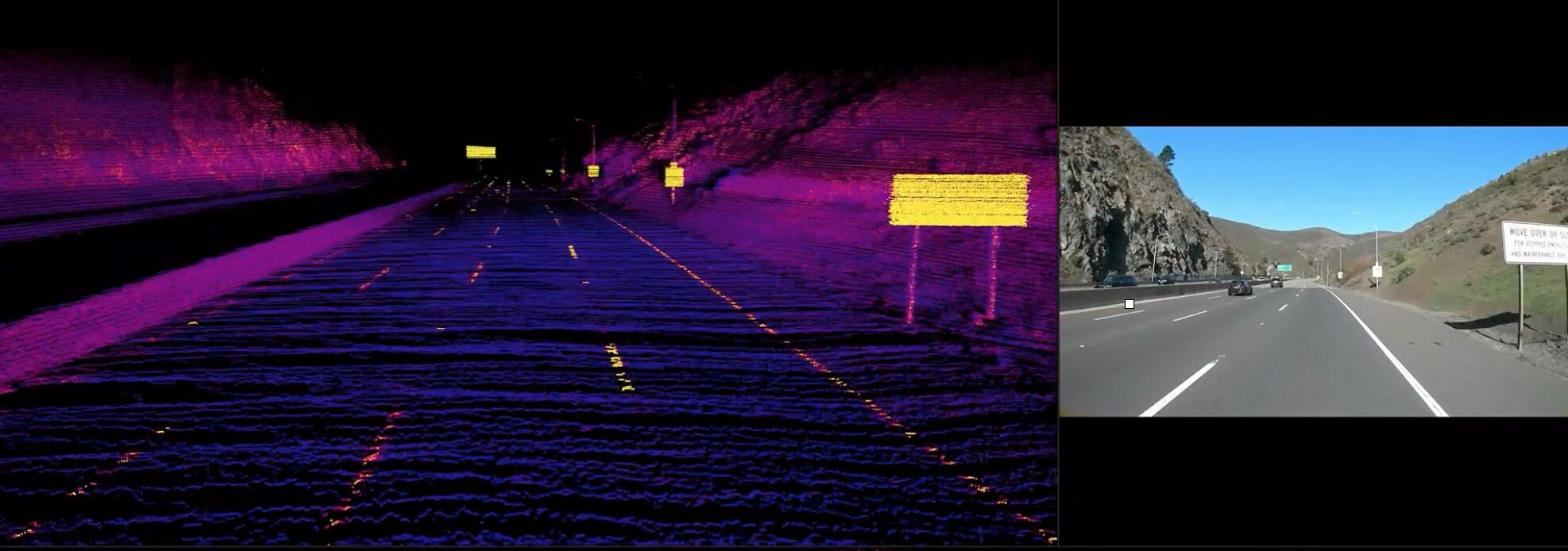}
    \caption{No blooming on retro-reflectors.}
\end{subfigure}\hfill
\begin{subfigure}[t]{0.57\linewidth}
    \includegraphics[width=\linewidth]{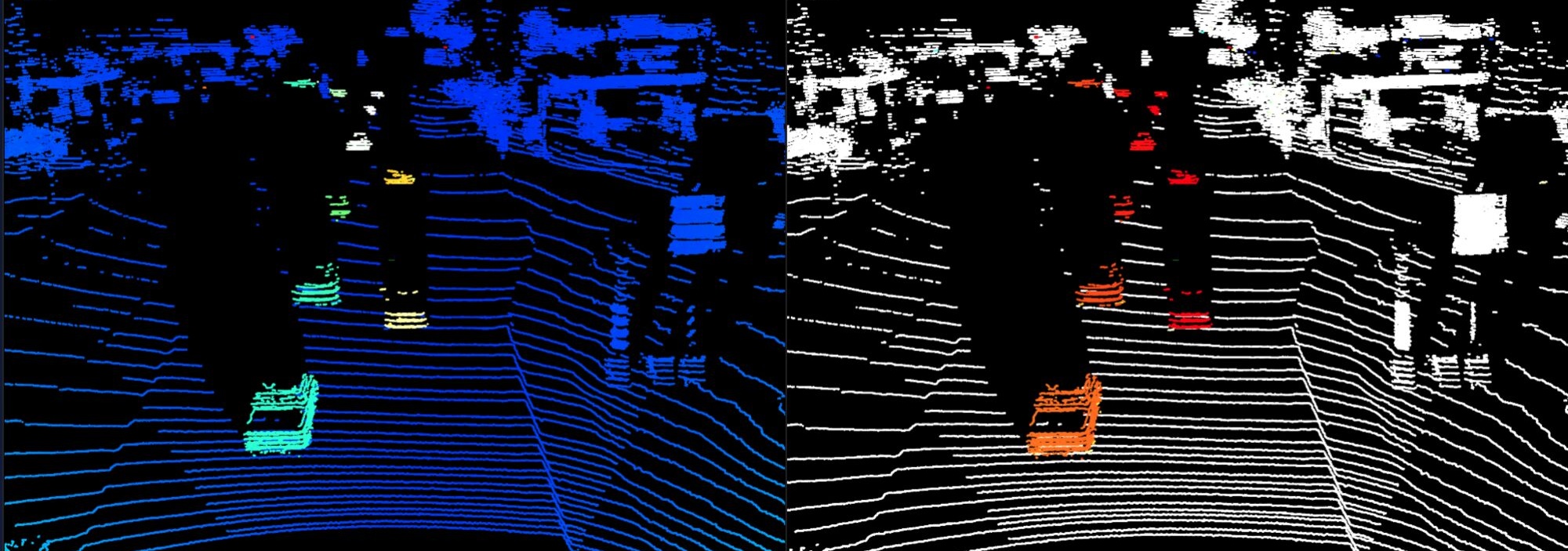}
    \caption{In-sensor ego-motion compensation.}
\end{subfigure}\\[3pt]
\begin{subfigure}[t]{\linewidth}
    \includegraphics[width=0.42\linewidth,height=0.2004\linewidth]{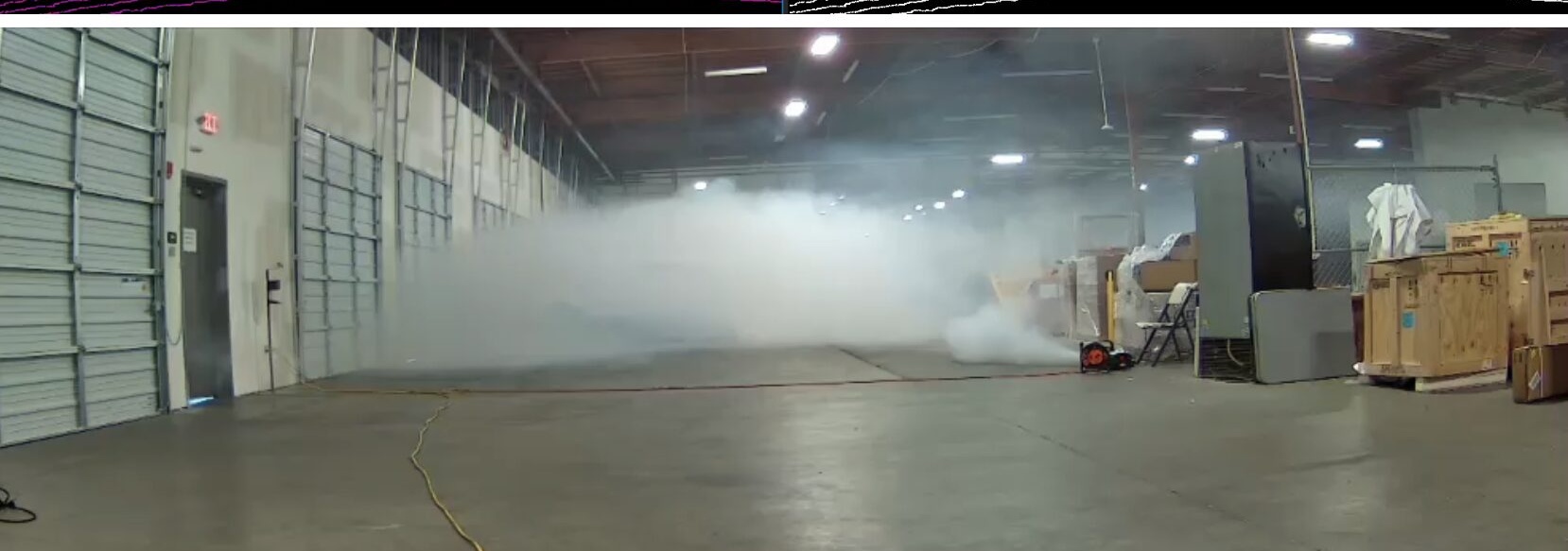}\hfill
    \includegraphics[width=0.57\linewidth]{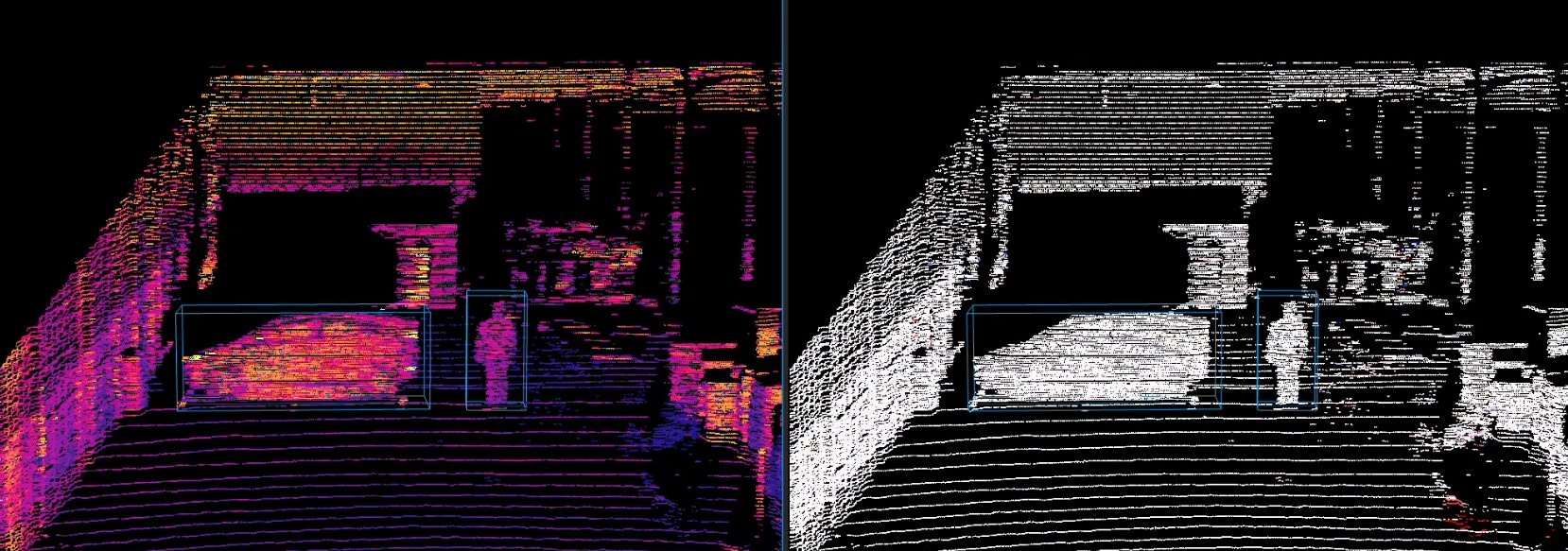}
    \caption{\textbf{Minimal degradation in fog.} (\textbf{Left:}) Foggy scene. (\textbf{Right:}) Minimal degradation in point-cloud quality.}
\end{subfigure}
\caption{Key advantages of FMCW LiDAR.}
\label{fig:fmcw_advantages}
\end{figure}

\begin{figure}[!t]
\centering
\captionsetup[sub]{skip=2pt}
\setcounter{subfigure}{0}
\begin{subfigure}[t]{0.323\linewidth}
    \includegraphics[width=\linewidth]{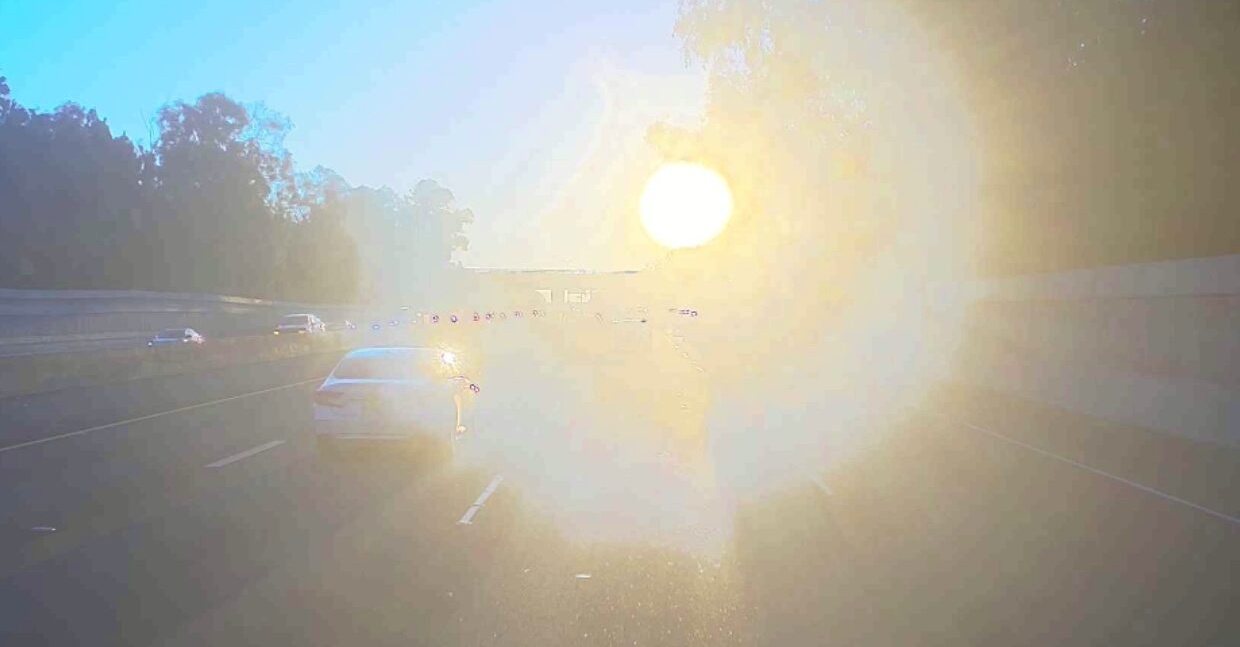}
    \caption{}
    \label{fig:sunlight_interference_sun_image}
\end{subfigure}\hfill
\begin{subfigure}[t]{0.323\linewidth}
    \includegraphics[width=\linewidth]{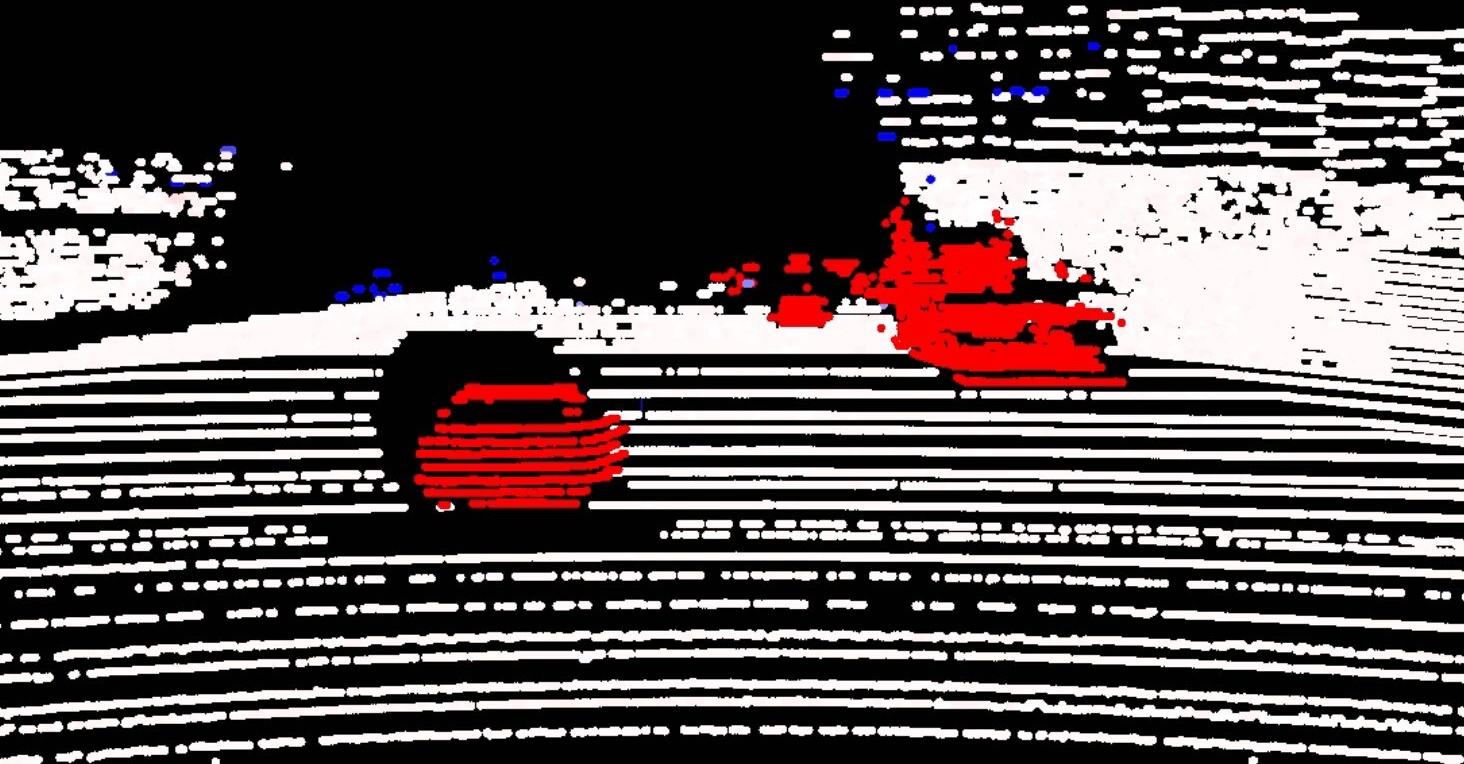}
    \caption{}
    \label{fig:sunlight_interference_pcd_image}
\end{subfigure}\hfill
\begin{subfigure}[t]{0.323\linewidth}
    \includegraphics[width=\linewidth]{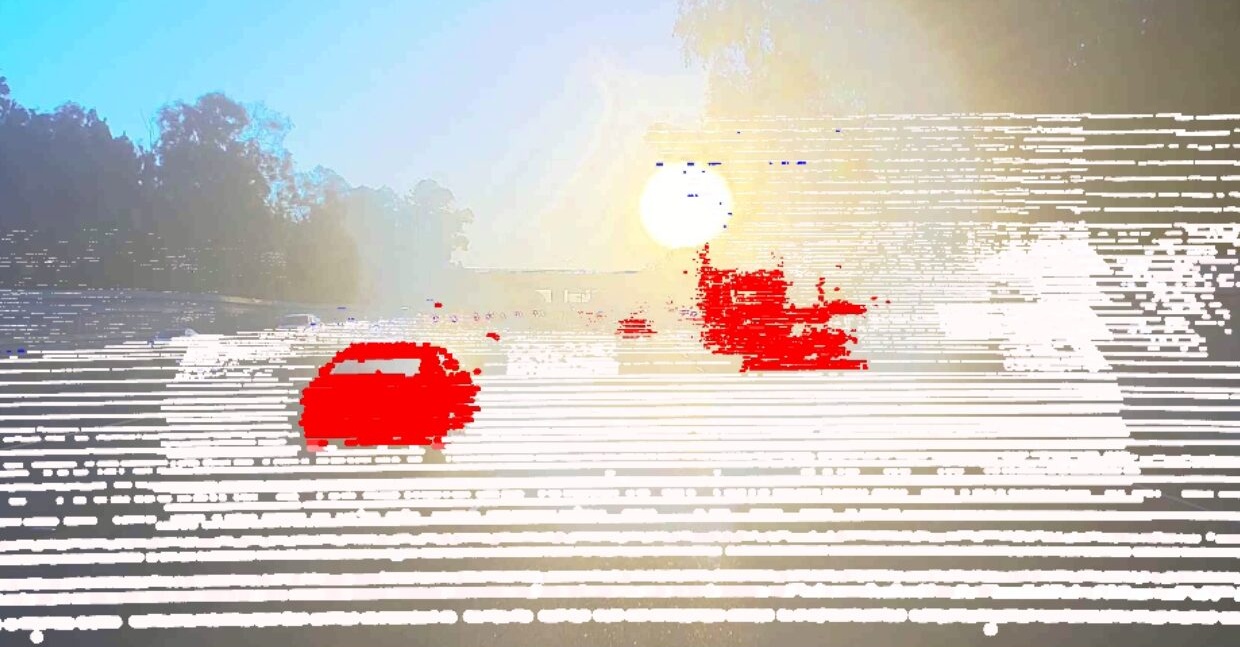}
    \caption{}
    \label{fig:sunlight_interference_superimposed}
\end{subfigure}\\[3pt]
\begin{subfigure}[t]{0.323\linewidth}
    \includegraphics[width=\linewidth]{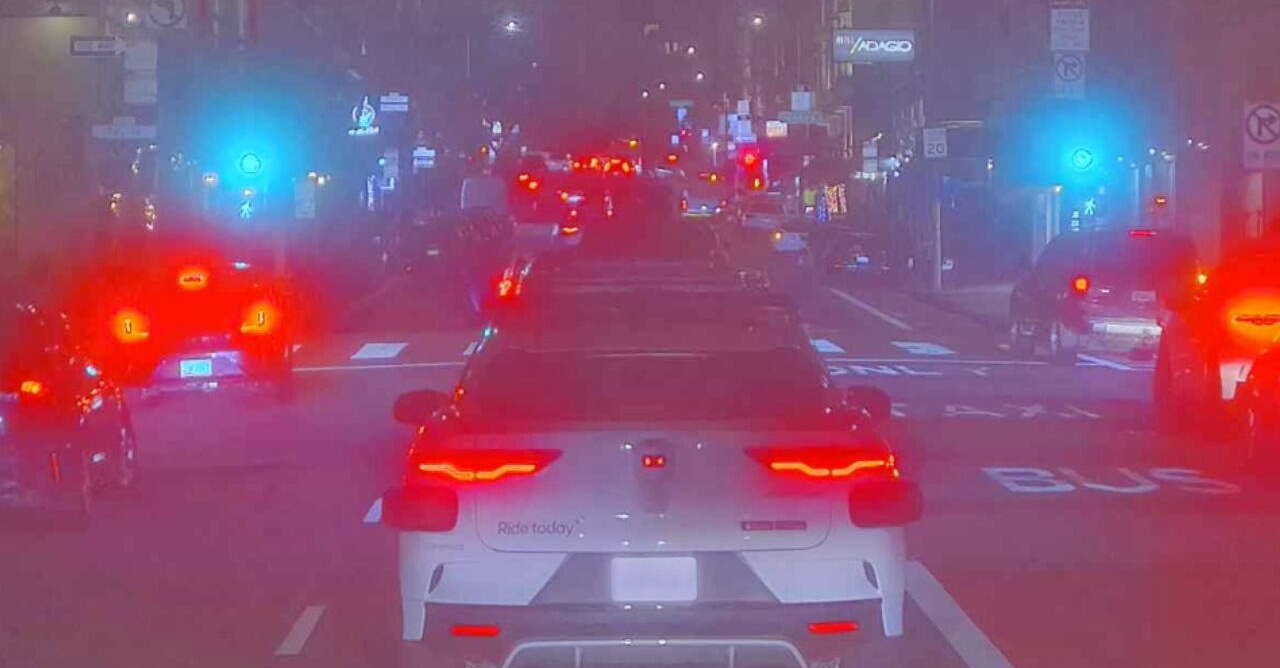}
    \caption{}
    \label{fig:f}
\end{subfigure}\hfill
\begin{subfigure}[t]{0.323\linewidth}
    \includegraphics[width=\linewidth]{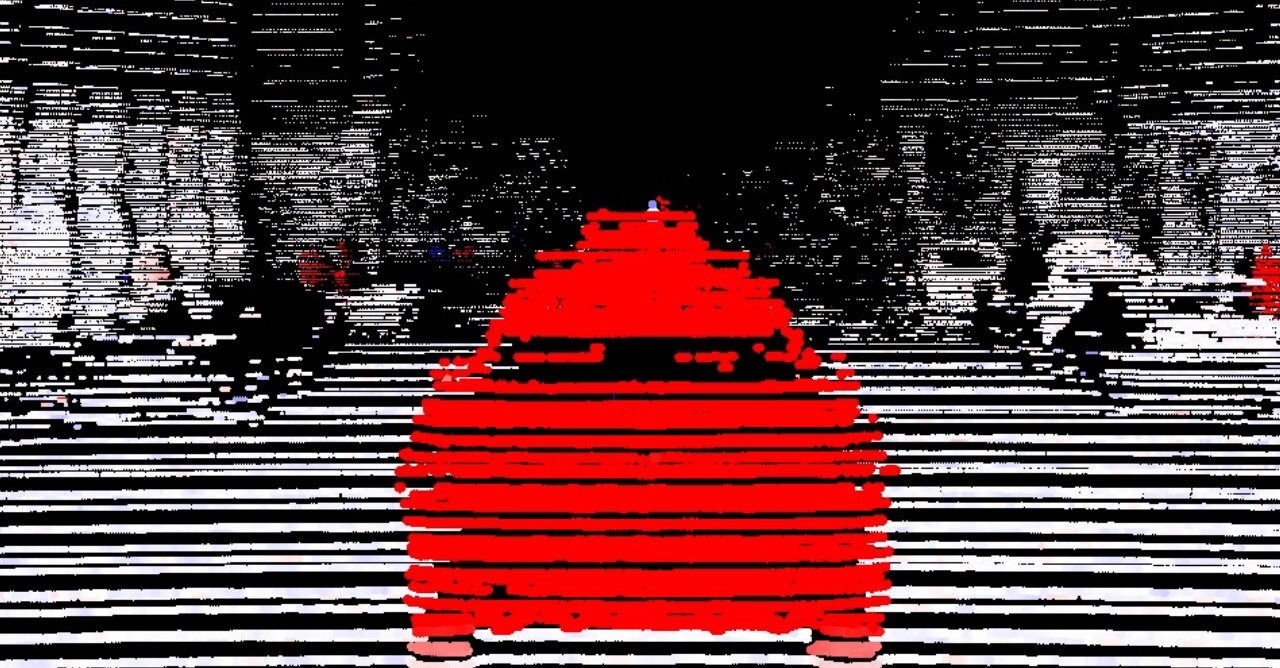}
    \caption{}
    \label{fig:d}
\end{subfigure}\hfill
\begin{subfigure}[t]{0.323\linewidth}
    \includegraphics[width=\linewidth]{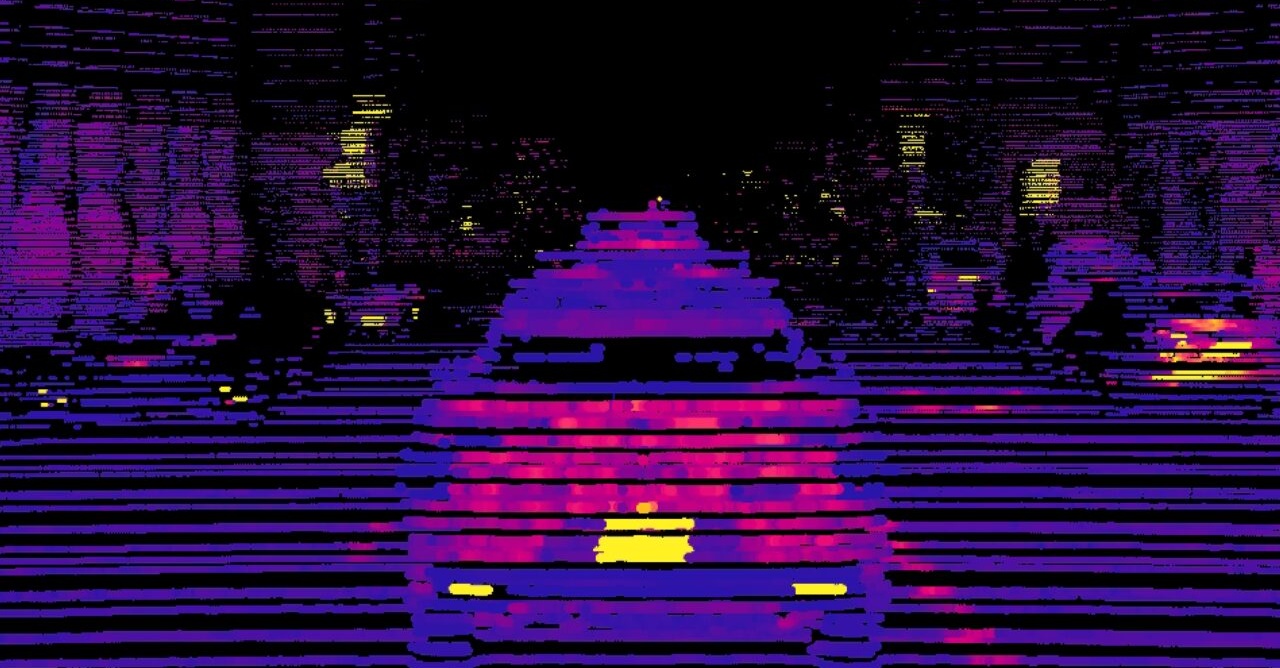}
    \caption{}
    \label{fig:e}
\end{subfigure}
\caption{\textbf{FMCW LiDAR under challenging sensing conditions.}
\textbf{Top: Direct sunlight.} ({\bf a}) Even under severe sun glare in the RGB image,
({\bf b}) FMCW LiDAR still produces accurate range and velocity measurements, which
({\bf c}) align well when overlaid.
\textbf{Bottom: Multi-LiDAR interference.} ({\bf d}) Here, the ego-vehicle is
surrounded by other active LiDAR sensors (the Waymo vehicle).
({\bf e}) FMCW LiDAR is immune to such interference and can still produce
accurate range, velocity, and ({\bf f}) reflectivity.}
\label{fig:fmcw_robustness}

\vspace{4pt}
\includegraphics[width=\columnwidth, height=0.16\textheight, keepaspectratio]{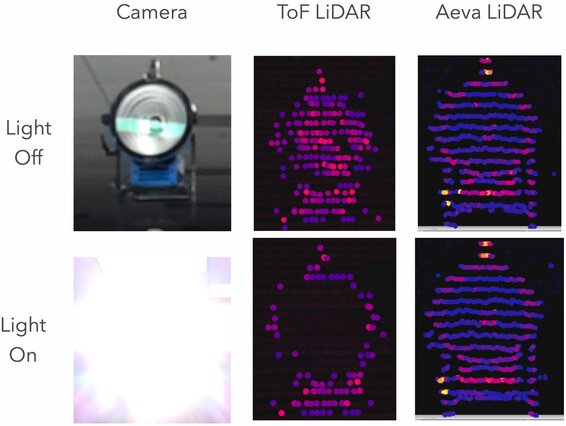}
\caption{Comparison of ToF vs FMCW LiDAR showing immunity to optical interference under direct illumination}
\label{fig:fmcw_vs_tof_light_interference}
\end{figure}

Compared to traditional (pulsed time-of-flight or ToF) LiDAR or Radar, {\em FMCW LiDAR directly measures per-point range and per-point radial velocity from a single sweep, without multi-frame aggregation or multi-sensor fusion} (\cref{fig:teaser}). Traditional radar offers poor spatial resolution, while traditional LiDAR infers point velocity only by differencing consecutive scans. This per-point velocity is especially valuable at long range, where geometric structure is sparse: with $\sim$15 points per object at 200--400\,m, velocity becomes the dominant cue for separating dynamic objects from static background. FMCW LiDAR also offers practical robustness advantages: because coherent processing correlates the returned signal with an internal copy, it is more robust to environmental obscurants such as fog and direct sunlight, interference from other LiDAR sensors, and blooming from retro-reflectors (Figure~\ref{fig:fmcw_advantages}, Figure~\ref{fig:fmcw_robustness}). 

\noindent {\bf FMCW LiDAR Datasets.} Public FMCW data is scarce. HeLiPR~\cite{jung2024helipr} includes FMCW data but lacks 3D annotations; its extension HeLiMOS~\cite{lee2024helimos} adds binary moving-object labels to a single sequence without bounding boxes or class labels.

Most related to us is TruckDrive~\cite{ghilotti2026truckdrive}, which provides an FMCW LiDAR dataset annotated with 3D cuboids to 400\,m, but targets heavy-truck highway driving (6.1\% urban, 9.6\% night) and does not explore the impact of Doppler cues for detection and velocity estimation.

\noindent {\bf Our Dataset.} We present a large-scale FMCW LiDAR dataset captured with six Aeva Aeries II FMCW LiDAR sensors and six paired 4K cameras across eight San Francisco Bay Area cities: 575 sequences and 57.5K synchronized frames at 10\,Hz ($\sim$508K points per frame). The dataset contains over 8 million annotated 3D boxes across 16 detection classes, with per-object tracking IDs, per-point labels across 24 semantic classes and annotations extending to 400\,m. It spans a more even urban/highway (55.8\%/44.2\%) and day/night (58.8\%/41.2\%) mix, with 237 nighttime sequences covering diverse traffic densities, road types, and lighting conditions; ego-motion poses are derived from fused GNSS, FMCW-based odometry, and offline registration, with paired cameras enabling future multimodal research.

\noindent {\bf Our Benchmark.} We define a benchmark with three tasks on a 10K-frame (100 sequences) test set: 3D object detection and scene flow estimation, both with a static/dynamic split and evaluated across three range bins (0--100, 100--200, 200--400\,m), and semantic segmentation over 24 classes. We evaluate four 3D detectors---CenterPoint~\cite{yin2021center}, DSVT~\cite{wang2023dsvt}, VoxelNeXt~\cite{chen2023voxelnext}, and TransFusion~\cite{bai2022transfusion}---with geometry-only and +Doppler velocity variants. For scene flow, we similarly evaluate DeFlow~\cite{zhang2024deflow}, Flow4D~\cite{kim2025flow4d}, and SSF~\cite{khoche2025ssf} with and without Doppler velocity. Preliminary Cylinder3D~\cite{zhu2021cylinder3d} and PTv3~\cite{wu2024ptv3} segmentation experiments show that naive velocity concatenation does not consistently improve dynamic-class IoU, motivating future motion-aware architectures.

\noindent {\bf Conclusions.}
We focus on the impact of adding Doppler velocity measurements on far-range and low-latency (single-frame) perception. Classic detectors like CenterPoint {\bf double} in accuracy from 33 AP to 67 AP for dynamic cars at range (200+ meters). Contemporary models like TransFusion improve from 50 to 72 AP. The effect is even more pronounced for challenging classes like pedestrians; TransFusion improves from 0 to 42 AP. Adding 5-frames of context improves overall performance (by adding 500\,ms of latency), but Doppler still more than doubles all-motion-state pedestrian AP from 13 AP to 33 AP. Doppler also consistently reduces scene-flow error, while preliminary semantic-segmentation baselines do not yet capture consistent gains from velocity on dynamic objects.

%% file: sections/related_work.tex
\section{Related Work}
\label{sec:related_work}

\begin{table*}[t]
    \centering
    \caption{Comparison with existing autonomous driving datasets. Our FMCW LiDAR dataset is the first large-scale dataset captured entirely with FMCW LiDAR, offering the highest point density (508K pts/fr), greatest annotation density (140 boxes/fr), longest effective range (400\,m), and the second largest number of night sequences (237) among all compared datasets.}
    \label{tab:dataset_comparison}
    \resizebox{\textwidth}{!}{%
    \begin{tabular}{@{}llrrrrrcccrc c@{}}
        \toprule
        & & \multicolumn{5}{c}{\textbf{Scale}} & \multicolumn{3}{c}{\textbf{Sensor Setup}} & \multicolumn{3}{c}{\textbf{Coverage}} \\
        \cmidrule(lr){3-7} \cmidrule(lr){8-10} \cmidrule(lr){11-13}
        \textbf{Year} & \textbf{Dataset} & \textbf{Scenes} & \textbf{Frames} & \textbf{Boxes/Fr} & \textbf{Pts/Fr} & \textbf{Objects} & \textbf{LiDARs} & \textbf{Type} & \textbf{Cameras} & \textbf{Range} & \textbf{Night (\#)} & \textbf{Classes} \\
        \midrule
        2012 & KITTI            & 22   & 15K  & 13   & 120K & 0.2M  & 1 & ToF        & 4   & 91\,m  & $\times$ & 8  \\
        2019 & nuScenes         & 1000 & 40K  & 35   & 34K  & 1.4M  & 1 & ToF        & 6   & 70\,m  & $\checkmark$ & 23 \\
        2020 & Waymo Open       & 1150 & 230K & 55   & 177K & 12.6M & 5 & ToF        & 5   & 80\,m  & 102 & 4  \\
        2023 & Argoverse V2     & 1000 & 150K & 33   & 107K & 5M    & 2 & ToF        & 9   & 214\,m & $\times$ & 30 \\
        2023 & HeLiPR           & 6    & --   & --   & --   & --    & 4 & ToF+FMCW   & --  & 200\,m & $\checkmark$ & -- \\
        2024 & HeLiMOS          & 1    & 12K  & --   & --   & --    & 4 & ToF+FMCW   & 0   & 200\,m & $\times$ & 3  \\
        2024 & MAN TruckScenes  & 747  & 30K  & 33   & --   & 1M    & 6 & ToF        & 4   & 230\,m & $\checkmark$ & 27 \\
        2026 & Boreas-RT        & 60   & --   & --   & --   & --    & 2 & ToF+FMCW   & 1   & 400\,m & -- & -- \\
        2026 & TruckDrive       & 3828 & 165K & --   & --   & --    & 7+3 & FMCW+ToF & 11--15 & 400\,m & 367 & 9  \\
        \midrule
        2026 & \textbf{Ours}    & \textbf{575} & \textbf{57.5K} & \textbf{140} & \textbf{508K} & \textbf{$>$8M} & \textbf{6} & \textbf{FMCW} & \textbf{6} & \textbf{400\,m} & \textbf{237} & \textbf{16} \\
        \bottomrule
    \end{tabular}%
    }
    \vspace{1ex}
    {\scriptsize $\checkmark$ indicates night data is available but sequence count is not reported; $\times$ indicates no night data; -- denotes not reported.}
\end{table*}

\noindent\textbf{LiDAR Datasets for Autonomous Driving.}
Large-scale annotated datasets have been central to advancing 3D perception for autonomous driving.
KITTI~\cite{geiger2012kitti} was the first large-scale benchmark to pair camera and LiDAR data with 3D bounding box annotations, catalyzing a decade of research in 3D object detection.
nuScenes~\cite{caesar2020nuscenes} significantly expanded scope by providing full 360$^\circ$ coverage with six cameras, five radars, and one LiDAR across 1{,}000 scenes in Boston and Singapore.
The Waymo Open Dataset~\cite{sun2020waymo} further scaled annotation density, offering over 12 million 3D boxes across diverse US geographies with high-quality 64-beam LiDAR.
Argoverse~2~\cite{wilson2023argoverse2} introduced the largest object taxonomy to date with 30 classes across six US cities, along with rich HD maps containing 3D lane geometry.
Beyond fully supervised settings, ONCE~\cite{mao2021once} provided one million LiDAR scenes to enable research in semi-supervised and self-supervised 3D detection.

While these datasets have driven enormous progress, they all employ time-of-flight (ToF) LiDAR sensors that measure only range and reflectance requiring either multi-frame temporal reasoning or auxiliary sensor fusion to infer per-point motion.

\noindent\textbf{4D Radar Datasets.}
Automotive 4D radar natively measures radial velocity alongside position. Astyx HiRes2019~\cite{meyer2019astyx} was the first public 4D radar dataset (500 frames, front-view only); View-of-Delft (VoD)~\cite{palffy2022vod} expanded to 8{,}693 frames focusing on vulnerable road users; TJ4DRadSet~\cite{zheng2022tj4dradset} added diverse lighting and road conditions; K-Radar~\cite{paek2022kradar} introduced 35{,}000 frames of full 4D tensors (range, azimuth, elevation, Doppler) with 3D annotations under adverse weather; and MAN TruckScenes~\cite{man2024truckscenes} provided the first 360$^\circ$ 4D radar coverage for heavy vehicles. Despite valuable Doppler measurements, 4D radar angular resolution remains fundamentally limited and produces two to three orders of magnitude fewer points per frame than LiDAR thereby restricting fine-grained detection of small or distant objects.

\textbf{FMCW LiDAR Sensing and Datasets.} 
Prior work utilizing FMCW LiDAR for perception has been limited in scope, scale, and public availability. Doppler velocity improves tracking~\cite{gu2022fmcw_tracking}, detection~\cite{dopplerptnet2024}, and predictive localization~\cite{shi2025pod}, but all rely on proprietary data, instance-only labels, or simulation without real sensor noise. 
HeLiPR~\cite{jung2024helipr} targets place recognition without 3D annotations; its extension HeLiMOS~\cite{lee2024helimos} adds binary moving-object labels to one sequence but lacks bounding boxes, class labels, or tracking IDs. MUSES~\cite{sakaridis2024muses} pairs an FMCW scanning \emph{radar} (not LiDAR) with a MEMS LiDAR, offering only 2D panoptic annotations without per-point velocity at LiDAR-level density.

A prior version of this dataset was released with a smaller set of 10K annotated frames~\cite{aevascenes2025}, limiting its use for training. 4DLidarOpen~\cite{qian2026_4dlidaropen} benchmarks Doppler-aware detection and BEV flow, but its 28K-frame, five-class human split uses one forward-facing FMCW sensor, reaches only up to 150\,m, and limits evaluation to one detector. TruckDrive~\cite{ghilotti2026truckdrive} provides an FMCW dataset annotated with 3D cuboids to 400\,m, but targets heavy-truck highway driving (6.1\% urban, 9.6\% night) and does not explore the impact of Doppler cues for detection and velocity estimation.

%% file: sections/dataset.tex
\section{AevaScenes Dataset}

\begin{figure}[!ht]
\centering
\includegraphics[width=0.86\columnwidth]{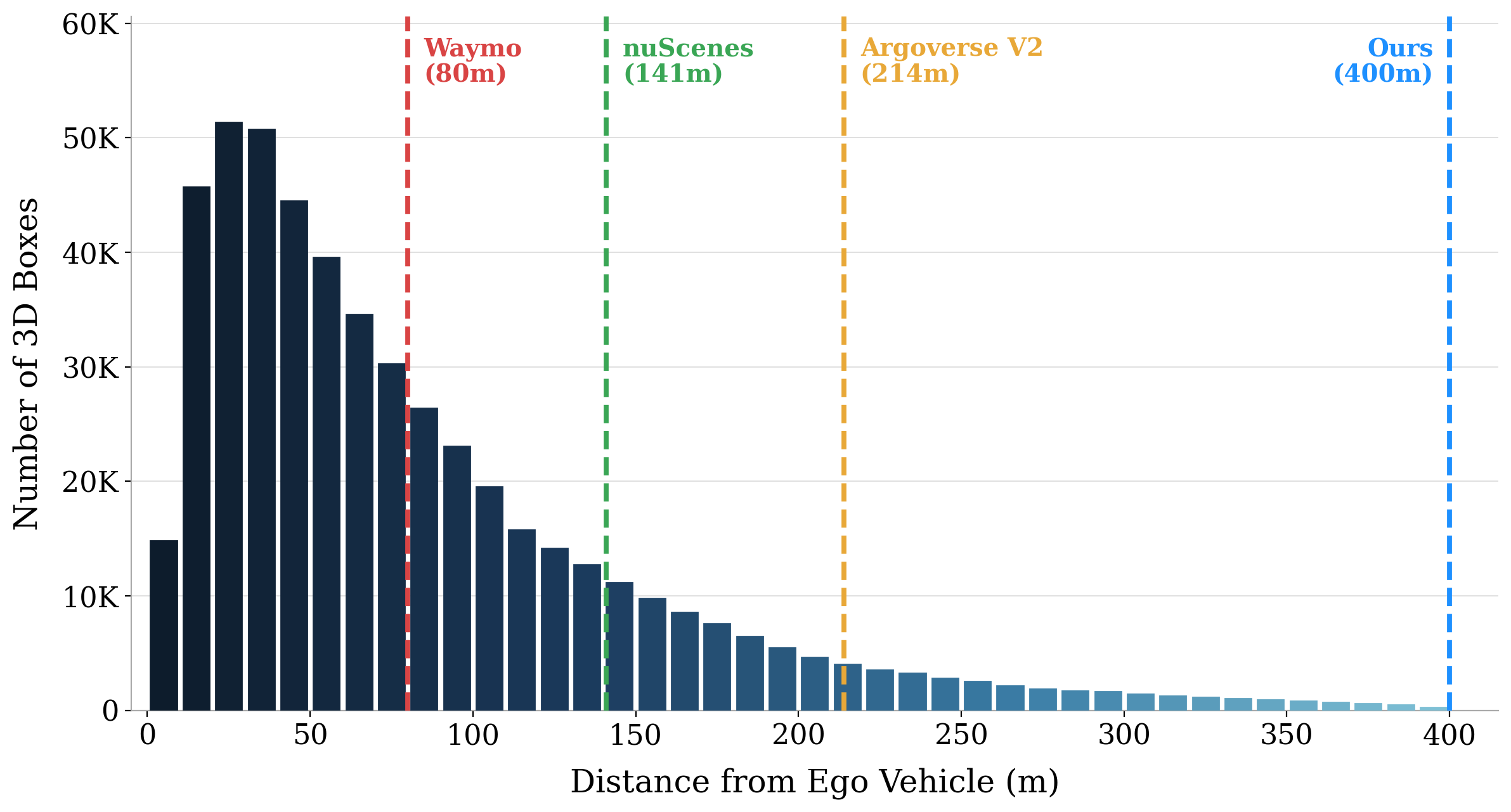}\\[1pt]
\includegraphics[width=0.86\columnwidth]{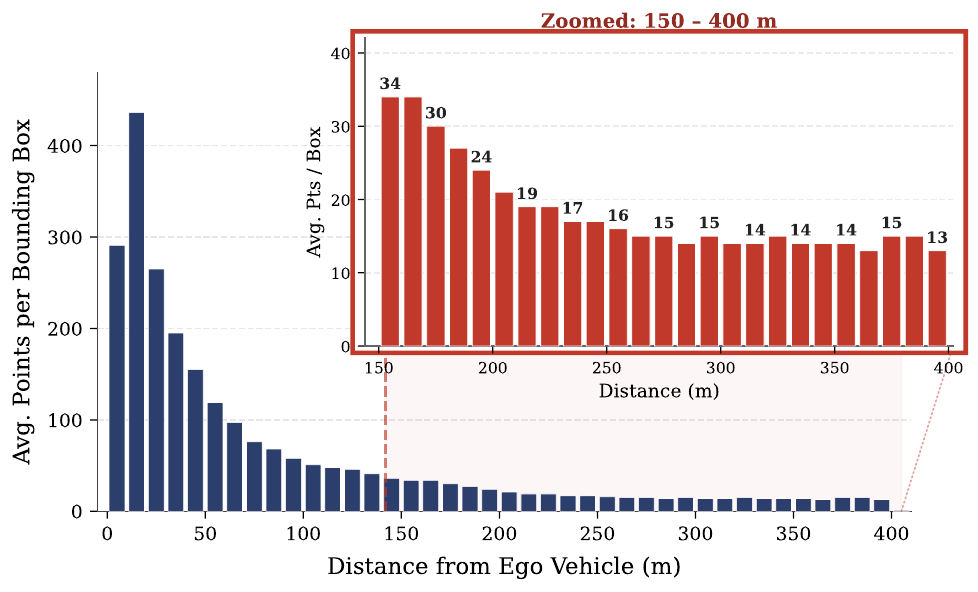}
\caption{\textbf{Long-range annotation statistics of our FMCW LiDAR dataset.} (\textbf{Top:}) distribution of 3D bounding boxes vs.\ distance from the ego vehicle; annotations extend up to 400\,m (Waymo 80\,m, nuScenes 141\,m, Argoverse~V2 214\,m). (\textbf{Bottom:}) average LiDAR points per box vs.\ distance, remaining $\sim$13 points per box to 400\,m.}
\label{fig:box_stats}
\end{figure}

\input{sections/miscellaneous/class_distribution}

{\setlength{\intextsep}{0pt}
\begin{figure}[!b]
    \centering
    \captionsetup{skip=1pt}
    \captionsetup[sub]{skip=1pt}
    \begin{subfigure}{\columnwidth}
        \centering
        \includegraphics[width=\linewidth]{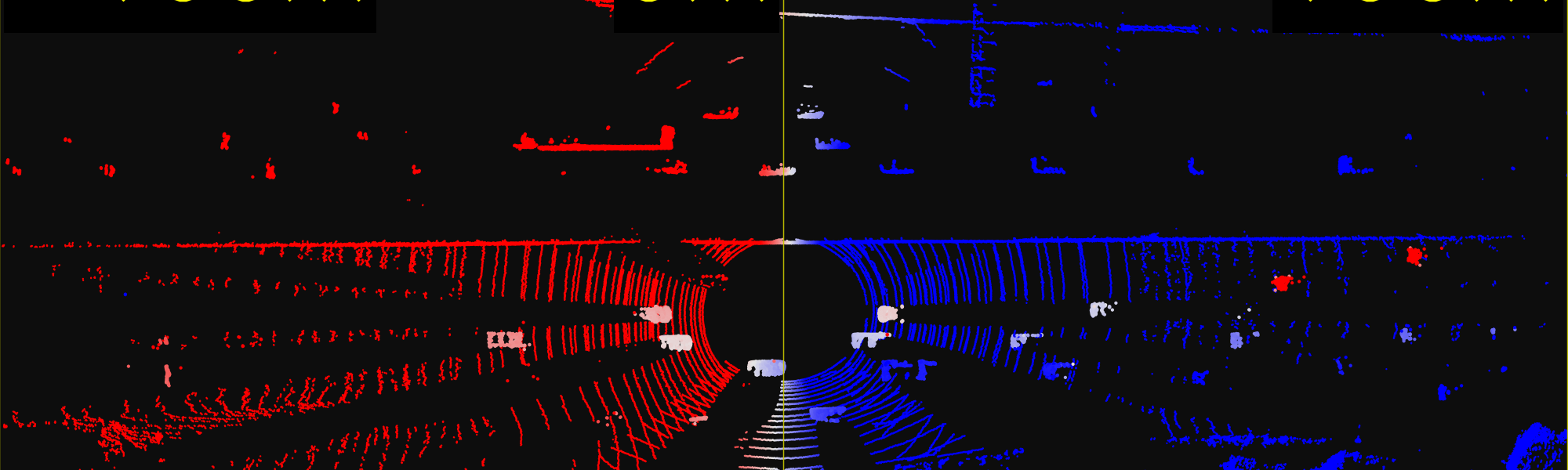}
        \caption{Raw Doppler $d'$.}
        \label{fig:egomotion_raw}
    \end{subfigure}\\[0pt]
    \begin{subfigure}{\columnwidth}
        \centering
        \includegraphics[width=\linewidth]{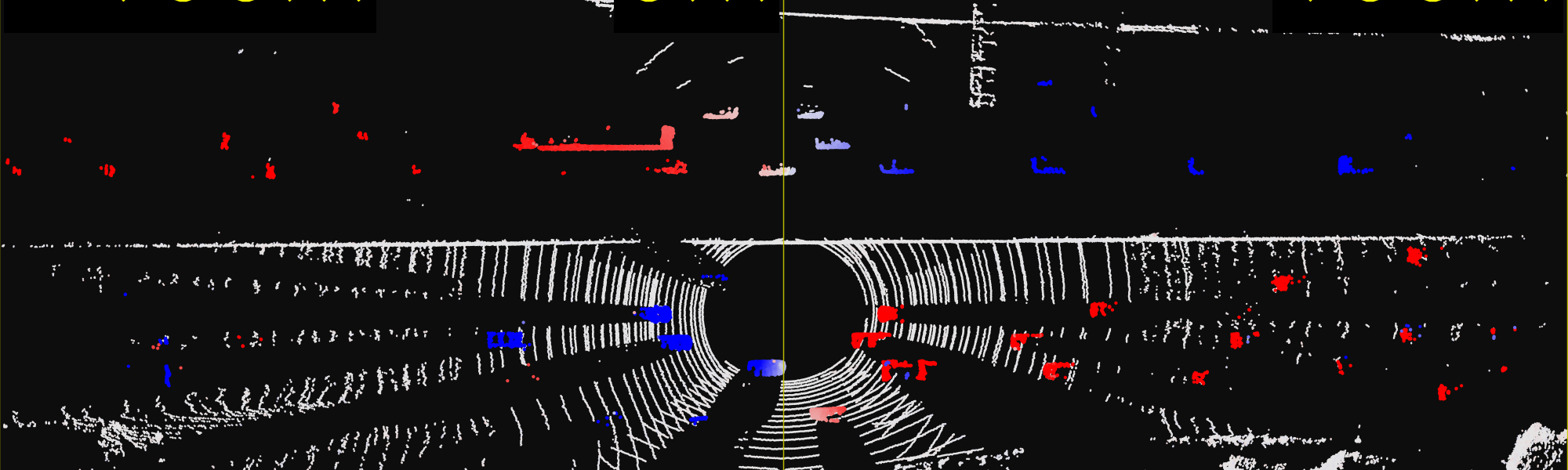}
        \caption{Ego-motion compensated Doppler $d$.}
        \label{fig:egomotion_comp}
    \end{subfigure}
    \caption{\textbf{Ego-motion compensation.}
    (\textbf{Top:}) raw radial velocity measured by the LiDAR sensors moving on a highway towards the right.
    (\textbf{Bottom:}) after ego-motion compensation, static surfaces (e.g. ground) have zero velocity.}
    \label{fig:egomotion}
\end{figure}
}

\subsection{FMCW LiDAR}

FMCW LiDAR demonstrates robust sensing performance across challenging environmental and operational conditions.
As shown in Fig.~\ref{fig:fmcw_robustness} (top row), range and per-point velocity measurements are preserved under direct sunlight. In Fig.~\ref{fig:fmcw_robustness} (bottom row), a multi-LiDAR traffic scene containing another autonomous vehicle remains fully observable with no degradation in range or velocity measurements. As illustrated in Fig.~\ref{fig:fmcw_advantages}, returns from retro-reflective infrastructure such as traffic signs exhibit no blooming artifacts, and under foggy conditions scene structure and motion cues remain observable despite significant visual degradation. Collectively, these properties enable consistent data collection across diverse lighting conditions, dense traffic, varying material properties, and adverse weather.

Per-point radial velocity directly benefits ego-motion estimation. In GNSS-denied and geometrically degenerate environments such as tunnels, highways, and long corridors, the Doppler channel constrains translational motion even when scene geometry is ambiguous, while dense spatial structure anchors the rotational estimate, significantly reducing drift~\cite{hexsel2022dicp, wu2022pickingupspeed, zhao2024fmcwlio}.
Automotive 4D radars also provide Doppler velocity, but their sparse, noisy point clouds and poor angular resolution~\cite{cao2025caoronet} limit odometry accuracy compared to the dense geometry available from FMCW LiDAR.
Per-point velocity further enables dynamic object rejection without an external tracker~\cite{wang2025dynamicicp}.
Combined with accurate ego-motion, this allows temporal accumulation of point clouds free of dynamic trails, improving downstream tasks like small object detection, road hazard identification, and lane boundary estimation.

\begin{figure}[!t]
\centering
\includegraphics[width=0.72\columnwidth, trim=0cm 2.5cm 0cm 2cm, clip]{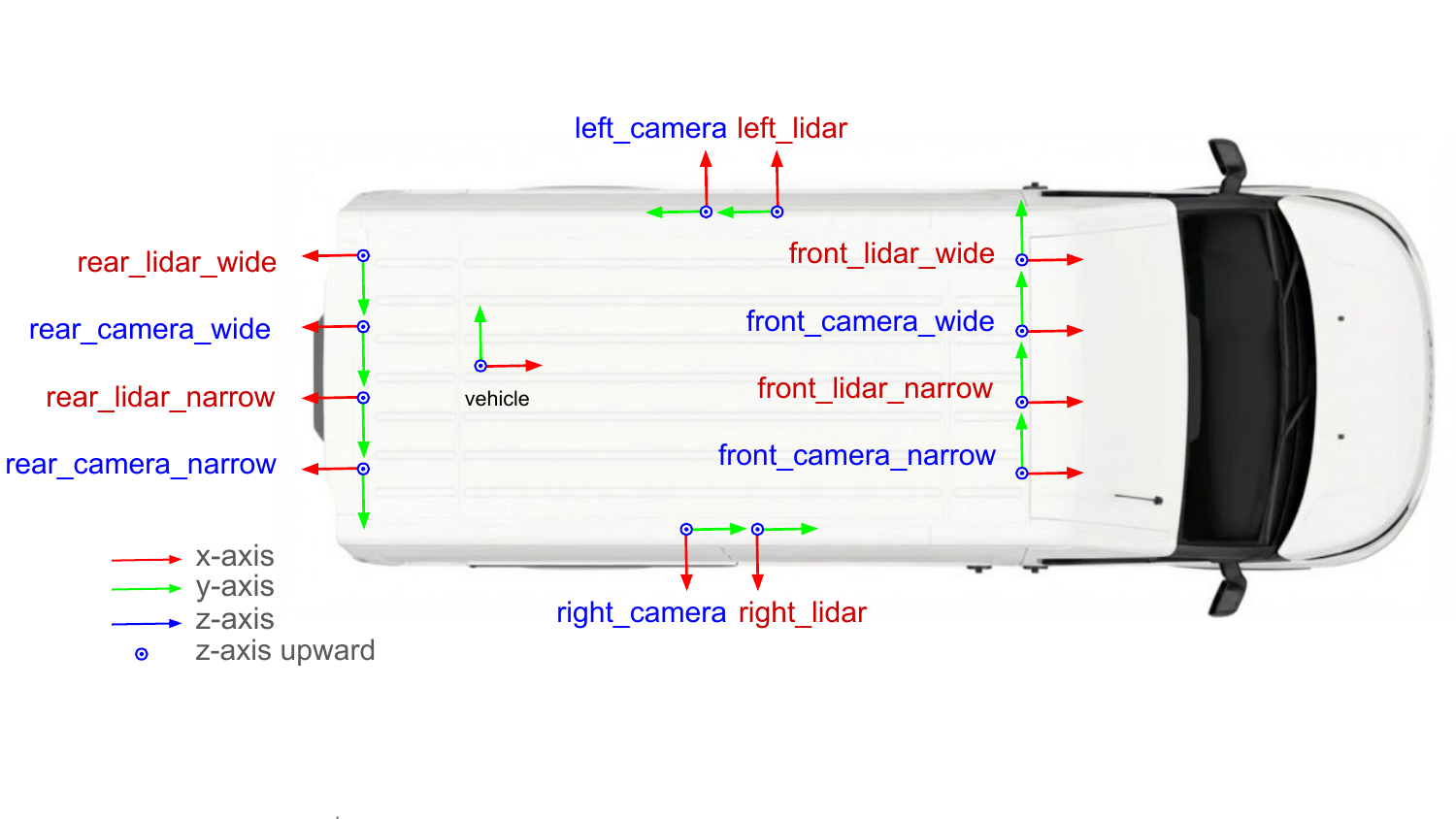}
\caption{Sensor setup and coordinate system}
\label{fig:vehicle_setup}

\vspace{0pt}
{\captionsetup{type=table,aboveskip=1pt,belowskip=1pt}
 \captionof{table}{Sensor specifications of the data collection vehicle.}
 \label{tab:sensor_config}}
{\footnotesize
\begin{tabular}{@{}lccccc@{}}
\toprule
\textbf{LiDAR} & \textbf{Count} & \textbf{Range} & \textbf{HFoV} & \textbf{VFoV} & \textbf{FPS} \\
\midrule
Aeries~II Wide   & $\times$4 & 250\,m & 110\textdegree & 19.4\textdegree & 10 \\
Aeries~II Narrow & $\times$2 & 400\,m & 35\textdegree  & 19.4\textdegree & 10 \\
\midrule
\textbf{Camera} & \textbf{Count} & \textbf{Resolution} & \textbf{HFoV} & \textbf{VFoV} & \textbf{FPS} \\
\midrule
OX08B40 Wide   & $\times$4 & 3840$\times$2160 & 120\textdegree & 67.5\textdegree & 30 \\
OX08B40 Narrow & $\times$2 & 3840$\times$2160 & 60\textdegree  & 36\textdegree   & 30 \\
\bottomrule
\end{tabular}}
\end{figure}

\noindent\textbf{Vehicle and Sensor Setup}
Our data collection vehicle is equipped with six Aeva Aeries II FMCW LiDAR sensors and six RGB cameras (\cref{fig:vehicle_setup,tab:sensor_config}). LiDARs follow a mixed wide/narrow FoV layout: front and rear positions each mount a wide, narrow pair for perception up to \SI{400}{\meter}, while the two side-facing wide-FoV units cover up to \SI{100}{\meter}. All LiDARs run at \SI{10}{\hertz}; each point encodes 3D position, reflectivity, instantaneous radial velocity, line index, and acquisition timestamp. The camera suite pairs four wide-FoV and two narrow-FoV 4K units at \SI{30}{\hertz}, with each camera spatially aligned to its corresponding LiDAR as shown in \cref{tab:sensor_config}. All sensors are calibrated to a common ego-vehicle frame centered at the rear axle and synchronized using PTP (IEEE 1588-2008), with LiDAR frame geometry compensated at the temporal mid-frame.

\begingroup
\setlength{\abovedisplayskip}{4pt}
\setlength{\belowdisplayskip}{4pt}
\setlength{\abovedisplayshortskip}{4pt}
\setlength{\belowdisplayshortskip}{4pt}
\noindent\textbf{Doppler velocity and ego-motion compensation.}
Alongside the range vector $\mathbf{p}_i-\mathbf{o}_i$ from the per-point sensor origin $\mathbf{o}_i$ to the return $\mathbf{p}_i$, the sensor measures a Doppler velocity $d_i$: the radial velocity of that point relative to the sensor along the unit line-of-sight (LOS) direction of the uncompensated geometry (\cref{fig:egomotion}),
\begin{equation}
  \hat{\mathbf{r}}_i = (\mathbf{p}_i - \mathbf{o}_i) / \|\mathbf{p}_i - \mathbf{o}_i\|.
  \label{eq:los}
\end{equation}
With sensor velocity $\mathbf{v}_s\in\mathbb{R}^3$ and point velocity $\mathbf{v}_i\in\mathbb{R}^3$ (both in m/s, in the sensor frame; $\mathbf{v}_i=\mathbf{0}$ for static points, the object velocity otherwise), $d_i$ is the LOS projection of the relative velocity, and adding back the projected sensor velocity yields the ego-motion-compensated Doppler $d_i'$, i.e., the radial velocity a stationary sensor would measure:
\begin{align}
  d_i &=\hat{\mathbf{r}}_i^\top(\mathbf{v}_i-\mathbf{v}_s)
       =\hat{\mathbf{r}}_i^\top\mathbf{v}_i-\hat{\mathbf{r}}_i^\top\mathbf{v}_s,
  \label{eq:doppler-raw}\\
  d_i'&=d_i+\hat{\mathbf{r}}_i^\top\mathbf{v}_s
       =\hat{\mathbf{r}}_i^\top\mathbf{v}_i.
  \label{eq:egomotion-doppler}
\end{align}
\endgroup

\noindent\textbf{Dataset Overview}
The dataset comprises 575 sequences of 10\,s at 10\,Hz (100 frames each, 57{,}500 total), with $\sim$508K LiDAR points per frame and over 8 million annotated 3D boxes across 16 detection classes. Collection spans urban and highway environments across eight San Francisco Bay Area cities (Mountain View, Sunnyvale, Palo Alto, San Jose, San Francisco, Fremont, Union City, and San Mateo), covering diverse road types (residential, arterial, highway), traffic densities, and lighting conditions (55.8\% urban / 44.2\% highway; 58.8\% day / 41.2\% night), including 237 dedicated nighttime sequences. We provide both the raw and ego-motion compensated (\cref{eq:egomotion-doppler}) point clouds (see supplementary \cref{sec:dataset-supplement,fig:bev_all}).

\noindent\textbf{Data Curation and Annotation}.
We curated the 57.5K frame dataset from several hundred hours of internally collected driving logs across various geographies. Front-camera images were embedded using CLIP and SigLIP~2~\cite{clip,siglip2}, and their t-SNE projections were inspected to assess scene diversity and identify underrepresented conditions. Text-to-image retrieval further targeted challenging scenes, including severe sun glare and nearby autonomous vehicles equipped with LiDAR (\cref{fig:fmcw_robustness}). We also used a 3.5B-parameter Qwen-VL model to caption scene activity, followed by a secondary model that surfaced unusual behaviors such as pedestrian jaywalking and vehicles cutting closely in front of the ego vehicle. Because offline perception was insufficiently reliable beyond 200\,m, where objects are sparse, frequently occluded, and can be tracked only for short durations, human annotators labeled the selected sequences from scratch. Every annotated frame was then processed by an in-house object detector to flag model--human disagreements for correction, followed by a final frame-by-frame QC process until we reach a 99\% accuracy.

\noindent\textbf{Long Range Annotations}.
A defining characteristic of our dataset is the extent and density of its long-range
supervision. Each frame contains $\sim$140 annotated objects; $\sim$40\% lie beyond
80\,m and $\sim$18\% beyond 141\,m, with the distribution tailing continuously to
400\,m, providing dense long-range labeled ground truth, including scene-flow labels
(Fig.~\ref{fig:box_stats}, Table~\ref{tab:dataset_comparison}). Point density degrades
gradually with range, with per-object point count plateauing at 13--15 points from
200--400\,m. Each long-range point carries instantaneous radial velocity, a signal absent from ToF sensors, which becomes
the dominant discriminative cue for dynamic/static separation when spatial structure
alone is ambiguous. 

Annotations comprise 3D bounding boxes across 16 categories, per-object tracking IDs, scene-flow labels, and per-point semantic labels (see supplementary
\cref{sec:semantic-segmentation-supplement,tab:semseg-v2}). Detection categories aggregate into four groups,
\textit{infrastructure}, \textit{vehicles}, \textit{Vulnerable Road Users (VRUs)}, and \textit{other}, revealing
a long-tailed distribution (Fig.~\ref{fig:class_distribution}). At long range, radial
velocity is essential for annotators to disambiguate moving objects from static
background where frame-to-frame geometric differencing is insufficient. Together with 237 nighttime sequences, these velocity-aware, long-range annotations support early detection, motion forecasting, and long-range perception, encouraging range-aware strategies in which Doppler velocity is increasingly valuable as geometric density diminishes.

%% file: sections/miscellaneous/class_distribution.tex
\begin{figure}[t]
\centering
\begin{tikzpicture}
\begin{axis}[
    ybar, bar width=8pt, bar shift=0pt,
    width=\columnwidth, height=4.2cm,
    ymode=log, log origin=infty,
    ymin=2500, ymax=5000000,
    ylabel={Number of instances}, ylabel style={font=\small},
    ytick={1e3, 1e4, 1e5, 1e6},
    yticklabels={$10^3$, $10^4$, $10^5$, $10^6$},
    yticklabel style={font=\footnotesize},
    xtick={0,1,2,3,4,5,6,7,8,9,10,11,12,13,14,15},
    xticklabels={
        {pole\_trunk}, {car}, {traffic\_sign}, {pedestrian}, {traffic\_item},
        {other\_structure}, {truck}, {trailer}, {bicycle},
        {other\_vehicle}, {bus}, {bicyclist}, {motorcycle}, {motorcyclist},
        {vehicle\_on\_rails}, {animal}
    },
    xticklabel style={font=\scriptsize, rotate=45, anchor=east},
    xmin=-0.8, xmax=15.8,
    ymajorgrids=true, grid style={gray!20, dashed},
    legend style={
        at={(0.98,0.98)}, anchor=north east, font=\footnotesize,
        draw=gray!30, fill=white, fill opacity=0.9, text opacity=1,
        legend columns=2, column sep=4pt},
]
\addplot[fill=colInfra, draw=colInfra!60!black, fill opacity=0.85] coordinates {(0, 2426366)};
\addlegendentry{Infrastructure}
\addplot[forget plot, fill=colInfra, draw=colInfra!60!black, fill opacity=0.85] coordinates {(2, 1209328)};
\addplot[forget plot, fill=colInfra, draw=colInfra!60!black, fill opacity=0.85] coordinates {(4, 339148)};
\addplot[forget plot, fill=colInfra, draw=colInfra!60!black, fill opacity=0.85] coordinates {(5, 137842)};
\addplot[fill=colVehicle, draw=colVehicle!60!black, fill opacity=0.85] coordinates {(1, 2014082)};
\addlegendentry{Vehicle}
\addplot[forget plot, fill=colVehicle, draw=colVehicle!60!black, fill opacity=0.85] coordinates {(6, 109194)};
\addplot[forget plot, fill=colVehicle, draw=colVehicle!60!black, fill opacity=0.85] coordinates {(7, 38917)};
\addplot[forget plot, fill=colVehicle, draw=colVehicle!60!black, fill opacity=0.85] coordinates {(8, 31612)};
\addplot[forget plot, fill=colVehicle, draw=colVehicle!60!black, fill opacity=0.85] coordinates {(9, 28575)};
\addplot[forget plot, fill=colVehicle, draw=colVehicle!60!black, fill opacity=0.85] coordinates {(10, 23581)};
\addplot[forget plot, fill=colVehicle, draw=colVehicle!60!black, fill opacity=0.85] coordinates {(12, 10984)};
\addplot[forget plot, fill=colVehicle, draw=colVehicle!60!black, fill opacity=0.85] coordinates {(14, 6680)};
\addplot[fill=colVRU, draw=colVRU!60!black, fill opacity=0.85] coordinates {(3, 660407)};
\addlegendentry{Vulnerable Road User}
\addplot[forget plot, fill=colVRU, draw=colVRU!60!black, fill opacity=0.85] coordinates {(11, 13343)};
\addplot[forget plot, fill=colVRU, draw=colVRU!60!black, fill opacity=0.85] coordinates {(13, 7642)};
\addplot[fill=colOther, draw=colOther!60!black, fill opacity=0.85] coordinates {(15, 3093)};
\addlegendentry{Other}
\end{axis}
\end{tikzpicture}
\caption{Class distribution of our annotated FMCW LiDAR dataset.}
\label{fig:class_distribution}
\end{figure}

%% file: sections/benchmark.tex
\subsection{Benchmark Tasks and Metrics}
\label{sec:benchmark}

Our benchmark is designed to evaluate 3D perception at ranges up to 400\,m and to encourage methods that exploit the per-point velocity signal unique to FMCW LiDAR. We define three tasks: 3D object detection, 3D scene flow estimation and semantic segmentation and present baseline results for all three. The goal of this benchmark is to encourage methods that exploit the per-point velocity signal unique to FMCW LiDAR, especially for single-frame detection---where Doppler matches five-frame geometry-only accuracy at 5$\times$ lower sensing latency (500\,ms vs 100\,ms for newly visible objects) and long-range detection.

\noindent\textbf{3D Object Detection.}
We evaluate detection across 16 classes grouped into \textit{vehicles} (car, bus, truck, trailer, vehicle\_on\_rails, other\_vehicle), \textit{vulnerable road users} (pedestrian, bicyclist, bicycle, motorcyclist, motorcycle, animal), and \textit{infrastructure} (pole\_trunk, traffic\_sign, traffic\_item, other\_structure).

Predictions are matched to ground truth by using center distance threshold of 2\,m for vehicles and 1\,m for pedestrians. The primary metric is Average Precision (AP) with 101-point interpolation. Ground-truth filtering is applied on the raw point cloud with a uniform minimum of 5 points per box, ensuring fairness independent of any model-specific preprocessing. Metrics are reported across three range bins (0--100, 100--200, 200--400\,m) and over the full 400\,m range on a test split of 10K frames (see supplementary \cref{sec:detection-results-supplement,tab:learned-implementation} for the dynamic/static partition and speed-error metric).

\noindent\textbf{Scene Flow Estimation.}
Scene flow labels are derived from ground-truth human-annotated object tracks.
Unlike conventional LiDAR, FMCW sensing directly measures the radial component of each point's velocity via Doppler, so the estimation problem reduces to recovering the unobserved tangential component; methods can treat the per-point radial measurement as a strong prior, both for gating static points (compensated Doppler ${\approx}0$) and for constraining the flow of dynamic ones.
The primary metric is absolute end-point error (EPE), reported on all, static, and dynamic points using the same three range bins as detection.
We do not prescribe a fixed frame pairing, allowing methods to use consecutive or multi-frame inputs.

\noindent\textbf{Semantic Segmentation.}
We define point-wise segmentation across 24 classes (excluding
\textit{unknown} from evaluation), spanning vehicles, vulnerable road users,
infrastructure, ground surfaces (road, sidewalk, lane\_boundary,
road\_marking, reflective\_marker, other\_ground), and vegetation. The primary
metric is mIoU across all evaluated classes, with per-class IoU reported.
Evaluation is restricted to points with ground-truth labels. We compare
metrics with and without compensated radial velocity under otherwise identical
settings; the full evaluation protocol is provided in supplementary
Table~\ref{tab:learned-implementation}(e).

%% file: sections/baseline_methods.tex
\section{Baseline Methods}

\subsection{Classical Dynamic Object Detection and Tracking}
\label{sec:classical-baseline}

We introduce a classical, learning-free Kalman filter baseline that exploits per-point Doppler measurements for detection and tracking. We provide detailed pseudocode and parameters in supplementary \cref{sec:classical-baseline-supplement,alg:pipeline,tab:supp-params,tab:supp-tiers}, but include a summary here. 

\noindent\textbf{Dynamic object detection.}
After ego-motion compensation, each return measures
$d_i'=\hat{\mathbf{r}}_i^\top\mathbf{v}_{\mathrm{obj}}$ from
\cref{eq:egomotion-doppler}. Static objects have
$d_i'\approx0$, while dynamic objects retain a non-zero radial component;
thus, $|d_i'|>0.7$\,m/s isolates dynamic \emph{seed} points without
multi-frame differencing or scan registration. The zero-Doppler blind
spot (\cref{fig:teaser}) makes motion perpendicular to all LOS directions
indistinguishable from static background, affecting perpendicular
cross-traffic, adjacent-lane vehicles observed by side-facing sensors,
and turning vehicles passing through the perpendicular-to-LOS direction.
DBSCAN~\cite{ester1996dbscan} clusters seeds in BEV, followed by
Doppler-based splitting and re-merging to separate overlapping objects
moving at different speeds.

\noindent\textbf{Cluster velocity estimation.}
For each cluster, we estimate ground-plane velocity
$\mathbf{v}=[v_x,v_y]^\top$ using
\begin{equation}
  d_i'=(\hat{\mathbf{r}}_i)_{xy}^\top\mathbf{v},
  \label{eq:doppler-model}
\end{equation}
where $(\cdot)_{xy}$ retains the ground-plane components of the LOS
direction in \cref{eq:los}. Weights
$w_i=\max(|d_i'|,0.5)$ emphasize returns with a large Doppler velocity
for robustness.
If narrow azimuth spread or an ill-conditioned observation matrix leaves
tangential velocity underdetermined, a radial-only fallback projects the
median Doppler velocity along the mean LOS, preserving speed while
leaving heading to the tracker.

\noindent\textbf{Tracking and Limitations.}
A UKF~\cite{julier1997ukf} with a CTRV model tracks
$\mathbf{x}=[x,y,z,s,\psi,\dot{\psi}]^T$ (position, speed, heading, yaw
rate) from $\mathbf{z}=[x,y,z,s,\psi]^T$; cluster velocity directly
initializes and updates $s$ and $\psi$, whereas ToF pipelines need
several noisy positions before finite differencing yields usable
velocity, leaving early tracks heading-ambiguous. For radial-only
detections, inflated yaw noise lets the motion model carry heading
through Doppler-blind phases. Beyond the
zero-Doppler blind spot, pedestrian recall is near zero since returns
are sparse and often below the Doppler threshold; learned methods could
exploit temporal context~\cite{chen2021lidarmos}, shape priors, and
per-point Doppler supervision for scene flow.

\subsection{Learning-Based Detection}
\label{sec:learning-baseline}

Learned detectors address the shortcomings of the classical pipeline
(Sec.~\ref{sec:classical-baseline}) by leveraging shape priors,
temporal context, and class-specific supervision.
CenterPoint~\cite{yin2021center} provides a convolutional BEV baseline,
predicting
an anchor-free heatmap of object centers and regressing a 3D box from
each peak. DSVT~\cite{wang2023dsvt} instead applies dynamic sparse
window attention to voxel tokens, alternating rotated set partitions
to exchange information across local groups before BEV detection.
VoxelNeXt~\cite{chen2023voxelnext} is fully sparse: a sparse 3D CNN
predicts boxes directly from occupied voxels without conversion to a
dense BEV feature map. Because computation follows occupied voxels
rather than the area of a dense grid, it scales efficiently to
long-range scenes. Finally, we use the
LiDAR-only variant of
TransFusion~\cite{bai2022transfusion}, which initializes class-aware
object queries from center-heatmap peaks and refines them by
cross-attending to BEV features. Together, these strong baselines span
convolutional and attention-based backbones, dense BEV and sparse voxel
representations, and center-, voxel-, and query-based prediction. This
provides architectural diversity while holding the sensing modality,
supervision, and input variants fixed, allowing Doppler gains to be
compared across detector families.

\noindent\textbf{Geometry-only.}
XYZ coordinates and reflectivity, with no velocity. We train both a
single-frame (100\,ms) model and a five-frame (500\,ms) model that
accumulates the current frame plus four ego-motion-compensated past
frames. This mirrors the standard ToF LiDAR detection setup applied to
FMCW data, at the cost of 5$\times$ sensing latency ($\sim$13\,m at
60\,mph).

\noindent\textbf{+vel: Doppler velocity as additional input.}
We append ego-motion-compensated per-point radial velocity as an extra
point channel. All other training settings are unchanged.

\subsection{Scene Flow}
\label{sec:scene-flow-baseline}
We evaluate three recent supervised LiDAR scene-flow methods designed
for full-scale driving point clouds. DeFlow~\cite{zhang2024deflow}
encodes two scans as dense PointPillars BEV maps with a 2D U-Net, then
uses an iterative GRU decoder to recover point-level detail lost during
pillarization. Flow4D~\cite{kim2025flow4d} instead uses five scans and
preserves height and time in a 4D voxel network, with decomposed 3D
spatial and 1D temporal convolutions for efficient spatiotemporal
fusion. SSF~\cite{khoche2025ssf} targets long-range scalability using a
sparse-convolutional U-Net and aligns the unequal occupied-voxel sets
of two scans with masked virtual voxels before feature fusion. Thus, the
baselines vary spatial representation, temporal context, and
fusion/decoding strategy while retaining a practical supervised,
feed-forward setting; for each, \emph{+vel} appends radial velocity to
the otherwise unchanged input and training configuration.

\subsection{Semantic Segmentation}
\label{sec:semantic-segmentation-baseline}
We evaluate Cylinder3D~\cite{zhu2021cylinder3d}, which applies asymmetrical
3D convolutions over cylindrical voxels, and Point Transformer V3
(PTv3)~\cite{wu2024ptv3}, which uses point-based transformer blocks.
For each method, \emph{+vel} appends compensated radial velocity to the
geometry and reflectivity features under otherwise unchanged settings
(see supplementary
\cref{sec:learned-implementation-supplement,tab:learned-implementation}).

%% file: sections/experiments.tex
\section{Experiments}
\label{sec:experiments}

We evaluate all baselines on the 10{,}000-frame test split using the
protocol in \cref{sec:benchmark}. Baseline methods use XYZ and reflectivity as input features, while \emph{+vel}
adds ego-motion-compensated radial velocity as an additional channel.
We evaluate one-frame input and five-frame input formed by combining
the current frame with four preceding ego-motion-compensated frames.
Within each architecture, the training and evaluation settings are
unchanged across input variants (see supplementary
\cref{sec:learned-implementation-supplement,tab:learned-implementation}).

\begin{figure*}[t]
\centering
\includegraphics[page=1, trim=1cm 0cm 1cm 1cm, clip, width=\textwidth]{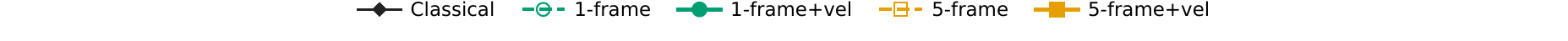}\\[-2pt]
\includegraphics[page=2, trim=0cm 0cm 0cm 0cm, clip, width=\textwidth]{images/metrics.pdf}\\[0pt]
\includegraphics[page=3, trim=0cm 0cm 0cm 0cm, clip, width=\textwidth]{images/metrics.pdf}\\[0pt]
\includegraphics[page=6, trim=0cm 0cm 0cm 0cm, clip, width=\textwidth]{images/metrics.pdf}\\[0pt]
\includegraphics[page=7, trim=0cm 0cm 0cm 0cm, clip, width=\textwidth]{images/metrics.pdf}

\caption{\textbf{Doppler improves single-frame and multi-frame long-range detection.}
We report AP ($\uparrow$) by range for dynamic and
all (static and dynamic) vehicles and pedestrians. At 200--400\,m, adding Doppler to single-frame detectors consistently raises
dynamic-vehicle AP from $0.33{\rightarrow}0.67$ (CenterPoint),
$0.31{\rightarrow}0.63$ (DSVT), $0.39{\rightarrow}0.68$ (VoxelNeXt), and
$0.50{\rightarrow}0.72$ (TransFusion). These velocity-aware single-frame
detectors match or outperform their five-frame XYZ-only counterparts
($0.52$, $0.62$, $0.56$, and $0.60$ AP, respectively), using a 100\,ms
rather than 500\,ms observation window. At the same range, all four
single-frame XYZ-only models obtain $0.00$ dynamic-pedestrian AP, while
Doppler raises VoxelNeXt to $0.12$ and TransFusion to $0.42$. With five
frames, Doppler further raises all-pedestrian AP from
$0.01{\rightarrow}0.13$ for CenterPoint and $0.13{\rightarrow}0.33$ for
TransFusion, showing that instantaneous motion and accumulated geometry
provide complementary cues.}
\label{fig:object_detection_metrics}
\end{figure*}

\subsection{Classical Object Detection}
The classical learning-free detector makes use of
thresholded Doppler and geometric clustering
(\cref{sec:classical-baseline}). At 200--400\,m, it obtains $0.41$
dynamic-vehicle AP, a higher reported point estimate than one-frame
no-Doppler CenterPoint ($0.33$), DSVT ($0.31$), and VoxelNeXt
($0.39$), and a lower estimate than TransFusion ($0.50$). Candidate
generation requires $|d_i|>0.7$\,m/s. Across points enclosed by the
over 8 million annotated 3D boxes, 97\% have
$|d_i|>0.5$\,m/s, indicating that informative radial motion is
available for nearly all annotated object points. The classical result
therefore provides a competitive long-range reference without
learning, while the learned detectors additionally cover stationary
and low-radial-velocity objects.

\subsection{Learned Object Detection}

\noindent\textbf{Single-frame detection.}
Adding ego-motion-compensated radial velocity increasingly benefits single-frame detection as geometry becomes sparse: dynamic-vehicle AP increases from $0.09$--$0.20$ at 0--100\,m to $0.22$--$0.34$ at 200--400\,m. At 200--400\,m, one-frame \emph{+vel} matches or exceeds five-frame XYZ-only AP for CenterPoint ($0.67$ vs.\ $0.52$), DSVT ($0.63$ vs.\ $0.62$), VoxelNeXt ($0.68$ vs.\ $0.56$), and TransFusion ($0.72$ vs.\ $0.60$), showing that instantaneous Doppler can replace much of the evidence obtained through temporal accumulation. Pedestrian performance is architecture-dependent: TransFusion reaches $0.42$ dynamic AP compared with $0.00$ without velocity >100m, whereas VoxelNeXt reaches $0.12$ and the other detectors remain near zero >200m. Single-frame inference also avoids maintaining and aligning temporal history, supporting immediate startup and recovery from dropped frames while potentially reducing memory, computation, latency, and cross-frame registration errors.

\noindent\textbf{Multi-frame detection.}
With five-frame input, velocity remains valuable where accumulated geometry is sparse. At 200--400\,m, \emph{+vel} improves dynamic-vehicle AP by $0.01$--$0.05$ and all-motion-state vehicle AP by $0.04$--$0.06$ across all detectors, while full-range DSVT improves from $0.72{\rightarrow}0.78$. Long-range pedestrian gains are larger: all-motion-state AP increases from $0.01{\rightarrow}0.13$ for CenterPoint, $0.23{\rightarrow}0.29$ for VoxelNeXt, and $0.13{\rightarrow}0.33$ for TransFusion; dynamic-pedestrian AP increases from $0.04{\rightarrow}0.28$ for CenterPoint and $0.24{\rightarrow}0.30$ for TransFusion. These results show that temporal stacking and Doppler provide complementary evidence: stacking increases spatial support and recovers object shape, whereas radial velocity supplies a per-point motion cue helping separate moving objects from static background.

\begin{figure*}[!t]
\centering
\includegraphics[page=10, trim=0cm 0cm 0cm 1cm, clip, width=\textwidth]{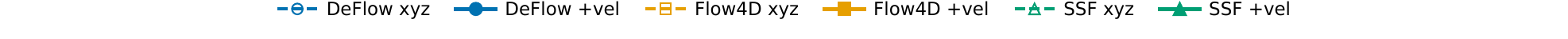}

\includegraphics[page=11, trim=0cm 0.2cm 0cm 1cm, clip, width=\textwidth]{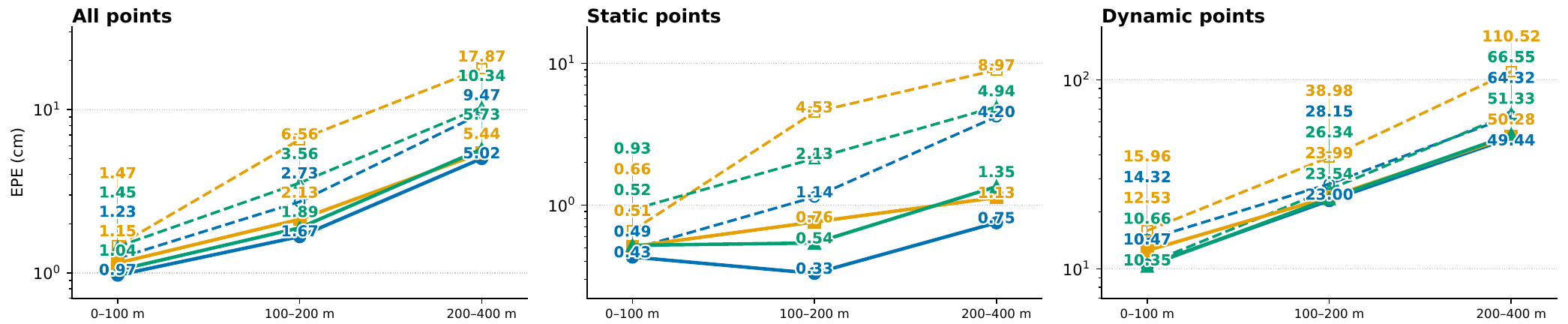}

\caption{\textbf{Scene-flow EPE ($\downarrow$) by input configuration and range.}
Adding radial velocity reduces EPE for every plotted method, range, and
point subset. At 200--400\,m, all-point EPE decreases from
$9.47{\rightarrow}5.02$\,cm for DeFlow,
$17.87{\rightarrow}5.44$\,cm for Flow4D, and
$10.34{\rightarrow}5.73$\,cm for SSF.}
\label{fig:scene_flow_metrics}
\end{figure*}

\noindent\textbf{Long-range detection (200--400\,m).}
At 200--400\,m, where annotated objects contain approximately 13--15 returns (\cref{fig:box_stats}), velocity provides its largest gains. One-frame \emph{+vel} matches or exceeds five-frame XYZ-only dynamic-vehicle AP for all four detectors, with all-motion-state margins of $0.19$, $0.04$, $0.12$, and $0.14$ for CenterPoint, DSVT, VoxelNeXt, and TransFusion, respectively. The classical Doppler detector reaches $0.41$ dynamic-vehicle AP, exceeding one-frame XYZ-only CenterPoint ($0.33$), DSVT ($0.31$), and VoxelNeXt ($0.39$), \emph{+vel} raises all learned detectors to $0.63$--$0.72$. For pedestrians, all one-frame XYZ-only models obtain $0.00$ dynamic AP at the reported precision, whereas \emph{+vel} reaches $0.42$ for TransFusion and $0.12$ for VoxelNeXt. Thus, FMCW velocity provides an effective per-return cue when sparse geometry and cross-frame misalignment limit temporal accumulation (see supplementary \cref{tab:dynamic-range-ablation,tab:all-metric-ablation,fig:point-binned,tab:point-binned-dynamic,tab:point-binned-all}).

\subsection{Scene Flow}
Adding ego-motion-compensated velocity consistently improves scene-flow estimation, reducing EPE in all 36 method--point-type--range combinations. Averaged across methods, all-point EPE decreases by 23.9\% at 0--100\,m, 55.7\% at 100--200\,m, and 57.0\% at 200--400\,m, showing that velocity becomes increasingly valuable as geometric evidence grows sparse. Static-point estimation benefits particularly strongly, with far-range EPE reductions of 72.7--87.4\%. For dynamic points, Doppler directly constrains the radial component of motion, reducing flow ambiguity and lowering far-range EPE by 23.1\% for DeFlow, 54.5\% for Flow4D, and 22.9\% for SSF. At 200--400\,m, velocity brings all-point EPE to a consistent 5.0--5.7\,cm and dynamic-point EPE to 49--51\,cm across methods. The smaller full-range reductions reflect the dominance of dense near-range points, where geometry is already informative. Overall, velocity provides a direct motion prior that consistently reduces error and becomes especially effective at longer ranges (see supplementary \cref{sec:scene-flow-supplement,tab:scene-flow-metric-v2}).

%% file: sections/conclusion.tex
\section{Conclusion}
\label{sec:conclusion}

We presented a large-scale FMCW LiDAR dataset with 575 sequences (57{,}500 frames), over 8 million annotated 3D boxes across 16 detection classes, including 237 nighttime sequences, and labels extending to 400\,m. The dataset also includes per-point labels across 24 semantic classes. Our analysis shows that Doppler becomes increasingly valuable as geometric evidence grows sparse. Across all four detection architectures, one-frame velocity-aware models match or exceed five-frame geometry-only vehicle detection at long range, while also improving pedestrian detection. Temporal aggregation and Doppler provide complementary geometric and motion evidence. Doppler also consistently reduces scene-flow error, although tangential motion must still be inferred. We hope this work accelerates perception systems that fully leverage the rich motion information provided by FMCW LiDAR.

%% file: sections/supplementary.tex
\section{Supplementary Material}
\label{sec:supplement}
\renewcommand{\bottomfraction}{0.55}
\makeatletter
\newcommand{\allowAlgorithmHere}{\@twocolumnfalse}
\makeatother

\subsection{Dataset}
\label{sec:dataset-supplement}

Fig.~\ref{fig:bev_all} presents bird's-eye-view visualizations of a
representative scene from our dataset. Unlike conventional ToF LiDAR, each
point carries a direct radial velocity measurement. Ego-motion compensation
removes the velocity contribution of the ego vehicle, leaving only the
residual motion of dynamic objects in the scene, making moving agents
immediately distinguishable from static background points. Reflectivity
returns remain clean and artifact-free at long range, with no blooming or
saturation effects. The dataset further provides dense 3D bounding box
annotations and per-point semantic labels, spanning a full 800\,m field of
view ($-$400\,m to $+$400\,m).

\newcommand{\datasetFigure}{%
\begin{figure*}[p]
\centering
\captionsetup[sub]{
  font=small,
  skip=2pt,
  justification=centering,
  singlelinecheck=false,
  labelsep=space
}
\begin{subfigure}[t]{0.16\textwidth}
  \centering
  \resizebox{\linewidth}{!}{%
    \includegraphics[angle=90]{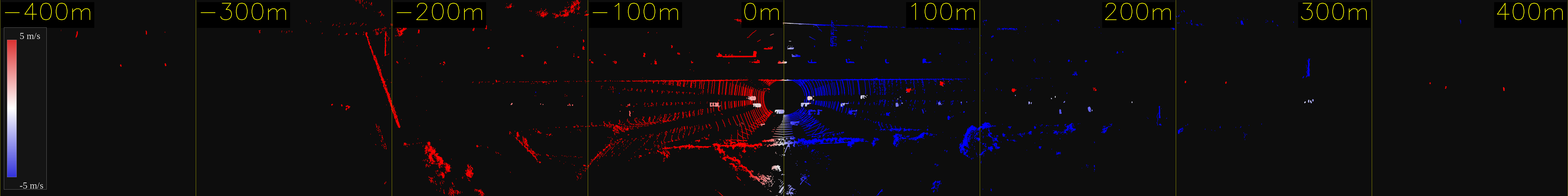}
  }
  \caption{\scriptsize Doppler Velocity (raw)}
\end{subfigure}\hspace{0.001\textwidth}%
\begin{subfigure}[t]{0.16\textwidth}
  \centering
  \resizebox{\linewidth}{!}{%
    \includegraphics[angle=90]{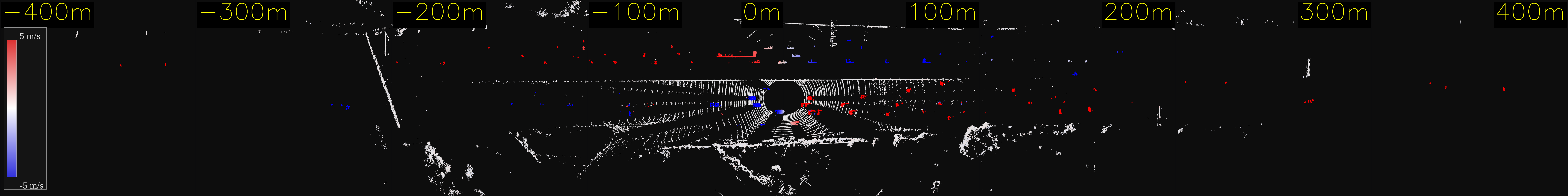}
  }
  \caption{\scriptsize Doppler Velocity (ego-motion compensated)}
\end{subfigure}\hspace{0.001\textwidth}%
\begin{subfigure}[t]{0.16\textwidth}
  \centering
  \resizebox{\linewidth}{!}{%
    \includegraphics[angle=90]{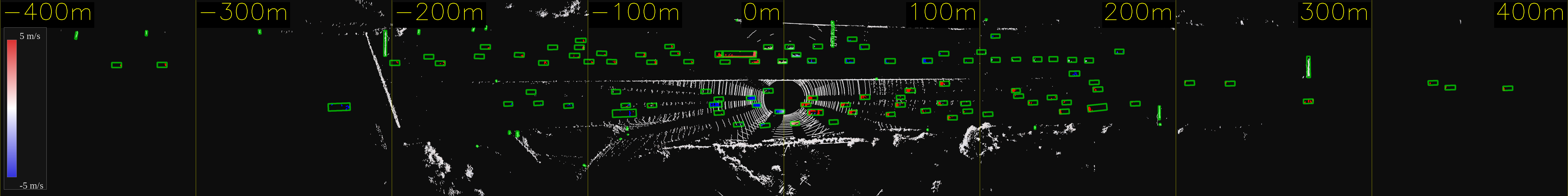}
  }
  \caption{\scriptsize Doppler Velocity (ego-motion compensated) with labelled boxes}
\end{subfigure}\hspace{0.001\textwidth}%
\begin{subfigure}[t]{0.16\textwidth}
  \centering
  \resizebox{\linewidth}{!}{%
    \includegraphics[angle=90]{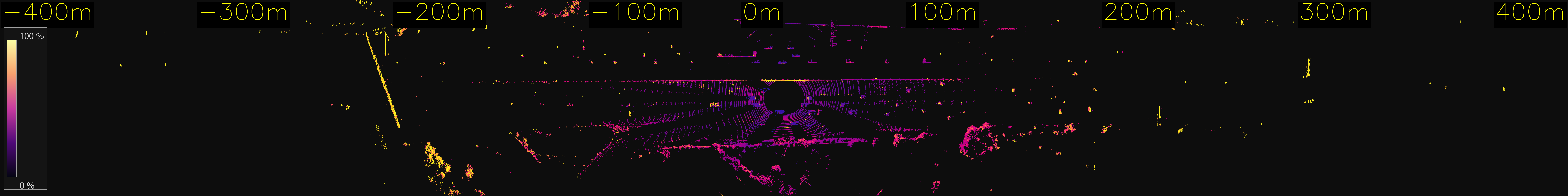}
  }
  \caption{\scriptsize Reflectivity}
\end{subfigure}\hspace{0.001\textwidth}%
\begin{subfigure}[t]{0.16\textwidth}
  \centering
  \resizebox{\linewidth}{!}{%
    \includegraphics[angle=90]{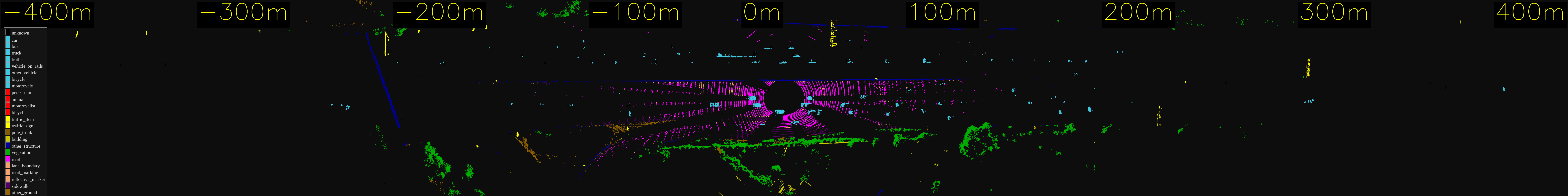}
  }
  \caption{\scriptsize Semantic Labels}
\end{subfigure}
\captionsetup{skip=0pt}
\caption{Bird's-eye-view visualization of FMCW LiDAR point cloud.}
\label{fig:bev_all}
\end{figure*}
}

\subsection{Classical Dynamic Object Detection and Tracking}
\label{sec:classical-baseline-supplement}

We provide a classical, learning-free detection and tracking baseline
exploiting per-point Doppler. Unlike ToF LiDAR's multi-frame differencing,
FMCW measures instantaneous radial velocity per return, enabling single-frame
detection at 200--400\,m despite sparse geometry.

\begin{figure}[t]
\centering
\resizebox{0.92\columnwidth}{!}{%
\begin{minipage}{\columnwidth}
\allowAlgorithmHere
\begin{algorithm}[H]
\caption{Classical Dynamic Object Detection and Tracking}
\label{alg:pipeline}
\DontPrintSemicolon
\footnotesize
\KwIn{Fused point cloud $\{(\mathbf{p}_i, d_i', \mathbf{o}_i)\}$ with positions, compensated Doppler, and sensor origins; ego pose $T_k$}
\KwOut{Tracked objects with position, velocity, heading, dimensions}
\tcp{Detection}
$\mathcal{S} \gets \{\mathbf{p} \in \mathcal{P} : \|\mathbf{p}_{xy}\| \leq 400\text{\,m},\; |d_i'| > 0.7\text{\,m/s}\}$ \tcp*{Dynamic seeds}
$\mathcal{S} \gets \text{VoxelDownsample}(\mathcal{S},\; \delta\!=\!0.2\text{\,m})$ \tcp*{Median Doppler per voxel}
$\mathcal{L} \gets \text{DBSCAN}(\mathcal{S}_{xy},\; \varepsilon\!=\!1.0\text{\,m},\; n_{\min}\!=\!5)$ \cite{ester1996dbscan}\;
Velocity split: run 1D DBSCAN ($\varepsilon_v\!=\!2.0$\,m/s) on Doppler within each cluster\;
Velocity merge: rejoin pairs with $\|\bar{\mathbf{p}}_j-\bar{\mathbf{p}}_k\|<3$\,m and $|\tilde{d}_j-\tilde{d}_k|<2$\,m/s\;
\tcp{Cluster velocity estimation}
\ForEach{cluster $C_j$}{
  $\hat{\mathbf{r}}_i \gets (\mathbf{p}_i-\mathbf{o}_i)_{xy}/\|\mathbf{p}_i-\mathbf{o}_i\|$ \tcp*{LOS unit vectors}
  \eIf{azimuth span $\geq8^{\circ}$ and well-conditioned}{
    $\mathbf{v}_j \gets \arg\min \|W(R\mathbf{v}-\mathbf{d})\|^2$; flip if $\langle R\mathbf{v}_j,\mathbf{d}\rangle<0$ \tcp*{WLS (Eq.~\ref{eq:doppler-model})}
  }{
    $\mathbf{v}_j \gets \text{median}(d_i')\cdot[\cos\bar{\theta},\,\sin\bar{\theta}]^T$ \tcp*{Radial fallback}
  }
  Fit heading-aligned box $\mathbf{b}_j$ with $\psi_j\!=\!\text{atan2}(v_y,v_x)$; discard outlier clusters\;
}
$\mathcal{D} \gets \text{NMS}(\{\mathbf{b}_j\})$ with floored dims and velocity gate\;
\tcp{Tracking}
Ego-motion compensate all tracks; predict forward with CTRV model\;
$(\mathcal{M},\mathcal{U}_t,\mathcal{U}_d)\gets\text{Hungarian}(\mathcal{T},\mathcal{D})$ using BEV IoU + velocity cost\;
\ForEach{matched $(t_i,d_j)\in\mathcal{M}$}{
  \lIf{radial fallback}{inflate $R_{\psi}$ by $10^4$}
  UKF update \cite{julier1997ukf} with displacement heading blending\;
}
Init new tracks from $\mathcal{U}_d$; prune stale tracks; post-process and NMS\;
\Return $\mathcal{T}$\;
\end{algorithm}
\vspace{3pt}
\input{tables/pipeline_parameters}
\end{minipage}
}
\end{figure}

Table~\ref{tab:supp-params} lists all configurable parameters for the
classical pipeline, and Table~\ref{tab:supp-tiers} details the range-tier
overrides that adapt clustering and tracking behavior across the full
400\,m operating range.

\datasetFigure

\begin{table}[htbp]
\centering
\caption{Range-tier overrides. Tiers are checked in order; the first tier
whose max range covers the object is used.}
\label{tab:supp-tiers}
\resizebox{\linewidth}{!}{%
\begin{tabular}{@{}rrrrrr@{}}
\toprule
Range & Min pts & Max resid & Hits & Skip & Assoc dist \\
(m) & (cluster) & (m/s) & (mature) & (max) & (m) \\
\midrule
0--50 & 8 & 5.0 & 3 & 5 & 8 \\
50--100 & 3 & 5.0 & 2 & 5 & 8 \\
100--200 & 3 & 8.0 & 2 & 5 & 12 \\
200--300 & 3 & 8.0 & 2 & 5 & 15 \\
300--400 & 4 & 6.0 & 3 & 3 & 15 \\
\bottomrule
\end{tabular}%
}
\end{table}

\paragraph{Ego-motion compensation}
Between frames, all track states (position, velocity heading) are
transformed into the current ego frame using the relative pose
$T_{k-1}^{-1}T_k$. The position history used for displacement heading is
similarly compensated.

\paragraph{CTRV motion model}
The prediction step propagates position via
\begin{equation}
  \begin{bmatrix}x\\y\end{bmatrix}_{k+1}
  =
  \begin{bmatrix}x\\y\end{bmatrix}_k
  +\frac{s}{\dot{\psi}}
  \begin{bmatrix}
    \sin(\psi+\dot{\psi}\Delta t)-\sin\psi\\
    \cos\psi-\cos(\psi+\dot{\psi}\Delta t)
  \end{bmatrix},
\end{equation}
with a straight-line approximation when $|\dot{\psi}|<0.001$\,rad/s.
UKF sigma points use Merwe scaling ($\alpha\!=\!1$, $\beta\!=\!2$,
$\kappa\!=\!0$).

\paragraph{Adaptive measurement noise}
For radial-only fallback detections, the yaw measurement noise $R_{\psi}$ is
inflated by $10^4\!\times$, effectively zeroing the Kalman gain for heading
so the CTRV prediction carries it forward.

\paragraph{Displacement heading injection}
For tracks with at least 3 frames of position history and net displacement
above a speed-dependent threshold, a displacement-derived heading is computed
from ego-compensated position deltas and blended into the UKF update, clamped
to $8^{\circ}$/frame.

\paragraph{Data association}
The Hungarian algorithm is used with cost
$C_{ij}=(1-\text{IoU}_{ij})+0.3\cdot\|\mathbf{v}_i-\mathbf{v}_j\|/5.0$\,m/s.
Bounding box dimensions are floored to $5.0\times2.0$\,m before IoU
computation, and a center offset along the heading compensates for
single-face visibility bias.

\paragraph{Dimension smoothing}
Cluster dimensions are smoothed with an asymmetric EMA (shrink
$\alpha\!=\!0.7$, grow $\alpha\!=\!0.3$) to reject transient spikes while
allowing over-estimated dimensions to decay quickly.

\newcommand{\learnedImplementationTable}{%
\begin{figure*}[!b]
\centering
\begin{minipage}[t]{\textwidth}
\centering
\footnotesize
\setlength{\tabcolsep}{3pt}
\renewcommand{\arraystretch}{1.08}
\captionof{table}{\textbf{Learned-baseline implementation and evaluation.} Crops are
specified as $[x_{\min},y_{\min},z_{\min},x_{\max},y_{\max},z_{\max}]$.}
\label{tab:learned-implementation}

\textbf{(a) 3D object detection}\\[2pt]
\begin{tabular*}{\textwidth}{@{\extracolsep{\fill}}
  >{\raggedright\arraybackslash}p{0.11\textwidth}
  >{\raggedright\arraybackslash}p{0.20\textwidth}
  >{\raggedright\arraybackslash}p{0.18\textwidth}
  >{\raggedright\arraybackslash}p{0.43\textwidth}@{}}
\toprule
Method & Spatial crop (m) & Voxelization (m) & Optimizer and schedule \\
\midrule
CenterPoint & $[0,-72,-3,400,72,5]$ & $[0.1,0.1,0.2]$
& Adam OneCycle, LR $10^{-3}$, weight decay $0.01$,
  $p_{\mathrm{start}}=0.4$, division factor $10$ \\
DSVT & $[0,-74,-2,400,74,4]$ & Dynamic $[0.4,0.4,0.2]$
& Adam OneCycle, LR $3{\times}10^{-3}$, weight decay $0.05$,
  $p_{\mathrm{start}}=0.1$, division factor $100$ \\
VoxelNeXt & $[0,-74,-2,400,74,4]$ & $[0.1,0.1,0.15]$
& Adam OneCycle, LR $3{\times}10^{-3}$, weight decay $0.05$,
  $p_{\mathrm{start}}=0.1$, division factor $100$ \\
TransFusion & $[0,-72,-3,400,72,5]$ & $[0.1,0.1,0.2]$
& Adam OneCycle, LR $10^{-3}$, weight decay $0.01$,
  $p_{\mathrm{start}}=0.4$, division factor $10$ \\
\bottomrule
\end{tabular*}

\vspace{2pt}
\textbf{(b) Detection point features}\\[2pt]
\begin{tabular*}{\textwidth}{@{\extracolsep{\fill}}
  >{\raggedright\arraybackslash}p{0.20\textwidth}
  >{\raggedright\arraybackslash}p{0.72\textwidth}@{}}
\toprule
Input variant & Per-point features \\
\midrule
1-frame $xyz$ & $[x,y,z,\text{reflectivity}]$ \\
1-frame $+\mathrm{vel}$ & $[x,y,z,\text{reflectivity},v_r]$ \\
5-frame $xyz$ & $[x,y,z,\text{frame index},\text{reflectivity}]$ \\
5-frame $+\mathrm{vel}$
& $[x,y,z,\text{frame index},\text{reflectivity},v_r]$ \\
\bottomrule
\end{tabular*}
\end{minipage}
\end{figure*}
}

\newcommand{\learnedImplementationTableContinuation}{%
\begin{figure*}[!t]
\centering
\begin{minipage}[t]{\textwidth}
\centering
\footnotesize
\setlength{\tabcolsep}{3pt}
\renewcommand{\arraystretch}{1.08}
\textbf{Table~\ref{tab:learned-implementation} (continued)}\\[2pt]

\vspace{2pt}
\textbf{(c) Scene flow}\\[2pt]
\begin{tabular*}{\textwidth}{@{\extracolsep{\fill}}
  >{\raggedright\arraybackslash}p{0.10\textwidth}
  >{\raggedright\arraybackslash}p{0.18\textwidth}
  >{\raggedright\arraybackslash}p{0.24\textwidth}
  >{\raggedright\arraybackslash}p{0.16\textwidth}
  >{\raggedright\arraybackslash}p{0.24\textwidth}@{}}
\toprule
Method & Input & Spatial crop (m) & Voxelization (m) & Optimization \\
\midrule
DeFlow & $xyz$ or $xyz+\mathrm{vel}$ & $[0,-100,-3,400,100,3]$
& $[0.4,0.2,6.0]$ & Adam, LR $2{\times}10^{-4}$, clip $5.0$ \\
Flow4D & 2-frame $xyz$ or $xyz+\mathrm{vel}$ & $[0,-100,-3,400,100,3]$
& $[0.4,0.4,0.3]$ & LR $10^{-3}$, clip $5.0$ \\
SSF & $xyz$ or $xyz+\mathrm{vel}$ & $[0,-100,-3,400,100,3]$
& $[0.2,0.2,6.0]$ & LR $2{\times}10^{-3}$, clip $5.0$ \\
\bottomrule
\end{tabular*}

\vspace{2pt}
\textbf{(d) Semantic segmentation}\\[2pt]
\begin{tabular*}{\textwidth}{@{\extracolsep{\fill}}
  >{\raggedright\arraybackslash}p{0.10\textwidth}
  >{\raggedright\arraybackslash}p{0.26\textwidth}
  >{\raggedright\arraybackslash}p{0.24\textwidth}
  >{\raggedright\arraybackslash}p{0.32\textwidth}@{}}
\toprule
Method & Point input & Spatial discretization & Optimizer and schedule \\
\midrule
Cylinder3D & $xyz+r$ or $xyz+r+\tanh(v_r/60)$
& Cylindrical grid $480{\times}360{\times}32$
& AdamW, LR $10^{-3}$, weight decay $0.01$; 1K-iteration linear warmup;
  step decay at epoch 30 ($\gamma=0.1$) \\
PTv3 & $xyz+r$ or $xyz+r+\tanh(v_r/60)$
& $0.05$\,m train/validation; $0.025{\rightarrow}0.05$\,m test
& AdamW, LR $2{\times}10^{-3}$, weight decay $0.005$;
  OneCycle with cosine annealing \\
\bottomrule
\end{tabular*}

\vspace{2pt}
\textbf{(e) Evaluation protocol}\\[2pt]
\begin{tabular*}{\textwidth}{@{\extracolsep{\fill}}
  >{\raggedright\arraybackslash}p{0.16\textwidth}
  >{\raggedright\arraybackslash}p{0.76\textwidth}@{}}
\toprule
Task & Evaluation protocol \\
\midrule
3D object detection
& BEV-center matching with a 2\,m threshold for vehicles/unknown and 1\,m
for pedestrians. AP is reported over 0--100, 100--200, and 200--400\,m and
over the full 0--400\,m range. Dynamic and static objects are separated at
1.0\,m/s. \\
Semantic segmentation
& Per-class IoU and mIoU over all 24 semantic classes. \\
Scene flow
& Absolute end-point error over all, static, and dynamic points in the
0--100, 100--200, and 200--400\,m intervals and over the full range. \\
\bottomrule
\end{tabular*}
\end{minipage}

\par\vspace{14pt}
\input{tables/detection_table_v2_style}
\setcounter{table}{9}
\input{tables/scene_flow_metric_table_v2}
\par\vspace{8pt}
\input{tables/semseg_metric_table_v2}
\setcounter{table}{5}
\end{figure*}
}
\learnedImplementationTable

\subsection{Learned Baseline Implementation}
\label{sec:learned-implementation-supplement}

Each physical scan is divided into front and rear hemispheres. The rear
hemisphere, including its annotations and Cartesian object velocities, is
rotated by $180^\circ$, yielding two samples spanning 0--400\,m. Radial
velocity is invariant to this rotation and is not sign-flipped. Unless
otherwise noted, the baselines use all six LiDARs and ego-motion-compensated
point clouds. Where reflectivity is used, values below 100 are multiplied by
0.8 and values above 100 are clipped to 100.

\subsubsection{3D Object Detection}

Table~\ref{tab:learned-implementation} summarizes the detector-specific
spatial discretization, optimization, and point features. Five-frame inputs
contain the current scan and four preceding scans transformed into the
current ego frame. No geometric, copy-paste, or ground-truth-sampling
augmentation is enabled.

\subsubsection{Scene Flow}

Scene-flow methods receive either ego-motion-compensated coordinates
($xyz$) or coordinates augmented with compensated radial velocity
($xyz+\mathrm{vel}$). The retained DeFlow configuration uses a frame stride
of two and randomly selects a hemisphere during training. Flow4D uses two
input frames. All three methods remove ground points at $z\leq-1.4$\,m.

\subsubsection{Semantic Segmentation}

Both segmentation methods use the crop $[0,-100,-2,400,100,4]$\,m.
Radial velocity is normalized as $\tanh(v_r/60)$ before being appended to
the point features.

\paragraph{Data augmentation}
Cylinder3D uses no augmentation beyond sensor fusion, spatial cropping, and
the hemisphere transform. PTv3 applies a random rotation about the vertical
axis in $[-1,1]$\,rad with probability $0.5$, scaling in $[0.9,1.1]$, random
flipping with probability $0.5$, and Gaussian jitter with standard deviation
$0.005$ clipped to $0.02$. Mixup is disabled.

\newpage
\paragraph{Evaluation protocol}
Task-specific metrics, matching thresholds, and range bins are summarized in
Table~\ref{tab:learned-implementation}(e).

\subsection{Object Detection Results}
\label{sec:detection-results-supplement}

Following Table~\ref{tab:learned-implementation}(e), we report vehicle and
pedestrian AP with 101-point interpolation for dynamic objects and for all
objects. Evaluation uses the 10K-frame test split. We additionally report
speed error (SpdE), defined as the L2 norm of the 3D velocity error over
matched detections.

\subsection{Scene Flow Estimation}
\label{sec:scene-flow-supplement}

Scene flow labels are derived from ground-truth human-annotated object tracks.
Unlike conventional LiDAR, FMCW sensing directly measures the radial
component of each point's velocity via Doppler, so the estimation problem
reduces to recovering the unobserved tangential component; methods can treat
the per-point radial measurement as a strong prior, both for gating static
points (compensated Doppler ${\approx}0$) and for constraining the flow of
dynamic ones. The primary metric is absolute end-point error (EPE), reported
on all, static, and dynamic points using the same three range bins as
detection. We do not prescribe a fixed frame pairing, allowing methods to use
consecutive or multi-frame inputs.

\subsection{Semantic Segmentation}
\label{sec:semantic-segmentation-supplement}

We evaluate Cylinder3D~\cite{zhu2021cylinder3d} and Point Transformer
V3 (PTv3)~\cite{wu2024ptv3}. Cylinder3D partitions the point cloud in
cylindrical coordinates and processes it with asymmetrical 3D convolutions,
while PTv3 applies point-based transformer blocks. Each method is evaluated
with geometry-only ($xyz$) input and with ego-motion-compensated radial
velocity appended as an additional input channel ($+$vel).

The primary metrics are per-class intersection over union (IoU) and mean IoU
(mIoU) over all 24 semantic classes. Table~\ref{tab:semseg-v2} reports the 
complete results. Adding velocity changes mIoU from 41.7 to 39.7 for Cylinder3D
and from 49.1 to 44.3 for PTv3, with mixed effects across individual classes.
Although isolated dynamic classes improve, neither architecture produces
consistent gains across vehicles and vulnerable road users. These preliminary
results suggest that appending radial velocity as a generic feature does not
expose its motion structure effectively; motion-aware architectures and
objectives remain future work.

\begin{figure*}[!t]
\centering
\begin{minipage}[t]{\textwidth}
\centering
\footnotesize
\setlength{\tabcolsep}{3pt}
\renewcommand{\arraystretch}{1.08}
\textbf{Table~\ref{tab:learned-implementation} (continued)}\\[2pt]

\vspace{2pt}
\textbf{(c) Scene flow}\\[2pt]
\begin{tabular*}{\textwidth}{@{\extracolsep{\fill}}
  >{\raggedright\arraybackslash}p{0.10\textwidth}
  >{\raggedright\arraybackslash}p{0.18\textwidth}
  >{\raggedright\arraybackslash}p{0.24\textwidth}
  >{\raggedright\arraybackslash}p{0.16\textwidth}
  >{\raggedright\arraybackslash}p{0.24\textwidth}@{}}
\toprule
Method & Input & Spatial crop (m) & Voxelization (m) & Optimization \\
\midrule
DeFlow & $xyz$ or $xyz+\mathrm{vel}$ & $[0,-100,-3,400,100,3]$
& $[0.4,0.2,6.0]$ & Adam, LR $2{\times}10^{-4}$, clip $5.0$ \\
Flow4D & 2-frame $xyz$ or $xyz+\mathrm{vel}$ & $[0,-100,-3,400,100,3]$
& $[0.4,0.4,0.3]$ & LR $10^{-3}$, clip $5.0$ \\
SSF & $xyz$ or $xyz+\mathrm{vel}$ & $[0,-100,-3,400,100,3]$
& $[0.2,0.2,6.0]$ & LR $2{\times}10^{-3}$, clip $5.0$ \\
\bottomrule
\end{tabular*}

\vspace{2pt}
\textbf{(d) Semantic segmentation}\\[2pt]
\begin{tabular*}{\textwidth}{@{\extracolsep{\fill}}
  >{\raggedright\arraybackslash}p{0.10\textwidth}
  >{\raggedright\arraybackslash}p{0.26\textwidth}
  >{\raggedright\arraybackslash}p{0.24\textwidth}
  >{\raggedright\arraybackslash}p{0.32\textwidth}@{}}
\toprule
Method & Point input & Spatial discretization & Optimizer and schedule \\
\midrule
Cylinder3D & $xyz+r$ or $xyz+r+\tanh(v_r/60)$
& Cylindrical grid $480{\times}360{\times}32$
& AdamW, LR $10^{-3}$, weight decay $0.01$; 1K-iteration linear warmup;
  step decay at epoch 30 ($\gamma=0.1$) \\
PTv3 & $xyz+r$ or $xyz+r+\tanh(v_r/60)$
& $0.05$\,m train/validation; $0.025{\rightarrow}0.05$\,m test
& AdamW, LR $2{\times}10^{-3}$, weight decay $0.005$;
  OneCycle with cosine annealing \\
\bottomrule
\end{tabular*}

\vspace{2pt}
\textbf{(e) Evaluation protocol}\\[2pt]
\begin{tabular*}{\textwidth}{@{\extracolsep{\fill}}
  >{\raggedright\arraybackslash}p{0.16\textwidth}
  >{\raggedright\arraybackslash}p{0.76\textwidth}@{}}
\toprule
Task & Evaluation protocol \\
\midrule
3D object detection
& BEV-center matching with a 2\,m threshold for vehicles/unknown and 1\,m
for pedestrians. AP is reported over 0--100, 100--200, and 200--400\,m and
over the full 0--400\,m range. Dynamic and static objects are separated at
1.0\,m/s. \\
Semantic segmentation
& Per-class IoU and mIoU over all 24 semantic classes. \\
Scene flow
& Absolute end-point error over all, static, and dynamic points in the
0--100, 100--200, and 200--400\,m intervals and over the full range. \\
\bottomrule
\end{tabular*}
\end{minipage}

\par\vspace{14pt}
\input{tables/detection_table_v2_style}
\setcounter{table}{9}
\input{tables/scene_flow_metric_table_v2}
\par\vspace{8pt}
\input{tables/semseg_metric_table_v2}
\setcounter{table}{5}
\end{figure*}

\begin{figure*}[p]
\centering
\begingroup
\def\best#1{#1}
\input{tables/detection_range_table}
\endgroup
\end{figure*}

\begin{figure*}[p]
\centering
\begingroup
\def\best#1{#1}
\input{tables/detection_all_metric_table}
\endgroup
\end{figure*}

\clearpage
\begin{figure*}[p]
\centering
\makebox[\textwidth][c]{%
\includegraphics[page=1, trim=0cm 0cm 0cm 1cm, clip,
width=1.05\textwidth]{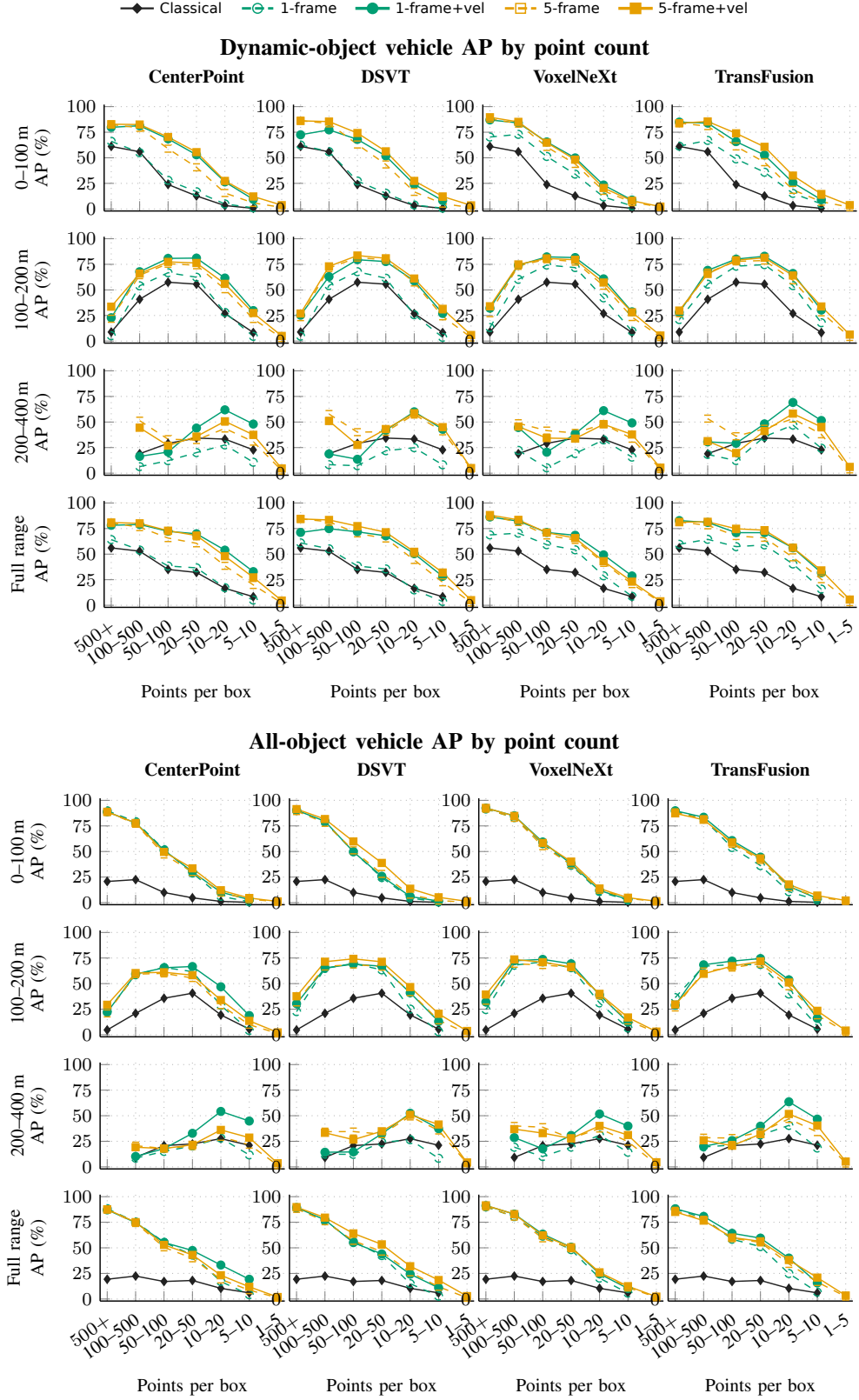}}\\[-2pt]
{\normalsize\bfseries Dynamic-object vehicle AP by point count\par}
\vspace{1pt}
\input{tables/point_binned_dynamic_plot}\\[5pt]
{\normalsize\bfseries All-object vehicle AP by point count\par}
\vspace{1pt}
\input{tables/point_binned_all_plot}
\caption{\textbf{Vehicle AP by point count for dynamic objects
(top) and all objects (bottom).} Rows denote evaluation range and columns
denote detector architecture; the classical baseline is repeated in each
column. At 200--400\,m with 5--10 points per box, Doppler improves one-frame
AP by 27.1--37.3 points for dynamic vehicles and 25.6--33.1 points for all
vehicles; five-frame gains are 4.1--7.2 and 5.8--6.9 points, respectively.}
\label{fig:point-binned}
\end{figure*}

\clearpage
\begin{figure*}[p]
\centering
\input{tables/point_binned_dynamic_table}
\par\vspace{2pt}\noindent
\input{tables/point_binned_all_table}
\end{figure*}
\FloatBarrier

%% file: tables/pipeline_parameters.tex
\centering
\captionof{table}{Pipeline parameters.}
\label{tab:supp-params}
\resizebox{0.94\linewidth}{!}{%
\begin{tabular}{@{}llr@{}}
\toprule
Stage & Parameter & Value \\
\midrule
Filter & Voxel size & 0.2\,m \\
 & Min dynamic speed & 0.7\,m/s \\
 & Max range & 400\,m \\
\midrule
Cluster & DBSCAN $\varepsilon$ & 1.0\,m \\
 & DBSCAN min points & 5 \\
 & Velocity split $\varepsilon_v$ & 2.0\,m/s \\
 & Merge distance & 3.0\,m \\
 & Merge velocity gate & 2.0\,m/s \\
\midrule
Velocity & Method & WLS \\
 & Min azimuth spread & 8$^{\circ}$ (12$^{\circ}$ $<$30\,m) \\
 & Min observability ratio & 0.12 \\
 & Max residual & 5.0\,m/s \\
 & Max speed & 50\,m/s \\
\midrule
Tracker & Process noise (accel) & 2.0\,m/s$^2$ \\
 & Process noise (yaw rate) & 0.3\,rad/s \\
 & Measurement noise (pos) & 0.5\,m \\
 & Measurement noise (vel) & 1.0\,m/s \\
 & Min bbox dims & $5.0 \times 2.0 \times 1.6$\,m \\
 & Min export speed & 1.0\,m/s \\
 & Max center shift & 2.0\,m \\
\midrule
Association & Method & BEV IoU + velocity \\
 & IoU threshold & 0.01 \\
 & Velocity weight $\lambda$ & 0.3 \\
 & Velocity gate & 5.0\,m/s \\
\bottomrule
\end{tabular}%
}

%% file: tables/detection_table_v2_style.tex
\providecommand{\best}[1]{\bfseries #1}
\definecolor{tblHead}{gray}{0.86}
\definecolor{tblSub}{gray}{0.93}
\definecolor{tblOneF}{gray}{0.935}
\definecolor{tblRule}{gray}{0.52}
\providecommand{\onfv}{\rowcolor{tblOneF}}
\providecommand{\mhv}[2]{\multicolumn{#1}{c|}{\cellcolor{tblHead}\textbf{#2}}}
\providecommand{\mhlv}[1]{\multicolumn{4}{c}{\cellcolor{tblHead}\textbf{#1}}}
\providecommand{\rngv}[1]{\rotatebox[origin=l]{48}{\scriptsize\hspace{0.15em}#1}}
\providecommand{\clsv}[1]{%
  \raisebox{0.2ex}[1.55cm][0pt]{%
    \rotatebox[origin=lb]{52}{\scriptsize\makebox[0pt][l]{#1}}}}
\providecommand{\stubv}[1]{\cellcolor{tblHead}\raisebox{1.15em}{#1}}
\providecommand{\gapf}{\addlinespace[2.8pt]}

%% file: tables/scene_flow_metric_table_v2.tex
\begin{minipage}[t]{\textwidth}
\centering
\scriptsize
\setlength{\tabcolsep}{3.2pt}
\renewcommand{\arraystretch}{1.18}
\setlength{\aboverulesep}{0pt}
\setlength{\belowrulesep}{0pt}
\setlength{\extrarowheight}{1.4pt}
\captionof{table}{Scene flow absolute EPE~(m)}
\label{tab:scene-flow-metric-v2}
\arrayrulecolor{tblRule}
\begin{tabular}{@{}ll!{\color{tblRule}\vrule\hspace{3.6pt}}
  *{4}{S[table-format=1.4]}
  !{\hspace{3.6pt}\color{tblRule}\vrule\hspace{3.6pt}}
  *{4}{S[table-format=1.4]}
  !{\hspace{3.6pt}\color{tblRule}\vrule\hspace{3.6pt}}
  *{4}{S[table-format=1.4]}
  @{}}
\arrayrulecolor{black}
\toprule
\arrayrulecolor{tblRule}
\rowcolor{tblHead}
& & \mhv{4}{All Points} & \mhv{4}{Static Points} & \mhlv{Dynamic Points} \\
\rowcolor{tblSub}
\rule{0pt}{2.55em}\stubv{Method} & \stubv{Input}
  & {\rngv{0--100\,m}} & {\rngv{100--200\,m}} & {\rngv{200--400\,m}} & {\rngv{Full Range}}
  & {\rngv{0--100\,m}} & {\rngv{100--200\,m}} & {\rngv{200--400\,m}} & {\rngv{Full Range}}
  & {\rngv{0--100\,m}} & {\rngv{100--200\,m}} & {\rngv{200--400\,m}} & {\rngv{Full Range}} \\
\midrule
\onfv \textbf{DeFlow} & $xyz$
  & 0.0123 & 0.0273 & 0.0947 & 0.0127
  & 0.0049 & 0.0114 & 0.0420 & 0.0051
  & 0.1432 & 0.2815 & 0.6432 & 0.1470 \\
& +vel
  & \best{0.0097} & \best{0.0167} & \best{0.0502} & \best{0.0099}
  & \best{0.0043} & \best{0.0033} & \best{0.0075} & \best{0.0043}
  & 0.1047 & \best{0.2300} & \best{0.4944} & 0.1080 \\
\midrule
\onfv \textbf{Flow4D} & $xyz$
  & 0.0147 & 0.0656 & 0.1787 & 0.0159
  & 0.0066 & 0.0453 & 0.0897 & 0.0074
  & 0.1596 & 0.3898 & 1.1052 & 0.1660 \\
& +vel
  & 0.0115 & 0.0213 & 0.0544 & 0.0118
  & 0.0051 & 0.0076 & 0.0113 & 0.0051
  & 0.1253 & 0.2399 & 0.5028 & 0.1284 \\
\midrule
\onfv \textbf{SSF} & $xyz$
  & 0.0145 & 0.0356 & 0.1034 & 0.0150
  & 0.0093 & 0.0213 & 0.0494 & 0.0095
  & 0.1066 & 0.2634 & 0.6655 & 0.1108 \\
& +vel
  & 0.0104 & 0.0189 & 0.0573 & 0.0106
  & 0.0052 & 0.0054 & 0.0135 & 0.0052
  & \best{0.1035} & 0.2354 & 0.5133 & \best{0.1069} \\
\arrayrulecolor{black}
\bottomrule
\end{tabular}
\arrayrulecolor{black}
\end{minipage}

%% file: tables/semseg_metric_table_v2.tex
\begin{minipage}[t]{\textwidth}
\centering
\scriptsize
\setlength{\tabcolsep}{2.6pt}
\renewcommand{\arraystretch}{1.18}
\setlength{\aboverulesep}{0pt}
\setlength{\belowrulesep}{0pt}
\setlength{\extrarowheight}{1.2pt}
\captionof{table}{Semantic segmentation on the test split (IoU~\%)}
\label{tab:semseg-v2}
\resizebox{\textwidth}{!}{%
\begin{tabular}{@{}ll *{25}{S[table-format=2.1]} @{}}
\toprule
\cellcolor{tblHead}\raisebox{0.62cm}{Method}
  & \cellcolor{tblHead}\raisebox{0.62cm}{Input}
  & {\clsv{Car}} & {\clsv{Bus}} & {\clsv{Truck}} & {\clsv{Trailer}}
  & {\clsv{Veh.\ on rails}} & {\clsv{Other veh.}}
  & {\clsv{Bicycle}} & {\clsv{Motorcycle}} & {\clsv{Pedestrian}} & {\clsv{Animal}}
  & {\clsv{Motorcyclist}} & {\clsv{Bicyclist}}
  & {\clsv{Traffic item}} & {\clsv{Traffic sign}} & {\clsv{Pole / trunk}}
  & {\clsv{Building}} & {\clsv{Other struct.}}
  & {\clsv{Vegetation}} & {\clsv{Road}} & {\clsv{Lane bound.}}
  & {\clsv{Road mark.}} & {\clsv{Refl.\ marker}} & {\clsv{Sidewalk}}
  & {\clsv{Other ground}}
  & {\clsv{\textbf{mIoU}}} \\
\midrule
\onfv \textbf{Cylinder3D} & $xyz$
  & 87.4 & 59.0 & 39.8 & 15.4 & 38.6 & 0.7
  & 18.7 & 17.0 & 66.8 & 2.1 & \best{56.3} & 5.1
  & 22.8 & 47.9 & 59.6 & 75.0 & 45.1
  & 75.7 & 85.3 & 31.0 & 34.1 & 4.3 & 55.6 & 58.8
  & 41.7 \\
& +vel
  & 86.8 & 42.6 & 32.4 & 17.5 & 14.7 & 1.0
  & 32.9 & 7.5 & 64.6 & 1.5 & 52.1 & 3.3
  & 22.4 & 50.5 & 57.8 & 76.5 & 50.4
  & 76.5 & 84.8 & 29.2 & 28.6 & \best{5.5} & 54.7 & 60.3
  & 39.7 \\
\midrule
\onfv \textbf{PTv3} & $xyz$
  & \best{87.9} & \best{82.8} & \best{51.0} & \best{20.2} & \best{66.5} & 2.5
  & \best{38.9} & 21.5 & 78.4 & 6.7 & 34.9 & \best{13.2}
  & \best{26.4} & \best{59.0} & \best{66.4} & \best{78.6} & 54.8
  & \best{81.5} & \best{87.2} & 40.1 & \best{44.5} & 5.4 & \best{66.3} & \best{64.6}
  & \best{49.1} \\
& +vel
  & 87.2 & 77.1 & 47.0 & 15.8 & 3.1 & \best{4.1}
  & 25.9 & \best{22.7} & \best{79.5} & \best{7.5} & 49.0 & 12.9
  & 21.5 & 56.9 & 62.0 & 77.6 & \best{58.6}
  & 75.6 & 86.1 & \best{41.0} & 43.9 & 5.4 & 57.9 & 46.2
  & 44.3 \\
\bottomrule
\end{tabular}}
\end{minipage}

%% file: tables/detection_range_table.tex
\begin{minipage}[t]{\textwidth}
\centering
\scriptsize
\setlength{\tabcolsep}{2.2pt}
\renewcommand{\arraystretch}{1.22}
\setlength{\aboverulesep}{0pt}
\setlength{\belowrulesep}{0pt}
\setlength{\extrarowheight}{1.2pt}
\providecommand{\best}[1]{\bfseries #1}
\definecolor{tblHead}{gray}{0.88}
\definecolor{pointOne}{HTML}{009E73}
\definecolor{pointFive}{HTML}{E69F00}
\definecolor{tblRule}{gray}{0.62}
\providecommand{\onf}{\rowcolor{pointOne!8}}
\providecommand{\fivef}{\rowcolor{pointFive!10}}
\providecommand{\oneinput}[1]{\textcolor{pointOne}{\bfseries #1}}
\providecommand{\fiveinput}[1]{\textcolor{pointFive}{\bfseries #1}}
\providecommand{\onebest}[1]{\textcolor{pointOne}{\bfseries #1}}
\providecommand{\fivebest}[1]{\textcolor{pointFive}{\bfseries #1}}
\providecommand{\mh}[2]{\multicolumn{#1}{c|}{\cellcolor{tblHead}\textbf{#2}}}
\providecommand{\mhl}[1]{\multicolumn{4}{c}{\cellcolor{tblHead}\textbf{#1}}}
\providecommand{\rng}[1]{\shortstack[c]{\scriptsize #1}}
\captionof{table}{Dynamic object detection. Best AP ($\uparrow$) and SpdE ($\downarrow$) values are bold within each frame setting; one-frame configurations are green and five-frame configurations are orange.}

\label{tab:dynamic-range-ablation}
\arrayrulecolor{tblRule}
\resizebox{\linewidth}{!}{%
\begin{tabular}{@{}ll|
  *{4}{S[table-format=1.2]}|
  *{4}{S[table-format=2.2]}|
  *{4}{S[table-format=1.2]}|
  *{4}{S[table-format=1.2]}
  @{}}
\arrayrulecolor{black}
\toprule
\arrayrulecolor{tblRule}
\rowcolor{tblHead}
& & \mh{8}{Vehicle} & \multicolumn{8}{c}{\cellcolor{tblHead}\textbf{Pedestrian}} \\
\rowcolor{tblHead}
& & \mh{4}{AP} & \mh{4}{SpdE (m/s)} & \mh{4}{AP} & \mhl{SpdE (m/s)} \\
\rowcolor{tblHead}
Method & Input
  & {\rng{0--100\\m}} & {\rng{100--200\\m}} & {\rng{200--400\\m}} & {\rng{Full\\Range}}
  & {\rng{0--100\\m}} & {\rng{100--200\\m}} & {\rng{200--400\\m}} & {\rng{Full\\Range}}
  & {\rng{0--100\\m}} & {\rng{100--200\\m}} & {\rng{200--400\\m}} & {\rng{Full\\Range}}
  & {\rng{0--100\\m}} & {\rng{100--200\\m}} & {\rng{200--400\\m}} & {\rng{Full\\Range}} \\
\midrule
\onf \textbf{Classical} & \oneinput{1-frame}
  & 0.61 & 0.59 & 0.41 & 0.58
  & \onebest{0.84} & \onebest{0.98} & \onebest{1.64} & \onebest{0.98}
  & {--} & {--} & {--} & {--}
  & {--} & {--} & {--} & {--} \\
\midrule
\onf \textbf{CenterPoint} & \oneinput{1-frame}
  & 0.64 & 0.65 & 0.33 & 0.62
  & 8.99 & 10.54 & 12.99 & 9.91
  & 0.03 & 0.02 & 0.00 & 0.03
  & 1.27 & 2.32 & {--} & 1.32 \\
\onf & \oneinput{\quad+vel}
  & 0.79 & 0.81 & 0.67 & 0.80
  & 3.99 & 3.28 & 3.41 & 3.68
  & 0.04 & 0.04 & 0.00 & 0.04
  & 0.99 & 0.54 & {--} & 0.94 \\
\fivef & \fiveinput{5-frame}
  & 0.79 & 0.75 & 0.52 & 0.75
  & 1.65 & 1.88 & 3.37 & 1.92
  & 0.36 & 0.22 & 0.04 & 0.35
  & 0.28 & 0.32 & 2.12 & 0.28 \\
\fivef & \fiveinput{\quad+vel}
  & 0.81 & 0.77 & 0.57 & 0.78
  & 1.50 & 1.64 & 3.00 & 1.72
  & 0.25 & 0.22 & 0.28 & 0.25
  & 0.30 & 0.33 & 1.21 & 0.30 \\
\midrule
\onf \textbf{DSVT} & \oneinput{1-frame}
  & 0.64 & 0.66 & 0.31 & 0.62
  & 9.04 & 10.92 & 12.54 & 10.02
  & 0.02 & 0.02 & 0.00 & 0.02
  & 2.40 & 3.16 & 5.03 & 2.46 \\
\onf & \oneinput{\quad+vel}
  & 0.73 & 0.77 & 0.63 & 0.74
  & 5.41 & 6.27 & 8.53 & 6.04
  & 0.05 & 0.02 & 0.03 & 0.05
  & 0.97 & 1.84 & 2.94 & 1.00 \\
\fivef & \fiveinput{5-frame}
  & 0.81 & 0.79 & 0.62 & 0.80
  & 1.92 & 2.09 & 4.08 & 2.21
  & 0.38 & 0.22 & 0.04 & 0.37
  & 0.34 & 0.39 & \fivebest{0.75} & 0.34 \\
\fivef & \fiveinput{\quad+vel}
  & 0.82 & 0.81 & \fivebest{0.63} & 0.81
  & 2.10 & 2.83 & 6.15 & 2.79
  & 0.36 & 0.13 & 0.01 & 0.34
  & 0.26 & 0.37 & 1.34 & 0.27 \\
\midrule
\onf \textbf{VoxelNeXt} & \oneinput{1-frame}
  & 0.72 & 0.72 & 0.39 & 0.71
  & 3.93 & 6.26 & 9.75 & 5.32
  & 0.16 & 0.05 & 0.00 & 0.15
  & 0.41 & 0.78 & \onebest{0.00} & 0.42 \\
\onf & \oneinput{\quad+vel}
  & 0.85 & 0.83 & 0.68 & \onebest{0.84}
  & 1.68 & 2.32 & 3.65 & 2.09
  & 0.24 & 0.21 & 0.12 & 0.24
  & 0.32 & 0.40 & 1.48 & 0.32 \\
\fivef & \fiveinput{5-frame}
  & 0.84 & 0.79 & 0.56 & 0.81
  & 1.30 & 1.69 & 3.39 & 1.65
  & 0.61 & 0.43 & \fivebest{0.46} & 0.59
  & \fivebest{0.22} & \fivebest{0.27} & 0.80 & \fivebest{0.23} \\
\fivef & \fiveinput{\quad+vel}
  & \fivebest{0.87} & \fivebest{0.82} & 0.59 & \fivebest{0.83}
  & \fivebest{1.22} & \fivebest{1.57} & 3.20 & \fivebest{1.55}
  & \fivebest{0.66} & \fivebest{0.54} & 0.43 & \fivebest{0.65}
  & 0.26 & 0.28 & 0.78 & 0.26 \\
\midrule
\onf \textbf{TransFusion} & \oneinput{1-frame}
  & 0.66 & 0.73 & 0.50 & 0.68
  & 5.10 & 5.88 & 8.51 & 5.68
  & 0.02 & 0.00 & 0.00 & 0.01
  & 2.24 & 2.38 & {--} & 2.24 \\
\onf & \oneinput{\quad+vel}
  & \onebest{0.86} & \onebest{0.84} & \onebest{0.72} & \onebest{0.84}
  & 1.80 & 1.71 & 2.32 & 1.82
  & \onebest{0.38} & \onebest{0.42} & \onebest{0.42} & \onebest{0.39}
  & \onebest{0.27} & \onebest{0.30} & 1.12 & \onebest{0.28} \\
\fivef & \fiveinput{5-frame}
  & 0.82 & 0.80 & 0.60 & 0.80
  & 1.61 & 1.76 & \fivebest{2.76} & 1.78
  & 0.35 & 0.29 & 0.24 & 0.34
  & 0.30 & 0.39 & 1.93 & 0.31 \\
\fivef & \fiveinput{\quad+vel}
  & 0.84 & 0.81 & \fivebest{0.63} & 0.82
  & 1.55 & 1.96 & 3.04 & 1.85
  & 0.21 & 0.21 & 0.30 & 0.21
  & 0.34 & 0.40 & 1.77 & 0.35 \\
\arrayrulecolor{black}
\bottomrule
\end{tabular}}
\arrayrulecolor{black}
\end{minipage}

%% file: tables/detection_all_metric_table.tex
\begin{minipage}[t]{\textwidth}
\centering
\scriptsize
\setlength{\tabcolsep}{2.2pt}
\renewcommand{\arraystretch}{1.22}
\setlength{\aboverulesep}{0pt}
\setlength{\belowrulesep}{0pt}
\setlength{\extrarowheight}{1.2pt}
\providecommand{\best}[1]{\bfseries #1}
\definecolor{tblHead}{gray}{0.88}
\definecolor{pointOne}{HTML}{009E73}
\definecolor{pointFive}{HTML}{E69F00}
\definecolor{tblRule}{gray}{0.62}
\providecommand{\onf}{\rowcolor{pointOne!8}}
\providecommand{\fivef}{\rowcolor{pointFive!10}}
\providecommand{\oneinput}[1]{\textcolor{pointOne}{\bfseries #1}}
\providecommand{\fiveinput}[1]{\textcolor{pointFive}{\bfseries #1}}
\providecommand{\onebest}[1]{\textcolor{pointOne}{\bfseries #1}}
\providecommand{\fivebest}[1]{\textcolor{pointFive}{\bfseries #1}}
\providecommand{\mh}[2]{\multicolumn{#1}{c|}{\cellcolor{tblHead}\textbf{#2}}}
\providecommand{\mhl}[1]{\multicolumn{4}{c}{\cellcolor{tblHead}\textbf{#1}}}
\providecommand{\rng}[1]{\shortstack[c]{\scriptsize #1}}
\captionof{table}{All-object detection (static + dynamic). Best AP ($\uparrow$) and SpdE ($\downarrow$) values are bold within each frame setting; one-frame configurations are green and five-frame configurations are orange.}
\label{tab:all-metric-ablation}
\arrayrulecolor{tblRule}
\resizebox{\linewidth}{!}{%
\begin{tabular}{@{}ll|
  *{4}{S[table-format=1.2]}|
  *{4}{S[table-format=2.2]}|
  *{4}{S[table-format=1.2]}|
  *{4}{S[table-format=1.2]}
  @{}}
\arrayrulecolor{black}
\toprule
\arrayrulecolor{tblRule}
\rowcolor{tblHead}
& & \mh{8}{Vehicle} & \multicolumn{8}{c}{\cellcolor{tblHead}\textbf{Pedestrian}} \\
\rowcolor{tblHead}
& & \mh{4}{AP} & \mh{4}{SpdE (m/s)} & \mh{4}{AP} & \mhl{SpdE (m/s)} \\
\rowcolor{tblHead}
Method & Input
  & {\rng{0--100\\m}} & {\rng{100--200\\m}} & {\rng{200--400\\m}} & {\rng{Full\\Range}}
  & {\rng{0--100\\m}} & {\rng{100--200\\m}} & {\rng{200--400\\m}} & {\rng{Full\\Range}}
  & {\rng{0--100\\m}} & {\rng{100--200\\m}} & {\rng{200--400\\m}} & {\rng{Full\\Range}}
  & {\rng{0--100\\m}} & {\rng{100--200\\m}} & {\rng{200--400\\m}} & {\rng{Full\\Range}} \\
\midrule
\onf \textbf{Classical} & \oneinput{1-frame}
  & 0.23 & 0.39 & 0.33 & 0.27
  & 0.90 & \onebest{1.07} & \onebest{2.61} & 1.15
  & {--} & {--} & {--} & {--}
  & {--} & {--} & {--} & {--} \\
\midrule
\onf \textbf{CenterPoint} & \oneinput{1-frame}
  & 0.77 & 0.60 & 0.34 & 0.71
  & 3.86 & 8.89 & 12.41 & 5.35
  & 0.53 & 0.12 & 0.00 & 0.50
  & 0.73 & 0.99 & {--} & 0.74 \\
\onf & \oneinput{\quad+vel}
  & 0.76 & 0.68 & 0.58 & 0.74
  & 1.77 & 2.85 & 4.34 & 2.18
  & 0.65 & 0.37 & 0.01 & 0.63
  & 0.76 & 0.99 & {--} & 0.77 \\
\fivef & \fiveinput{5-frame}
  & 0.79 & 0.63 & 0.39 & 0.74
  & 0.78 & 1.60 & 3.81 & 1.15
  & 0.75 & 0.46 & 0.01 & 0.72
  & 0.41 & 0.56 & 2.73 & 0.42 \\
\fivef & \fiveinput{\quad+vel}
  & 0.80 & 0.66 & 0.45 & 0.75
  & 0.73 & 1.47 & 3.62 & 1.08
  & 0.72 & 0.53 & 0.13 & 0.71
  & 0.48 & 0.59 & 1.33 & 0.48 \\
\midrule
\onf \textbf{DSVT} & \oneinput{1-frame}
  & 0.76 & 0.64 & 0.32 & 0.72
  & 3.94 & 8.84 & 11.90 & 5.41
  & 0.56 & 0.26 & 0.00 & 0.54
  & 0.77 & 1.02 & 3.67 & 0.79 \\
\onf & \oneinput{\quad+vel}
  & 0.78 & 0.70 & 0.56 & 0.75
  & 2.19 & 4.89 & 8.26 & 3.13
  & 0.57 & 0.26 & 0.02 & 0.54
  & 0.73 & 1.08 & 2.82 & 0.74 \\
\fivef & \fiveinput{5-frame}
  & 0.75 & 0.66 & 0.52 & 0.72
  & 0.92 & 1.81 & 4.50 & 1.34
  & 0.57 & 0.35 & 0.02 & 0.56
  & 0.35 & 0.50 & \fivebest{0.75} & 0.36 \\
\fivef & \fiveinput{\quad+vel}
  & 0.80 & 0.72 & \fivebest{0.57} & 0.78
  & 0.98 & 2.34 & 6.31 & 1.63
  & 0.59 & 0.38 & 0.00 & 0.57
  & 0.40 & 0.69 & 1.61 & 0.42 \\
\midrule
\onf \textbf{VoxelNext} & \oneinput{1-frame}
  & 0.84 & 0.71 & 0.39 & 0.80
  & 1.68 & 4.77 & 9.22 & 2.79
  & 0.77 & 0.46 & 0.00 & 0.74
  & 0.56 & 0.81 & 4.89 & 0.58 \\
\onf & \oneinput{\quad+vel}
  & 0.86 & 0.76 & 0.60 & 0.83
  & \onebest{0.79} & 1.92 & 4.15 & 1.23
  & \onebest{0.79} & 0.58 & 0.06 & \onebest{0.77}
  & 0.47 & 0.60 & 1.65 & 0.48 \\
\fivef & \fiveinput{5-frame}
  & 0.83 & 0.70 & 0.48 & 0.79
  & 0.65 & 1.47 & 4.08 & 1.05
  & 0.80 & 0.54 & 0.23 & 0.79
  & \fivebest{0.28} & 0.42 & 0.89 & \fivebest{0.29} \\
\fivef & \fiveinput{\quad+vel}
  & \fivebest{0.86} & \fivebest{0.74} & 0.53 & \fivebest{0.82}
  & \fivebest{0.61} & \fivebest{1.37} & 3.63 & \fivebest{0.97}
  & \fivebest{0.81} & \fivebest{0.61} & 0.29 & \fivebest{0.80}
  & \fivebest{0.28} & \fivebest{0.41} & 0.80 & \fivebest{0.29} \\
\midrule
\onf \textbf{TransFusion} & \oneinput{1-frame}
  & 0.86 & 0.75 & 0.47 & 0.82
  & 2.04 & 4.45 & 7.54 & 2.81
  & 0.53 & 0.41 & 0.00 & 0.52
  & 0.79 & 1.01 & {--} & 0.80 \\
\onf & \oneinput{\quad+vel}
  & \onebest{0.87} & \onebest{0.79} & \onebest{0.67} & \onebest{0.84}
  & 0.86 & 1.53 & 3.25 & \onebest{1.14}
  & 0.56 & \onebest{0.59} & \onebest{0.15} & 0.56
  & \onebest{0.39} & \onebest{0.47} & \onebest{1.12} & \onebest{0.40} \\
\fivef & \fiveinput{5-frame}
  & 0.84 & 0.73 & 0.53 & 0.80
  & 0.79 & 1.53 & \fivebest{3.34} & 1.10
  & 0.54 & 0.50 & 0.13 & 0.54
  & 0.40 & 0.56 & 2.19 & 0.41 \\
\fivef & \fiveinput{\quad+vel}
  & 0.84 & \fivebest{0.74} & \fivebest{0.57} & 0.80
  & 0.77 & 1.73 & 3.66 & 1.18
  & 0.54 & 0.50 & \fivebest{0.33} & 0.53
  & 0.53 & 0.68 & 1.74 & 0.54 \\
\arrayrulecolor{black}
\bottomrule
\end{tabular}}
\arrayrulecolor{black}
\end{minipage}

%% file: tables/point_binned_dynamic_plot.tex
\definecolor{figNineGreen}{HTML}{009E73}
\definecolor{figNineOrange}{HTML}{E69F00}
\definecolor{figNineBlack}{HTML}{222222}
\begin{tikzpicture}
\begin{groupplot}[
group style={group size=4 by 4, horizontal sep=0.004\textwidth, vertical sep=0.022\textwidth, x descriptions at=edge bottom},
width=0.247\textwidth, height=0.180\textwidth,
xmin=-0.25, xmax=6.25, ymin=-2, ymax=102,
xtick={0,1,2,3,4,5,6},
xticklabels={$500{+}$,100--500,50--100,20--50,10--20,5--10,1--5},
xticklabel style={font=\footnotesize, rotate=35, anchor=north east},
ytick={0,25,50,75,100},
tick label style={font=\footnotesize},
label style={font=\footnotesize}, title style={font=\footnotesize\bfseries},
grid=major, grid style={dotted, gray!55}, axis lines*=left,
axis line style={line width=0.55pt}, every axis plot/.append style={line width=0.55pt},
every mark/.append style={solid},
xlabel={Points per box},
]
\nextgroupplot[title={CenterPoint}, ylabel={\shortstack{0--100\,m\\AP (\%)}}]
\addplot+[figNineBlack, solid, mark=diamond*, mark size=1.7pt] coordinates {(0,61.1) (1,55.9) (2,23.9) (3,12.7) (4,3.3) (5,0.6)};
\addplot+[figNineGreen, dashed, mark=o, mark size=1.7pt] coordinates {(0,66.1) (1,55.0) (2,28.0) (3,16.8) (4,4.9) (5,1.1)};
\addplot+[figNineGreen, solid, mark=*, mark size=1.7pt] coordinates {(0,79.8) (1,81.3) (2,68.5) (3,52.9) (4,26.4) (5,9.4)};
\addplot+[figNineOrange, dashed, mark=square, mark size=1.5pt] coordinates {(0,82.4) (1,79.3) (2,58.9) (3,40.7) (4,15.6) (5,5.9) (6,1.5)};
\addplot+[figNineOrange, solid, mark=square*, mark size=1.5pt] coordinates {(0,82.8) (1,82.5) (2,70.3) (3,55.5) (4,27.4) (5,12.2) (6,3.7)};
\nextgroupplot[title={DSVT}]
\addplot+[figNineBlack, solid, mark=diamond*, mark size=1.7pt] coordinates {(0,61.1) (1,55.9) (2,23.9) (3,12.7) (4,3.3) (5,0.6)};
\addplot+[figNineGreen, dashed, mark=o, mark size=1.7pt] coordinates {(0,62.5) (1,56.0) (2,27.3) (3,15.3) (4,3.9) (5,0.8)};
\addplot+[figNineGreen, solid, mark=*, mark size=1.7pt] coordinates {(0,72.5) (1,77.3) (2,68.0) (3,51.8) (4,24.0) (5,8.2)};
\addplot+[figNineOrange, dashed, mark=square, mark size=1.5pt] coordinates {(0,86.2) (1,83.6) (2,63.0) (3,43.5) (4,16.7) (5,5.2) (6,1.3)};
\addplot+[figNineOrange, solid, mark=square*, mark size=1.5pt] coordinates {(0,86.1) (1,85.4) (2,74.2) (3,56.3) (4,27.4) (5,12.2) (6,3.5)};
\nextgroupplot[title={VoxelNeXt}]
\addplot+[figNineBlack, solid, mark=diamond*, mark size=1.7pt] coordinates {(0,61.1) (1,55.9) (2,23.9) (3,12.7) (4,3.3) (5,0.6)};
\addplot+[figNineGreen, dashed, mark=o, mark size=1.7pt] coordinates {(0,70.3) (1,72.9) (2,51.3) (3,34.0) (4,11.7) (5,3.0)};
\addplot+[figNineGreen, solid, mark=*, mark size=1.7pt] coordinates {(0,87.0) (1,84.1) (2,65.5) (3,49.9) (4,23.4) (5,8.7)};
\addplot+[figNineOrange, dashed, mark=square, mark size=1.5pt] coordinates {(0,88.2) (1,83.4) (2,60.6) (3,44.0) (4,17.7) (5,6.6) (6,1.9)};
\addplot+[figNineOrange, solid, mark=square*, mark size=1.5pt] coordinates {(0,89.6) (1,85.1) (2,64.8) (3,48.2) (4,20.5) (5,8.0) (6,2.3)};
\nextgroupplot[title={TransFusion}]
\addplot+[figNineBlack, solid, mark=diamond*, mark size=1.7pt] coordinates {(0,61.1) (1,55.9) (2,23.9) (3,12.7) (4,3.3) (5,0.6)};
\addplot+[figNineGreen, dashed, mark=o, mark size=1.7pt] coordinates {(0,61.4) (1,67.0) (2,48.9) (3,35.9) (4,15.2) (5,5.0)};
\addplot+[figNineGreen, solid, mark=*, mark size=1.7pt] coordinates {(0,84.9) (1,83.9) (2,65.8) (3,52.7) (4,25.7) (5,9.0)};
\addplot+[figNineOrange, dashed, mark=square, mark size=1.5pt] coordinates {(0,83.8) (1,81.0) (2,61.1) (3,45.7) (4,18.7) (5,7.1) (6,1.7)};
\addplot+[figNineOrange, solid, mark=square*, mark size=1.5pt] coordinates {(0,83.8) (1,85.5) (2,73.9) (3,60.7) (4,32.6) (5,14.6) (6,3.7)};
\nextgroupplot[ylabel={\shortstack{100--200\,m\\AP (\%)}}]
\addplot+[figNineBlack, solid, mark=diamond*, mark size=1.7pt] coordinates {(0,8.8) (1,40.8) (2,57.5) (3,55.6) (4,26.9) (5,8.3)};
\addplot+[figNineGreen, dashed, mark=o, mark size=1.7pt] coordinates {(0,5.2) (1,54.0) (2,66.8) (3,62.4) (4,28.0) (5,4.6)};
\addplot+[figNineGreen, solid, mark=*, mark size=1.7pt] coordinates {(0,22.8) (1,67.8) (2,80.8) (3,81.1) (4,61.6) (5,29.8)};
\addplot+[figNineOrange, dashed, mark=square, mark size=1.5pt] coordinates {(0,21.8) (1,64.4) (2,75.6) (3,74.1) (4,50.7) (5,21.6) (6,3.0)};
\addplot+[figNineOrange, solid, mark=square*, mark size=1.5pt] coordinates {(0,33.6) (1,66.0) (2,77.4) (3,76.3) (4,55.9) (5,27.4) (6,5.0)};
\nextgroupplot[]
\addplot+[figNineBlack, solid, mark=diamond*, mark size=1.7pt] coordinates {(0,8.8) (1,40.8) (2,57.5) (3,55.6) (4,26.9) (5,8.3)};
\addplot+[figNineGreen, dashed, mark=o, mark size=1.7pt] coordinates {(0,4.8) (1,54.3) (2,67.8) (3,61.4) (4,25.7) (5,3.6)};
\addplot+[figNineGreen, solid, mark=*, mark size=1.7pt] coordinates {(0,25.3) (1,63.0) (2,79.5) (3,77.7) (4,58.8) (5,27.0)};
\addplot+[figNineOrange, dashed, mark=square, mark size=1.5pt] coordinates {(0,23.4) (1,71.4) (2,81.6) (3,79.4) (4,57.3) (5,24.2) (6,3.5)};
\addplot+[figNineOrange, solid, mark=square*, mark size=1.5pt] coordinates {(0,26.9) (1,73.0) (2,83.6) (3,80.9) (4,61.1) (5,31.5) (6,5.9)};
\nextgroupplot[]
\addplot+[figNineBlack, solid, mark=diamond*, mark size=1.7pt] coordinates {(0,8.8) (1,40.8) (2,57.5) (3,55.6) (4,26.9) (5,8.3)};
\addplot+[figNineGreen, dashed, mark=o, mark size=1.7pt] coordinates {(0,13.2) (1,60.3) (2,74.3) (3,71.8) (4,42.2) (5,9.7)};
\addplot+[figNineGreen, solid, mark=*, mark size=1.7pt] coordinates {(0,32.1) (1,74.1) (2,82.3) (3,81.6) (4,60.8) (5,28.6)};
\addplot+[figNineOrange, dashed, mark=square, mark size=1.5pt] coordinates {(0,27.2) (1,73.5) (2,80.0) (3,77.6) (4,54.1) (5,23.5) (6,3.9)};
\addplot+[figNineOrange, solid, mark=square*, mark size=1.5pt] coordinates {(0,34.0) (1,74.9) (2,80.8) (3,79.4) (4,57.2) (5,28.1) (6,5.4)};
\nextgroupplot[]
\addplot+[figNineBlack, solid, mark=diamond*, mark size=1.7pt] coordinates {(0,8.8) (1,40.8) (2,57.5) (3,55.6) (4,26.9) (5,8.3)};
\addplot+[figNineGreen, dashed, mark=o, mark size=1.7pt] coordinates {(0,21.0) (1,55.6) (2,73.2) (3,74.7) (4,54.2) (5,17.9)};
\addplot+[figNineGreen, solid, mark=*, mark size=1.7pt] coordinates {(0,28.0) (1,69.1) (2,80.1) (3,82.9) (4,66.1) (5,30.3)};
\addplot+[figNineOrange, dashed, mark=square, mark size=1.5pt] coordinates {(0,27.3) (1,66.9) (2,77.9) (3,78.7) (4,59.4) (5,28.1) (6,4.0)};
\addplot+[figNineOrange, solid, mark=square*, mark size=1.5pt] coordinates {(0,29.8) (1,65.6) (2,79.0) (3,81.4) (4,64.1) (5,33.8) (6,6.3)};
\nextgroupplot[ylabel={\shortstack{200--400\,m\\AP (\%)}}]
\addplot+[figNineBlack, solid, mark=diamond*, mark size=1.7pt] coordinates {(1,19.0) (2,29.1) (3,34.3) (4,33.3) (5,22.8)};
\addplot+[figNineGreen, dashed, mark=o, mark size=1.7pt] coordinates {(1,6.4) (2,12.2) (3,20.1) (4,27.6) (5,10.7)};
\addplot+[figNineGreen, solid, mark=*, mark size=1.7pt] coordinates {(1,16.4) (2,20.7) (3,44.1) (4,61.9) (5,48.0)};
\addplot+[figNineOrange, dashed, mark=square, mark size=1.5pt] coordinates {(1,51.3) (2,33.2) (3,32.4) (4,43.7) (5,30.8) (6,2.6)};
\addplot+[figNineOrange, solid, mark=square*, mark size=1.5pt] coordinates {(1,44.4) (2,26.9) (3,35.8) (4,50.7) (5,37.5) (6,4.6)};
\nextgroupplot[]
\addplot+[figNineBlack, solid, mark=diamond*, mark size=1.7pt] coordinates {(1,19.0) (2,29.1) (3,34.3) (4,33.3) (5,22.8)};
\addplot+[figNineGreen, dashed, mark=o, mark size=1.7pt] coordinates {(1,8.3) (2,7.3) (3,21.9) (4,24.9) (5,8.2)};
\addplot+[figNineGreen, solid, mark=*, mark size=1.7pt] coordinates {(1,18.8) (2,13.6) (3,41.9) (4,60.0) (5,42.9)};
\addplot+[figNineOrange, dashed, mark=square, mark size=1.5pt] coordinates {(1,57.8) (2,40.1) (3,40.5) (4,57.3) (5,40.9) (6,4.4)};
\addplot+[figNineOrange, solid, mark=square*, mark size=1.5pt] coordinates {(1,51.0) (2,27.6) (3,43.1) (4,58.6) (5,45.0) (6,5.0)};
\nextgroupplot[]
\addplot+[figNineBlack, solid, mark=diamond*, mark size=1.7pt] coordinates {(1,19.0) (2,29.1) (3,34.3) (4,33.3) (5,22.8)};
\addplot+[figNineGreen, dashed, mark=o, mark size=1.7pt] coordinates {(1,22.3) (2,5.2) (3,19.2) (4,32.5) (5,15.7)};
\addplot+[figNineGreen, solid, mark=*, mark size=1.7pt] coordinates {(1,44.6) (2,20.5) (3,38.5) (4,61.1) (5,49.1)};
\addplot+[figNineOrange, dashed, mark=square, mark size=1.5pt] coordinates {(1,48.8) (2,41.6) (3,39.3) (4,48.0) (5,33.7) (6,4.0)};
\addplot+[figNineOrange, solid, mark=square*, mark size=1.5pt] coordinates {(1,45.6) (2,34.3) (3,33.8) (4,47.8) (5,37.8) (6,5.4)};
\nextgroupplot[]
\addplot+[figNineBlack, solid, mark=diamond*, mark size=1.7pt] coordinates {(1,19.0) (2,29.1) (3,34.3) (4,33.3) (5,22.8)};
\addplot+[figNineGreen, dashed, mark=o, mark size=1.7pt] coordinates {(1,18.3) (2,12.3) (3,35.7) (4,47.0) (5,24.4)};
\addplot+[figNineGreen, solid, mark=*, mark size=1.7pt] coordinates {(1,30.6) (2,29.2) (3,48.2) (4,69.1) (5,51.5)};
\addplot+[figNineOrange, dashed, mark=square, mark size=1.5pt] coordinates {(1,53.3) (2,36.1) (3,43.4) (4,53.3) (5,38.0) (6,3.7)};
\addplot+[figNineOrange, solid, mark=square*, mark size=1.5pt] coordinates {(1,31.3) (2,19.7) (3,41.2) (4,58.2) (5,45.2) (6,6.1)};
\nextgroupplot[ylabel={\shortstack{Full range\\AP (\%)}}]
\addplot+[figNineBlack, solid, mark=diamond*, mark size=1.7pt] coordinates {(0,56.1) (1,52.9) (2,35.0) (3,32.1) (4,16.6) (5,8.4)};
\addplot+[figNineGreen, dashed, mark=o, mark size=1.7pt] coordinates {(0,64.2) (1,54.0) (2,38.4) (3,36.6) (4,17.2) (5,5.1)};
\addplot+[figNineGreen, solid, mark=*, mark size=1.7pt] coordinates {(0,78.2) (1,79.0) (2,72.3) (3,69.9) (4,53.9) (5,32.9)};
\addplot+[figNineOrange, dashed, mark=square, mark size=1.5pt] coordinates {(0,80.4) (1,76.4) (2,65.7) (3,60.6) (4,38.6) (5,19.9) (6,2.3)};
\addplot+[figNineOrange, solid, mark=square*, mark size=1.5pt] coordinates {(0,80.9) (1,80.1) (2,72.9) (3,68.2) (4,48.0) (5,27.3) (6,4.6)};
\nextgroupplot[]
\addplot+[figNineBlack, solid, mark=diamond*, mark size=1.7pt] coordinates {(0,56.1) (1,52.9) (2,35.0) (3,32.1) (4,16.6) (5,8.4)};
\addplot+[figNineGreen, dashed, mark=o, mark size=1.7pt] coordinates {(0,61.0) (1,55.2) (2,38.4) (3,35.5) (4,15.0) (5,3.7)};
\addplot+[figNineGreen, solid, mark=*, mark size=1.7pt] coordinates {(0,71.2) (1,74.9) (2,71.8) (3,67.9) (4,50.4) (5,28.3)};
\addplot+[figNineOrange, dashed, mark=square, mark size=1.5pt] coordinates {(0,84.5) (1,81.6) (2,70.3) (3,65.3) (4,44.3) (5,22.5) (6,2.8)};
\addplot+[figNineOrange, solid, mark=square*, mark size=1.5pt] coordinates {(0,84.4) (1,83.4) (2,77.4) (3,71.4) (4,52.3) (5,32.0) (6,5.0)};
\nextgroupplot[]
\addplot+[figNineBlack, solid, mark=diamond*, mark size=1.7pt] coordinates {(0,56.1) (1,52.9) (2,35.0) (3,32.1) (4,16.6) (5,8.4)};
\addplot+[figNineGreen, dashed, mark=o, mark size=1.7pt] coordinates {(0,68.8) (1,70.5) (2,59.2) (3,54.2) (4,28.8) (5,8.8)};
\addplot+[figNineGreen, solid, mark=*, mark size=1.7pt] coordinates {(0,86.3) (1,82.3) (2,71.2) (3,68.4) (4,49.3) (5,28.7)};
\addplot+[figNineOrange, dashed, mark=square, mark size=1.5pt] coordinates {(0,87.0) (1,81.4) (2,68.3) (3,63.9) (4,41.5) (5,20.7) (6,3.2)};
\addplot+[figNineOrange, solid, mark=square*, mark size=1.5pt] coordinates {(0,88.2) (1,83.5) (2,71.0) (3,65.7) (4,43.1) (5,23.1) (6,3.9)};
\nextgroupplot[]
\addplot+[figNineBlack, solid, mark=diamond*, mark size=1.7pt] coordinates {(0,56.1) (1,52.9) (2,35.0) (3,32.1) (4,16.6) (5,8.4)};
\addplot+[figNineGreen, dashed, mark=o, mark size=1.7pt] coordinates {(0,59.6) (1,64.5) (2,57.3) (3,59.0) (4,40.4) (5,15.8)};
\addplot+[figNineGreen, solid, mark=*, mark size=1.7pt] coordinates {(0,82.8) (1,81.0) (2,71.0) (3,71.0) (4,56.1) (5,31.8)};
\addplot+[figNineOrange, dashed, mark=square, mark size=1.5pt] coordinates {(0,81.4) (1,78.2) (2,67.7) (3,66.0) (4,46.9) (5,25.5) (6,3.1)};
\addplot+[figNineOrange, solid, mark=square*, mark size=1.5pt] coordinates {(0,81.2) (1,81.6) (2,74.8) (3,73.4) (4,56.3) (5,34.1) (6,5.6)};
\end{groupplot}
\end{tikzpicture}

%% file: tables/point_binned_all_plot.tex
\definecolor{figNineGreen}{HTML}{009E73}
\definecolor{figNineOrange}{HTML}{E69F00}
\definecolor{figNineBlack}{HTML}{222222}
\begin{tikzpicture}
\begin{groupplot}[
group style={group size=4 by 4, horizontal sep=0.004\textwidth, vertical sep=0.022\textwidth, x descriptions at=edge bottom},
width=0.247\textwidth, height=0.180\textwidth,
xmin=-0.25, xmax=6.25, ymin=-2, ymax=102,
xtick={0,1,2,3,4,5,6},
xticklabels={$500{+}$,100--500,50--100,20--50,10--20,5--10,1--5},
xticklabel style={font=\footnotesize, rotate=35, anchor=north east},
ytick={0,25,50,75,100},
tick label style={font=\footnotesize},
label style={font=\footnotesize}, title style={font=\footnotesize\bfseries},
grid=major, grid style={dotted, gray!55}, axis lines*=left,
axis line style={line width=0.55pt}, every axis plot/.append style={line width=0.55pt},
every mark/.append style={solid},
xlabel={Points per box},
]
\nextgroupplot[title={CenterPoint}, ylabel={\shortstack{0--100\,m\\AP (\%)}}]
\addplot+[figNineBlack, solid, mark=diamond*, mark size=1.7pt] coordinates {(0,20.7) (1,22.5) (2,9.9) (3,4.7) (4,1.2) (5,0.3)};
\addplot+[figNineGreen, dashed, mark=o, mark size=1.7pt] coordinates {(0,89.8) (1,79.2) (2,51.9) (3,28.9) (4,6.2) (5,0.9)};
\addplot+[figNineGreen, solid, mark=*, mark size=1.7pt] coordinates {(0,88.8) (1,77.9) (2,51.3) (3,30.1) (4,10.2) (5,3.0)};
\addplot+[figNineOrange, dashed, mark=square, mark size=1.5pt] coordinates {(0,88.6) (1,77.1) (2,47.1) (3,30.0) (4,9.8) (5,3.4) (6,0.9)};
\addplot+[figNineOrange, solid, mark=square*, mark size=1.5pt] coordinates {(0,88.8) (1,77.8) (2,49.6) (3,33.5) (4,12.1) (5,4.5) (6,1.2)};
\nextgroupplot[title={DSVT}]
\addplot+[figNineBlack, solid, mark=diamond*, mark size=1.7pt] coordinates {(0,20.7) (1,22.5) (2,9.9) (3,4.7) (4,1.2) (5,0.3)};
\addplot+[figNineGreen, dashed, mark=o, mark size=1.7pt] coordinates {(0,90.1) (1,78.7) (2,49.9) (3,24.5) (4,4.5) (5,0.6)};
\addplot+[figNineGreen, solid, mark=*, mark size=1.7pt] coordinates {(0,90.3) (1,79.6) (2,49.7) (3,25.8) (4,6.0) (5,1.5)};
\addplot+[figNineOrange, dashed, mark=square, mark size=1.5pt] coordinates {(0,89.5) (1,78.0) (2,51.1) (3,28.1) (4,7.8) (5,2.1) (6,0.4)};
\addplot+[figNineOrange, solid, mark=square*, mark size=1.5pt] coordinates {(0,91.4) (1,81.7) (2,59.8) (3,38.8) (4,13.6) (5,5.2) (6,1.3)};
\nextgroupplot[title={VoxelNeXt}]
\addplot+[figNineBlack, solid, mark=diamond*, mark size=1.7pt] coordinates {(0,20.7) (1,22.5) (2,9.9) (3,4.7) (4,1.2) (5,0.3)};
\addplot+[figNineGreen, dashed, mark=o, mark size=1.7pt] coordinates {(0,91.5) (1,83.1) (2,57.7) (3,36.5) (4,10.8) (5,2.4)};
\addplot+[figNineGreen, solid, mark=*, mark size=1.7pt] coordinates {(0,92.4) (1,84.9) (2,59.3) (3,38.0) (4,11.9) (5,3.1)};
\addplot+[figNineOrange, dashed, mark=square, mark size=1.5pt] coordinates {(0,91.8) (1,82.9) (2,55.2) (3,37.3) (4,12.6) (5,4.1) (6,1.2)};
\addplot+[figNineOrange, solid, mark=square*, mark size=1.5pt] coordinates {(0,92.6) (1,84.7) (2,58.3) (3,40.1) (4,13.8) (5,4.8) (6,1.4)};
\nextgroupplot[title={TransFusion}]
\addplot+[figNineBlack, solid, mark=diamond*, mark size=1.7pt] coordinates {(0,20.7) (1,22.5) (2,9.9) (3,4.7) (4,1.2) (5,0.3)};
\addplot+[figNineGreen, dashed, mark=o, mark size=1.7pt] coordinates {(0,89.4) (1,81.7) (2,53.9) (3,36.2) (4,10.7) (5,2.6)};
\addplot+[figNineGreen, solid, mark=*, mark size=1.7pt] coordinates {(0,89.8) (1,83.5) (2,60.7) (3,44.5) (4,15.9) (5,4.1)};
\addplot+[figNineOrange, dashed, mark=square, mark size=1.5pt] coordinates {(0,88.3) (1,81.6) (2,57.6) (3,41.7) (4,15.2) (5,5.6) (6,1.6)};
\addplot+[figNineOrange, solid, mark=square*, mark size=1.5pt] coordinates {(0,87.3) (1,81.0) (2,58.8) (3,43.2) (4,17.8) (5,6.9) (6,2.1)};
\nextgroupplot[ylabel={\shortstack{100--200\,m\\AP (\%)}}]
\addplot+[figNineBlack, solid, mark=diamond*, mark size=1.7pt] coordinates {(0,4.7) (1,20.9) (2,35.7) (3,40.6) (4,19.4) (5,5.5)};
\addplot+[figNineGreen, dashed, mark=o, mark size=1.7pt] coordinates {(0,22.4) (1,58.6) (2,65.8) (3,61.7) (4,28.7) (5,4.2)};
\addplot+[figNineGreen, solid, mark=*, mark size=1.7pt] coordinates {(0,22.1) (1,59.3) (2,65.5) (3,66.7) (4,46.9) (5,18.8)};
\addplot+[figNineOrange, dashed, mark=square, mark size=1.5pt] coordinates {(0,20.9) (1,59.2) (2,60.0) (3,55.5) (4,29.0) (5,9.8) (6,1.3)};
\addplot+[figNineOrange, solid, mark=square*, mark size=1.5pt] coordinates {(0,29.3) (1,60.3) (2,61.1) (3,58.2) (4,33.9) (5,13.6) (6,2.3)};
\nextgroupplot[]
\addplot+[figNineBlack, solid, mark=diamond*, mark size=1.7pt] coordinates {(0,4.7) (1,20.9) (2,35.7) (3,40.6) (4,19.4) (5,5.5)};
\addplot+[figNineGreen, dashed, mark=o, mark size=1.7pt] coordinates {(0,22.6) (1,64.8) (2,69.7) (3,63.4) (4,25.1) (5,3.0)};
\addplot+[figNineGreen, solid, mark=*, mark size=1.7pt] coordinates {(0,30.4) (1,65.3) (2,69.0) (3,67.0) (4,41.0) (5,13.0)};
\addplot+[figNineOrange, dashed, mark=square, mark size=1.5pt] coordinates {(0,31.6) (1,66.7) (2,68.7) (3,66.6) (4,41.7) (5,14.1) (6,1.6)};
\addplot+[figNineOrange, solid, mark=square*, mark size=1.5pt] coordinates {(0,37.6) (1,71.4) (2,74.2) (3,71.3) (4,46.6) (5,20.5) (6,3.7)};
\nextgroupplot[]
\addplot+[figNineBlack, solid, mark=diamond*, mark size=1.7pt] coordinates {(0,4.7) (1,20.9) (2,35.7) (3,40.6) (4,19.4) (5,5.5)};
\addplot+[figNineGreen, dashed, mark=o, mark size=1.7pt] coordinates {(0,24.6) (1,68.0) (2,71.2) (3,65.5) (4,31.1) (5,6.6)};
\addplot+[figNineGreen, solid, mark=*, mark size=1.7pt] coordinates {(0,32.0) (1,72.3) (2,73.7) (3,69.4) (4,38.9) (5,12.3)};
\addplot+[figNineOrange, dashed, mark=square, mark size=1.5pt] coordinates {(0,33.3) (1,69.4) (2,68.2) (3,66.0) (4,38.6) (5,14.0) (6,2.1)};
\addplot+[figNineOrange, solid, mark=square*, mark size=1.5pt] coordinates {(0,39.3) (1,73.6) (2,70.9) (3,66.3) (4,40.0) (5,16.9) (6,3.1)};
\nextgroupplot[]
\addplot+[figNineBlack, solid, mark=diamond*, mark size=1.7pt] coordinates {(0,4.7) (1,20.9) (2,35.7) (3,40.6) (4,19.4) (5,5.5)};
\addplot+[figNineGreen, dashed, mark=o, mark size=1.7pt] coordinates {(0,36.3) (1,67.0) (2,68.5) (3,68.7) (4,39.8) (5,8.6)};
\addplot+[figNineGreen, solid, mark=*, mark size=1.7pt] coordinates {(0,28.9) (1,68.4) (2,72.0) (3,74.5) (4,53.7) (5,16.1)};
\addplot+[figNineOrange, dashed, mark=square, mark size=1.5pt] coordinates {(0,26.7) (1,61.8) (2,65.7) (3,69.6) (4,47.4) (5,19.0) (6,2.7)};
\addplot+[figNineOrange, solid, mark=square*, mark size=1.5pt] coordinates {(0,29.6) (1,59.6) (2,67.1) (3,71.7) (4,51.1) (5,23.4) (6,4.1)};
\nextgroupplot[ylabel={\shortstack{200--400\,m\\AP (\%)}}]
\addplot+[figNineBlack, solid, mark=diamond*, mark size=1.7pt] coordinates {(1,9.2) (2,20.8) (3,22.4) (4,27.6) (5,21.1)};
\addplot+[figNineGreen, dashed, mark=o, mark size=1.7pt] coordinates {(1,8.5) (2,15.1) (3,20.4) (4,28.1) (5,11.7)};
\addplot+[figNineGreen, solid, mark=*, mark size=1.7pt] coordinates {(1,10.2) (2,17.8) (3,32.8) (4,54.2) (5,44.8)};
\addplot+[figNineOrange, dashed, mark=square, mark size=1.5pt] coordinates {(1,20.7) (2,18.0) (3,20.4) (4,29.4) (5,21.7) (6,1.7)};
\addplot+[figNineOrange, solid, mark=square*, mark size=1.5pt] coordinates {(1,19.2) (2,17.9) (3,21.6) (4,36.1) (5,28.6) (6,3.5)};
\nextgroupplot[]
\addplot+[figNineBlack, solid, mark=diamond*, mark size=1.7pt] coordinates {(1,9.2) (2,20.8) (3,22.4) (4,27.6) (5,21.1)};
\addplot+[figNineGreen, dashed, mark=o, mark size=1.7pt] coordinates {(1,12.4) (2,12.2) (3,24.0) (4,26.5) (5,8.4)};
\addplot+[figNineGreen, solid, mark=*, mark size=1.7pt] coordinates {(1,14.3) (2,14.6) (3,32.3) (4,52.4) (5,37.3)};
\addplot+[figNineOrange, dashed, mark=square, mark size=1.5pt] coordinates {(1,34.4) (2,34.5) (3,32.0) (4,49.3) (5,35.6) (6,3.5)};
\addplot+[figNineOrange, solid, mark=square*, mark size=1.5pt] coordinates {(1,33.3) (2,26.6) (3,34.8) (4,50.9) (5,41.4) (6,4.3)};
\nextgroupplot[]
\addplot+[figNineBlack, solid, mark=diamond*, mark size=1.7pt] coordinates {(1,9.2) (2,20.8) (3,22.4) (4,27.6) (5,21.1)};
\addplot+[figNineGreen, dashed, mark=o, mark size=1.7pt] coordinates {(1,19.2) (2,10.1) (3,19.3) (4,31.0) (5,14.3)};
\addplot+[figNineGreen, solid, mark=*, mark size=1.7pt] coordinates {(1,28.6) (2,17.7) (3,30.7) (4,51.7) (5,39.9)};
\addplot+[figNineOrange, dashed, mark=square, mark size=1.5pt] coordinates {(1,40.0) (2,38.8) (3,27.8) (4,37.5) (5,24.7) (6,2.5)};
\addplot+[figNineOrange, solid, mark=square*, mark size=1.5pt] coordinates {(1,36.9) (2,33.0) (3,27.8) (4,40.0) (5,31.1) (6,4.5)};
\nextgroupplot[]
\addplot+[figNineBlack, solid, mark=diamond*, mark size=1.7pt] coordinates {(1,9.2) (2,20.8) (3,22.4) (4,27.6) (5,21.1)};
\addplot+[figNineGreen, dashed, mark=o, mark size=1.7pt] coordinates {(1,19.5) (2,21.3) (3,31.1) (4,40.5) (5,18.4)};
\addplot+[figNineGreen, solid, mark=*, mark size=1.7pt] coordinates {(1,20.0) (2,25.7) (3,39.8) (4,63.6) (5,46.7)};
\addplot+[figNineOrange, dashed, mark=square, mark size=1.5pt] coordinates {(1,28.5) (2,28.2) (3,33.9) (4,46.9) (5,34.0) (6,3.3)};
\addplot+[figNineOrange, solid, mark=square*, mark size=1.5pt] coordinates {(1,26.0) (2,20.7) (3,32.0) (4,51.7) (5,40.4) (6,5.2)};
\nextgroupplot[ylabel={\shortstack{Full range\\AP (\%)}}]
\addplot+[figNineBlack, solid, mark=diamond*, mark size=1.7pt] coordinates {(0,19.3) (1,22.4) (2,17.1) (3,18.1) (4,10.4) (5,6.0)};
\addplot+[figNineGreen, dashed, mark=o, mark size=1.7pt] coordinates {(0,88.3) (1,75.3) (2,55.5) (3,42.9) (4,17.1) (5,4.2)};
\addplot+[figNineGreen, solid, mark=*, mark size=1.7pt] coordinates {(0,87.1) (1,74.7) (2,55.2) (3,47.4) (4,33.2) (5,19.3)};
\addplot+[figNineOrange, dashed, mark=square, mark size=1.5pt] coordinates {(0,87.4) (1,74.3) (2,50.5) (3,39.8) (4,19.1) (5,8.7) (6,1.2)};
\addplot+[figNineOrange, solid, mark=square*, mark size=1.5pt] coordinates {(0,87.6) (1,75.1) (2,52.9) (3,43.1) (4,23.2) (5,11.9) (6,1.9)};
\nextgroupplot[]
\addplot+[figNineBlack, solid, mark=diamond*, mark size=1.7pt] coordinates {(0,19.3) (1,22.4) (2,17.1) (3,18.1) (4,10.4) (5,6.0)};
\addplot+[figNineGreen, dashed, mark=o, mark size=1.7pt] coordinates {(0,88.8) (1,76.4) (2,56.1) (3,42.4) (4,15.0) (5,2.7)};
\addplot+[figNineGreen, solid, mark=*, mark size=1.7pt] coordinates {(0,89.4) (1,77.3) (2,55.1) (3,44.0) (4,24.3) (5,10.9)};
\addplot+[figNineOrange, dashed, mark=square, mark size=1.5pt] coordinates {(0,88.1) (1,76.0) (2,56.9) (3,46.3) (4,26.1) (5,12.0) (6,1.2)};
\addplot+[figNineOrange, solid, mark=square*, mark size=1.5pt] coordinates {(0,89.8) (1,79.5) (2,64.1) (3,53.4) (4,32.0) (5,18.5) (6,2.8)};
\nextgroupplot[]
\addplot+[figNineBlack, solid, mark=diamond*, mark size=1.7pt] coordinates {(0,19.3) (1,22.4) (2,17.1) (3,18.1) (4,10.4) (5,6.0)};
\addplot+[figNineGreen, dashed, mark=o, mark size=1.7pt] coordinates {(0,90.0) (1,80.5) (2,61.4) (3,48.2) (4,20.4) (5,5.7)};
\addplot+[figNineGreen, solid, mark=*, mark size=1.7pt] coordinates {(0,91.2) (1,83.0) (2,63.4) (3,50.7) (4,24.5) (5,10.3)};
\addplot+[figNineOrange, dashed, mark=square, mark size=1.5pt] coordinates {(0,90.3) (1,80.5) (2,59.3) (3,49.6) (4,25.7) (5,11.2) (6,1.7)};
\addplot+[figNineOrange, solid, mark=square*, mark size=1.5pt] coordinates {(0,91.5) (1,82.9) (2,62.2) (3,50.5) (4,26.0) (5,12.5) (6,2.3)};
\nextgroupplot[]
\addplot+[figNineBlack, solid, mark=diamond*, mark size=1.7pt] coordinates {(0,19.3) (1,22.4) (2,17.1) (3,18.1) (4,10.4) (5,6.0)};
\addplot+[figNineGreen, dashed, mark=o, mark size=1.7pt] coordinates {(0,88.0) (1,79.1) (2,58.4) (3,51.2) (4,24.9) (5,6.7)};
\addplot+[figNineGreen, solid, mark=*, mark size=1.7pt] coordinates {(0,88.2) (1,80.7) (2,64.2) (3,59.5) (4,40.0) (5,15.7)};
\addplot+[figNineOrange, dashed, mark=square, mark size=1.5pt] coordinates {(0,86.2) (1,77.4) (2,59.4) (3,55.3) (4,34.3) (5,16.6) (6,2.3)};
\addplot+[figNineOrange, solid, mark=square*, mark size=1.5pt] coordinates {(0,85.1) (1,76.6) (2,60.1) (3,56.4) (4,38.1) (5,20.9) (6,3.4)};
\end{groupplot}
\end{tikzpicture}

%% file: tables/point_binned_dynamic_table.tex
\begin{minipage}[t]{\textwidth}
\centering
\scriptsize
\setlength{\tabcolsep}{5.4pt}
\renewcommand{\arraystretch}{0.98}
\definecolor{tblHead}{gray}{0.88}
\definecolor{pointOne}{HTML}{009E73}
\definecolor{pointFive}{HTML}{E69F00}
\providecommand{\onebest}[1]{\textcolor{pointOne}{\bfseries #1}}
\providecommand{\fivebest}[1]{\textcolor{pointFive}{\bfseries #1}}
\providecommand{\onehead}{\multicolumn{2}{c}{\cellcolor{pointOne!12}\textcolor{pointOne}{\textbf{1-frame}}}}
\providecommand{\fivehead}{\multicolumn{2}{c|}{\cellcolor{pointFive!14}\textcolor{pointFive}{\textbf{5-frame}}}}
\captionof{table}{Dynamic-object vehicle AP by point count. Per row/input, the best detector is bold (green: one-frame; orange: five-frame). Unmatched predictions are false positives in every bin; cross-bin matches are ignored. One-frame 1--5 entries are removed by ground-truth filtering.}
\label{tab:point-binned-dynamic}
\resizebox{\linewidth}{!}{%
\begin{tabular}{@{}c|c|*{4}{cccc|}@{}}
\toprule
\rowcolor{tblHead}
& \cellcolor{tblHead}\textbf{Classical} & \multicolumn{4}{c|}{\cellcolor{tblHead}\textbf{CenterPoint}} & \multicolumn{4}{c|}{\cellcolor{tblHead}\textbf{DSVT}} & \multicolumn{4}{c|}{\cellcolor{tblHead}\textbf{VoxelNeXt}} & \multicolumn{4}{c|}{\cellcolor{tblHead}\textbf{TransFusion}} \\
\rowcolor{tblHead}
\textbf{Point count} & \textbf{AP} & \onehead & \fivehead & \onehead & \fivehead & \onehead & \fivehead & \onehead & \fivehead \\
\rowcolor{tblHead}
 & & \textit{w/o} & \textit{+vel} & \textit{w/o} & \textit{+vel} & \textit{w/o} & \textit{+vel} & \textit{w/o} & \textit{+vel} & \textit{w/o} & \textit{+vel} & \textit{w/o} & \textit{+vel} & \textit{w/o} & \textit{+vel} & \textit{w/o} & \textit{+vel} \\
\midrule
\multicolumn{18}{l}{\cellcolor{gray!12}\textbf{0--100\,m}} \\
$500{+}$ & 0.611 & 0.661 & 0.798 & 0.824 & 0.828 & 0.625 & 0.725 & 0.862 & 0.861 & \onebest{0.703} & \onebest{0.870} & \fivebest{0.882} & \fivebest{0.896} & 0.614 & 0.849 & 0.838 & 0.838 \\
100--500 & 0.559 & 0.550 & 0.813 & 0.793 & 0.825 & 0.560 & 0.773 & \fivebest{0.836} & 0.854 & \onebest{0.729} & \onebest{0.841} & 0.834 & 0.851 & 0.670 & 0.839 & 0.810 & \fivebest{0.855} \\
50--100 & 0.239 & 0.280 & \onebest{0.685} & 0.589 & 0.703 & 0.273 & 0.680 & \fivebest{0.630} & \fivebest{0.742} & \onebest{0.513} & 0.655 & 0.606 & 0.648 & 0.489 & 0.658 & 0.611 & 0.739 \\
20--50 & 0.127 & 0.168 & \onebest{0.529} & 0.407 & 0.555 & 0.153 & 0.518 & 0.435 & 0.563 & 0.340 & 0.499 & 0.440 & 0.482 & \onebest{0.359} & 0.527 & \fivebest{0.457} & \fivebest{0.607} \\
10--20 & 0.033 & 0.049 & \onebest{0.264} & 0.156 & 0.274 & 0.039 & 0.240 & 0.167 & 0.274 & 0.117 & 0.234 & 0.177 & 0.205 & \onebest{0.152} & 0.257 & \fivebest{0.187} & \fivebest{0.326} \\
5--10 & 0.006 & 0.011 & \onebest{0.094} & 0.059 & 0.122 & 0.008 & 0.082 & 0.052 & 0.122 & 0.030 & 0.087 & 0.066 & 0.080 & \onebest{0.050} & 0.090 & \fivebest{0.071} & \fivebest{0.146} \\
1--5 & -- & -- & -- & 0.015 & \fivebest{0.037} & -- & -- & 0.013 & 0.035 & -- & -- & \fivebest{0.019} & 0.023 & -- & -- & 0.017 & \fivebest{0.037} \\
\midrule
\multicolumn{18}{l}{\cellcolor{gray!12}\textbf{100--200\,m}} \\
$500{+}$ & 0.088 & 0.052 & 0.228 & 0.218 & 0.336 & 0.048 & 0.253 & 0.234 & 0.269 & 0.132 & \onebest{0.321} & 0.272 & \fivebest{0.340} & \onebest{0.210} & 0.280 & \fivebest{0.273} & 0.298 \\
100--500 & 0.408 & 0.540 & 0.678 & 0.644 & 0.660 & 0.543 & 0.630 & 0.714 & 0.730 & \onebest{0.603} & \onebest{0.741} & \fivebest{0.735} & \fivebest{0.749} & 0.556 & 0.691 & 0.669 & 0.656 \\
50--100 & 0.575 & 0.668 & 0.808 & 0.756 & 0.774 & 0.678 & 0.795 & \fivebest{0.816} & \fivebest{0.836} & \onebest{0.743} & \onebest{0.823} & 0.800 & 0.808 & 0.732 & 0.801 & 0.779 & 0.790 \\
20--50 & 0.556 & 0.624 & 0.811 & 0.741 & 0.763 & 0.614 & 0.777 & \fivebest{0.794} & 0.809 & 0.718 & 0.816 & 0.776 & 0.794 & \onebest{0.747} & \onebest{0.829} & 0.787 & \fivebest{0.814} \\
10--20 & 0.269 & 0.280 & 0.616 & 0.507 & 0.559 & 0.257 & 0.588 & 0.573 & 0.611 & 0.422 & 0.608 & 0.541 & 0.572 & \onebest{0.542} & \onebest{0.661} & \fivebest{0.594} & \fivebest{0.641} \\
5--10 & 0.083 & 0.046 & 0.298 & 0.216 & 0.274 & 0.036 & 0.270 & 0.242 & 0.315 & 0.097 & 0.286 & 0.235 & 0.281 & \onebest{0.179} & \onebest{0.303} & \fivebest{0.281} & \fivebest{0.338} \\
1--5 & -- & -- & -- & 0.030 & 0.050 & -- & -- & 0.035 & 0.059 & -- & -- & 0.039 & 0.054 & -- & -- & \fivebest{0.040} & \fivebest{0.063} \\
\midrule
\multicolumn{18}{l}{\cellcolor{gray!12}\textbf{200--400\,m}} \\
$500{+}$ & -- & -- & -- & -- & -- & -- & -- & -- & -- & -- & -- & -- & -- & -- & -- & -- & -- \\
100--500 & 0.190 & 0.064 & 0.164 & 0.513 & 0.444 & 0.083 & 0.188 & \fivebest{0.578} & \fivebest{0.510} & \onebest{0.223} & \onebest{0.446} & 0.488 & 0.456 & 0.183 & 0.306 & 0.533 & 0.313 \\
50--100 & 0.291 & 0.122 & 0.207 & 0.332 & 0.269 & 0.073 & 0.136 & 0.401 & 0.276 & 0.052 & 0.205 & \fivebest{0.416} & \fivebest{0.343} & \onebest{0.123} & \onebest{0.292} & 0.361 & 0.197 \\
20--50 & 0.343 & 0.201 & 0.441 & 0.324 & 0.358 & 0.219 & 0.419 & 0.405 & \fivebest{0.431} & 0.192 & 0.385 & 0.393 & 0.338 & \onebest{0.357} & \onebest{0.482} & \fivebest{0.434} & 0.412 \\
10--20 & 0.333 & 0.276 & 0.619 & 0.437 & 0.507 & 0.249 & 0.600 & \fivebest{0.573} & \fivebest{0.586} & 0.325 & 0.611 & 0.480 & 0.478 & \onebest{0.470} & \onebest{0.691} & 0.533 & 0.582 \\
5--10 & 0.228 & 0.107 & 0.480 & 0.308 & 0.375 & 0.082 & 0.429 & \fivebest{0.409} & 0.450 & 0.157 & 0.491 & 0.337 & 0.378 & \onebest{0.244} & \onebest{0.515} & 0.380 & \fivebest{0.452} \\
1--5 & -- & -- & -- & 0.026 & 0.046 & -- & -- & \fivebest{0.044} & 0.050 & -- & -- & 0.040 & 0.054 & -- & -- & 0.037 & \fivebest{0.061} \\
\midrule
\multicolumn{18}{l}{\cellcolor{gray!12}\textbf{Full range}} \\
$500{+}$ & 0.561 & 0.642 & 0.782 & 0.804 & 0.809 & 0.610 & 0.712 & 0.845 & 0.844 & \onebest{0.688} & \onebest{0.863} & \fivebest{0.870} & \fivebest{0.882} & 0.596 & 0.828 & 0.814 & 0.812 \\
100--500 & 0.529 & 0.540 & 0.790 & 0.764 & 0.801 & 0.552 & 0.749 & \fivebest{0.816} & 0.834 & \onebest{0.705} & \onebest{0.823} & 0.814 & \fivebest{0.835} & 0.645 & 0.810 & 0.782 & 0.816 \\
50--100 & 0.350 & 0.384 & \onebest{0.723} & 0.657 & 0.729 & 0.384 & 0.718 & \fivebest{0.703} & \fivebest{0.774} & \onebest{0.592} & 0.712 & 0.683 & 0.710 & 0.573 & 0.710 & 0.677 & 0.748 \\
20--50 & 0.321 & 0.366 & 0.699 & 0.606 & 0.682 & 0.355 & 0.679 & 0.653 & 0.714 & 0.542 & 0.684 & 0.639 & 0.657 & \onebest{0.590} & \onebest{0.710} & \fivebest{0.660} & \fivebest{0.734} \\
10--20 & 0.166 & 0.172 & 0.539 & 0.386 & 0.480 & 0.150 & 0.504 & 0.443 & 0.523 & 0.288 & 0.493 & 0.415 & 0.431 & \onebest{0.404} & \onebest{0.561} & \fivebest{0.469} & \fivebest{0.563} \\
5--10 & 0.084 & 0.051 & \onebest{0.329} & 0.199 & 0.273 & 0.037 & 0.283 & 0.225 & 0.320 & 0.088 & 0.287 & 0.207 & 0.231 & \onebest{0.158} & 0.318 & \fivebest{0.255} & \fivebest{0.341} \\
1--5 & -- & -- & -- & 0.023 & 0.046 & -- & -- & 0.028 & 0.050 & -- & -- & \fivebest{0.032} & 0.039 & -- & -- & 0.031 & \fivebest{0.056} \\
\bottomrule
\end{tabular}%
}
\end{minipage}

%% file: tables/point_binned_all_table.tex
\begin{minipage}[t]{\textwidth}
\centering
\scriptsize
\setlength{\tabcolsep}{5.4pt}
\renewcommand{\arraystretch}{0.98}
\definecolor{tblHead}{gray}{0.88}
\definecolor{pointOne}{HTML}{009E73}
\definecolor{pointFive}{HTML}{E69F00}
\providecommand{\onebest}[1]{\textcolor{pointOne}{\bfseries #1}}
\providecommand{\fivebest}[1]{\textcolor{pointFive}{\bfseries #1}}
\providecommand{\onehead}{\multicolumn{2}{c}{\cellcolor{pointOne!12}\textcolor{pointOne}{\textbf{1-frame}}}}
\providecommand{\fivehead}{\multicolumn{2}{c|}{\cellcolor{pointFive!14}\textcolor{pointFive}{\textbf{5-frame}}}}
\captionof{table}{All-object (static + dynamic) vehicle AP by point count. Per row/input, the best detector is bold (green: one-frame; orange: five-frame). Unmatched predictions are false positives in every bin; cross-bin matches are ignored. One-frame 1--5 entries are removed by ground-truth filtering.}
\label{tab:point-binned-all}
\resizebox{\linewidth}{!}{%
\begin{tabular}{@{}c|c|*{4}{cccc|}@{}}
\toprule
\rowcolor{tblHead}
& \cellcolor{tblHead}\textbf{Classical} & \multicolumn{4}{c|}{\cellcolor{tblHead}\textbf{CenterPoint}} & \multicolumn{4}{c|}{\cellcolor{tblHead}\textbf{DSVT}} & \multicolumn{4}{c|}{\cellcolor{tblHead}\textbf{VoxelNeXt}} & \multicolumn{4}{c|}{\cellcolor{tblHead}\textbf{TransFusion}} \\
\rowcolor{tblHead}
\textbf{Point count} & \textbf{AP} & \onehead & \fivehead & \onehead & \fivehead & \onehead & \fivehead & \onehead & \fivehead \\
\rowcolor{tblHead}
 & & \textit{w/o} & \textit{+vel} & \textit{w/o} & \textit{+vel} & \textit{w/o} & \textit{+vel} & \textit{w/o} & \textit{+vel} & \textit{w/o} & \textit{+vel} & \textit{w/o} & \textit{+vel} & \textit{w/o} & \textit{+vel} & \textit{w/o} & \textit{+vel} \\
\midrule
\multicolumn{18}{l}{\cellcolor{gray!12}\textbf{0--100\,m}} \\
$500{+}$ & 0.207 & 0.898 & 0.888 & 0.886 & 0.888 & 0.901 & 0.903 & 0.895 & 0.914 & \onebest{0.915} & \onebest{0.924} & \fivebest{0.918} & \fivebest{0.926} & 0.894 & 0.898 & 0.883 & 0.873 \\
100--500 & 0.225 & 0.792 & 0.779 & 0.771 & 0.778 & 0.787 & 0.796 & 0.780 & 0.817 & \onebest{0.831} & \onebest{0.849} & \fivebest{0.829} & \fivebest{0.847} & 0.817 & 0.835 & 0.816 & 0.810 \\
50--100 & 0.099 & 0.519 & 0.513 & 0.471 & 0.496 & 0.499 & 0.497 & 0.511 & \fivebest{0.598} & \onebest{0.577} & 0.593 & 0.552 & 0.583 & 0.539 & \onebest{0.607} & \fivebest{0.576} & 0.588 \\
20--50 & 0.047 & 0.289 & 0.301 & 0.300 & 0.335 & 0.245 & 0.258 & 0.281 & 0.388 & \onebest{0.365} & 0.380 & 0.373 & 0.401 & 0.362 & \onebest{0.445} & \fivebest{0.417} & \fivebest{0.432} \\
10--20 & 0.012 & 0.062 & 0.102 & 0.098 & 0.121 & 0.045 & 0.060 & 0.078 & 0.136 & \onebest{0.108} & 0.119 & 0.126 & 0.138 & 0.107 & \onebest{0.159} & \fivebest{0.152} & \fivebest{0.178} \\
5--10 & 0.003 & 0.009 & 0.030 & 0.034 & 0.045 & 0.006 & 0.015 & 0.021 & 0.052 & 0.024 & 0.031 & 0.041 & 0.048 & \onebest{0.026} & \onebest{0.041} & \fivebest{0.056} & \fivebest{0.069} \\
1--5 & -- & -- & -- & 0.009 & 0.012 & -- & -- & 0.004 & 0.013 & -- & -- & 0.012 & 0.014 & -- & -- & \fivebest{0.016} & \fivebest{0.021} \\
\midrule
\multicolumn{18}{l}{\cellcolor{gray!12}\textbf{100--200\,m}} \\
$500{+}$ & 0.047 & 0.224 & 0.221 & 0.209 & 0.293 & 0.226 & 0.304 & 0.316 & 0.376 & 0.246 & \onebest{0.320} & \fivebest{0.333} & \fivebest{0.393} & \onebest{0.363} & 0.289 & 0.267 & 0.296 \\
100--500 & 0.209 & 0.586 & 0.593 & 0.592 & 0.603 & 0.648 & 0.653 & 0.667 & 0.714 & \onebest{0.680} & \onebest{0.723} & \fivebest{0.694} & \fivebest{0.736} & 0.670 & 0.684 & 0.618 & 0.596 \\
50--100 & 0.357 & 0.658 & 0.655 & 0.600 & 0.611 & 0.697 & 0.690 & \fivebest{0.687} & \fivebest{0.742} & \onebest{0.712} & \onebest{0.737} & 0.682 & 0.709 & 0.685 & 0.720 & 0.657 & 0.671 \\
20--50 & 0.406 & 0.617 & 0.667 & 0.555 & 0.582 & 0.634 & 0.670 & 0.666 & 0.713 & 0.655 & 0.694 & 0.660 & 0.663 & \onebest{0.687} & \onebest{0.745} & \fivebest{0.696} & \fivebest{0.717} \\
10--20 & 0.194 & 0.287 & 0.469 & 0.290 & 0.339 & 0.251 & 0.410 & 0.417 & 0.466 & 0.311 & 0.389 & 0.386 & 0.400 & \onebest{0.398} & \onebest{0.537} & \fivebest{0.474} & \fivebest{0.511} \\
5--10 & 0.055 & 0.042 & \onebest{0.188} & 0.098 & 0.136 & 0.030 & 0.130 & 0.141 & 0.205 & 0.066 & 0.123 & 0.140 & 0.169 & \onebest{0.086} & 0.161 & \fivebest{0.190} & \fivebest{0.234} \\
1--5 & -- & -- & -- & 0.013 & 0.023 & -- & -- & 0.016 & 0.037 & -- & -- & 0.021 & 0.031 & -- & -- & \fivebest{0.027} & \fivebest{0.041} \\
\midrule
\multicolumn{18}{l}{\cellcolor{gray!12}\textbf{200--400\,m}} \\
$500{+}$ & -- & -- & -- & -- & -- & -- & -- & -- & -- & -- & -- & -- & -- & -- & -- & -- & -- \\
100--500 & 0.092 & 0.085 & 0.102 & 0.207 & 0.192 & 0.124 & 0.143 & 0.344 & 0.333 & 0.192 & \onebest{0.286} & \fivebest{0.400} & \fivebest{0.369} & \onebest{0.195} & 0.200 & 0.285 & 0.260 \\
50--100 & 0.208 & 0.151 & 0.178 & 0.180 & 0.179 & 0.122 & 0.146 & 0.345 & 0.266 & 0.101 & 0.177 & \fivebest{0.388} & \fivebest{0.330} & \onebest{0.213} & \onebest{0.257} & 0.282 & 0.207 \\
20--50 & 0.224 & 0.204 & 0.328 & 0.204 & 0.216 & 0.240 & 0.323 & 0.320 & \fivebest{0.348} & 0.193 & 0.307 & 0.278 & 0.278 & \onebest{0.311} & \onebest{0.398} & \fivebest{0.339} & 0.320 \\
10--20 & 0.276 & 0.281 & 0.542 & 0.294 & 0.361 & 0.265 & 0.524 & \fivebest{0.493} & 0.509 & 0.310 & 0.517 & 0.375 & 0.400 & \onebest{0.405} & \onebest{0.636} & 0.469 & \fivebest{0.517} \\
5--10 & 0.211 & 0.117 & 0.448 & 0.217 & 0.286 & 0.084 & 0.373 & \fivebest{0.356} & \fivebest{0.414} & 0.143 & 0.399 & 0.247 & 0.311 & \onebest{0.184} & \onebest{0.467} & 0.340 & 0.404 \\
1--5 & -- & -- & -- & 0.017 & 0.035 & -- & -- & \fivebest{0.035} & 0.043 & -- & -- & 0.025 & 0.045 & -- & -- & 0.033 & \fivebest{0.052} \\
\midrule
\multicolumn{18}{l}{\cellcolor{gray!12}\textbf{Full range}} \\
$500{+}$ & 0.193 & 0.883 & 0.871 & 0.874 & 0.876 & 0.888 & 0.894 & 0.881 & 0.898 & \onebest{0.900} & \onebest{0.912} & \fivebest{0.903} & \fivebest{0.915} & 0.880 & 0.882 & 0.862 & 0.851 \\
100--500 & 0.224 & 0.753 & 0.747 & 0.743 & 0.751 & 0.764 & 0.773 & 0.760 & 0.795 & \onebest{0.805} & \onebest{0.830} & \fivebest{0.805} & \fivebest{0.829} & 0.791 & 0.807 & 0.774 & 0.766 \\
50--100 & 0.171 & 0.555 & 0.552 & 0.505 & 0.529 & 0.561 & 0.551 & 0.569 & \fivebest{0.641} & \onebest{0.614} & 0.634 & 0.593 & 0.622 & 0.584 & \onebest{0.642} & \fivebest{0.594} & 0.601 \\
20--50 & 0.181 & 0.429 & 0.474 & 0.398 & 0.431 & 0.424 & 0.440 & 0.463 & 0.534 & 0.482 & 0.507 & 0.496 & 0.505 & \onebest{0.512} & \onebest{0.595} & \fivebest{0.553} & \fivebest{0.564} \\
10--20 & 0.104 & 0.171 & 0.332 & 0.191 & 0.232 & 0.150 & 0.243 & 0.261 & 0.320 & 0.204 & 0.245 & 0.257 & 0.260 & \onebest{0.249} & \onebest{0.400} & \fivebest{0.343} & \fivebest{0.381} \\
5--10 & 0.060 & 0.042 & \onebest{0.193} & 0.087 & 0.119 & 0.027 & 0.109 & 0.120 & 0.185 & 0.057 & 0.103 & 0.112 & 0.125 & \onebest{0.067} & 0.157 & \fivebest{0.166} & \fivebest{0.209} \\
1--5 & -- & -- & -- & 0.012 & 0.019 & -- & -- & 0.012 & 0.028 & -- & -- & 0.017 & 0.023 & -- & -- & \fivebest{0.023} & \fivebest{0.034} \\
\bottomrule
\end{tabular}%
}
\end{minipage}

%% file: main.bbl
\begin{thebibliography}{10}
\providecommand{\url}[1]{#1}
\csname url@samestyle\endcsname
\providecommand{\newblock}{\relax}
\providecommand{\bibinfo}[2]{#2}
\providecommand{\BIBentrySTDinterwordspacing}{\spaceskip=0pt\relax}
\providecommand{\BIBentryALTinterwordstretchfactor}{4}
\providecommand{\BIBentryALTinterwordspacing}{\spaceskip=\fontdimen2\font plus
\BIBentryALTinterwordstretchfactor\fontdimen3\font minus \fontdimen4\font\relax}
\providecommand{\BIBforeignlanguage}[2]{{%
\expandafter\ifx\csname l@#1\endcsname\relax
\typeout{** WARNING: IEEEtran.bst: No hyphenation pattern has been}%
\typeout{** loaded for the language `#1'. Using the pattern for}%
\typeout{** the default language instead.}%
\else
\language=\csname l@#1\endcsname
\fi
#2}}
\providecommand{\BIBdecl}{\relax}
\BIBdecl

\bibitem{geiger2012kitti}
A.~Geiger, P.~Lenz, and R.~Urtasun, ``Are we ready for autonomous driving? {T}he {KITTI} vision benchmark suite,'' in \emph{CVPR}, 2012, pp. 3354--3361.

\bibitem{caesar2020nuscenes}
H.~Caesar \emph{et~al.}, ``nu{S}cenes: A multimodal dataset for autonomous driving,'' in \emph{CVPR}, 2020, pp. 11\,621--11\,631.

\bibitem{sun2020waymo}
P.~Sun \emph{et~al.}, ``Scalability in perception for autonomous driving: {W}aymo open dataset,'' in \emph{CVPR}, 2020, pp. 2446--2454.

\bibitem{wilson2023argoverse2}
B.~Wilson \emph{et~al.}, ``Argoverse 2: Next generation datasets for self-driving perception and forecasting,'' in \emph{NeurIPS}, 2023.

\bibitem{jung2024helipr}
M.~Jung \emph{et~al.}, ``{HeLiPR}: Heterogeneous {LiDAR} dataset for inter-{LiDAR} place recognition under spatiotemporal variations,'' \emph{Int. J. Robot. Res.}, vol.~43, no.~12, pp. 1867--1883, 2024.

\bibitem{lee2024helimos}
H.~Lee \emph{et~al.}, ``{HeLiMOS}: A dataset for moving object segmentation in {3D} point clouds from heterogeneous {LiDAR} sensors,'' \emph{arXiv preprint arXiv:2408.06328}, 2024.

\bibitem{ghilotti2026truckdrive}
F.~Ghilotti \emph{et~al.}, ``Truckdrive: Long-range autonomous highway driving dataset,'' in \emph{CVPR}, 2026, pp. 10\,587--10\,598.

\bibitem{yin2021center}
T.~Yin, X.~Zhou, and P.~Kr{\"a}henb{\"u}hl, ``Center-based 3d object detection and tracking,'' in \emph{CVPR}, 2021, pp. 11\,784--11\,793.

\bibitem{wang2023dsvt}
H.~Wang \emph{et~al.}, ``{DSVT}: Dynamic sparse voxel transformer with rotated sets,'' in \emph{CVPR}, 2023, pp. 13\,520--13\,529.

\bibitem{chen2023voxelnext}
Y.~Chen \emph{et~al.}, ``{VoxelNeXt}: Fully sparse {VoxelNet} for {3D} object detection and tracking,'' in \emph{CVPR}, 2023, pp. 21\,674--21\,683.

\bibitem{bai2022transfusion}
X.~Bai \emph{et~al.}, ``{TransFusion}: Robust {LiDAR}-camera fusion for {3D} object detection with transformers,'' in \emph{CVPR}, 2022, pp. 1090--1099.

\bibitem{zhang2024deflow}
Q.~Zhang \emph{et~al.}, ``{DeFlow}: Decoder of scene flow network in autonomous driving,'' in \emph{ICRA}, 2024, pp. 2105--2111.

\bibitem{kim2025flow4d}
J.~Kim \emph{et~al.}, ``{Flow4D}: Leveraging {4D} voxel network for {LiDAR} scene flow estimation,'' \emph{IEEE Robot. Autom. Lett.}, vol.~10, no.~4, pp. 3462--3469, 2025.

\bibitem{khoche2025ssf}
A.~Khoche \emph{et~al.}, ``{SSF}: Sparse long-range scene flow for autonomous driving,'' in \emph{ICRA}, 2025, pp. 6394--6400.

\bibitem{zhu2021cylinder3d}
X.~Zhu \emph{et~al.}, ``Cylindrical and asymmetrical {3D} convolution networks for {LiDAR} segmentation,'' in \emph{CVPR}, 2021.

\bibitem{wu2024ptv3}
X.~Wu \emph{et~al.}, ``Point transformer {V3}: Simpler, faster, stronger,'' in \emph{CVPR}, 2024.

\bibitem{mao2021once}
J.~Mao \emph{et~al.}, ``One million scenes for autonomous driving: {ONCE} dataset,'' in \emph{NeurIPS}, 2021.

\bibitem{meyer2019astyx}
M.~Meyer and G.~Kuschk, ``Automotive radar dataset for deep learning based {3D} object detection,'' in \emph{EuRAD}, 2019, pp. 129--132.

\bibitem{palffy2022vod}
A.~Palffy \emph{et~al.}, ``Multi-class road user detection with {3+1D} radar in the {V}iew-of-{D}elft dataset,'' \emph{IEEE Robot. Autom. Lett.}, vol.~7, no.~2, pp. 4961--4968, 2022.

\bibitem{zheng2022tj4dradset}
L.~Zheng \emph{et~al.}, ``{TJ4DRadSet}: A {4D} radar dataset for autonomous driving,'' in \emph{ITSC}, 2022, pp. 493--498.

\bibitem{paek2022kradar}
D.-H. Paek, S.-H. Kong, and K.~T. Wijaya, ``{K-Radar}: {4D} radar object detection for autonomous driving in various weather conditions,'' in \emph{NeurIPS}, 2022.

\bibitem{man2024truckscenes}
F.~Fent \emph{et~al.}, ``{MAN} {T}ruck{S}cenes: A multimodal dataset for autonomous trucking in diverse conditions,'' in \emph{NeurIPS}, 2024.

\bibitem{gu2022fmcw_tracking}
Y.~Gu \emph{et~al.}, ``Learning moving-object tracking with {FMCW} {LiDAR},'' in \emph{IROS}, 2022, pp. 3747--3753.

\bibitem{dopplerptnet2024}
G.~Xi \emph{et~al.}, ``{DopplerPTNet}: Object detection network with {D}oppler velocity information for {FMCW} {LiDAR} point cloud,'' \emph{J. Phys.: Conf. Ser.}, vol. 2809, p. 012006, 2024.

\bibitem{shi2025pod}
Y.~Shi \emph{et~al.}, ``{POD}: Predictive object detection with single-frame {FMCW} {LiDAR} point cloud,'' \emph{arXiv preprint arXiv:2504.05649}, 2025.

\bibitem{sakaridis2024muses}
T.~Br{\"o}dermann \emph{et~al.}, ``{MUSES}: The multi-sensor semantic perception dataset for driving under uncertainty,'' in \emph{ECCV}, 2024.

\bibitem{aevascenes2025}
G.~N. Narasimhan \emph{et~al.}, ``Aevascenes: A dataset and benchmark for fmcw lidar perception,'' 2025.

\bibitem{qian2026_4dlidaropen}
K.~Qian \emph{et~al.}, ``{4DLidarOpen}: An open {4D FMCW LiDAR} dataset for motion-aware autonomous driving,'' \emph{arXiv preprint arXiv:2605.18074}, 2026.

\bibitem{hexsel2022dicp}
B.~Hexsel, H.~Vhavle, and Y.~Chen, ``{DICP}: {D}oppler iterative closest point algorithm,'' in \emph{Robotics: Science and Systems ({RSS})}, 2022.

\bibitem{wu2022pickingupspeed}
Y.~Wu \emph{et~al.}, ``Picking up speed: Continuous-time lidar-only odometry using doppler velocity measurements,'' \emph{IEEE Robot. Autom. Lett.}, vol.~8, no.~1, pp. 264--271, 2022.

\bibitem{zhao2024fmcwlio}
M.~Zhao \emph{et~al.}, ``{FMCW-LIO}: A doppler {LiDAR}-inertial odometry,'' \emph{IEEE Robot. Autom. Lett.}, 2024.

\bibitem{cao2025caoronet}
Z.~Li \emph{et~al.}, ``{CAO-RONet}: A robust 4{D} radar odometry with exploring more information from low-quality points,'' \emph{arXiv preprint arXiv:2503.01438}, 2025.

\bibitem{wang2025dynamicicp}
D.~Wang \emph{et~al.}, ``{Dynamic-ICP}: Doppler-aware iterative closest point registration for dynamic scenes,'' \emph{arXiv preprint arXiv:2511.20292}, 2025.

\bibitem{clip}
A.~Radford \emph{et~al.}, ``Learning transferable visual models from natural language supervision,'' in \emph{Proc. ICML}, 2021.

\bibitem{siglip2}
M.~Tschannen \emph{et~al.}, ``{SigLIP 2}: Multilingual vision-language encoders with improved semantic understanding, localization, and dense features,'' \emph{arXiv:2502.14786}, 2025.

\bibitem{ester1996dbscan}
M.~Ester \emph{et~al.}, ``A density-based algorithm for discovering clusters in large spatial databases with noise,'' in \emph{KDD}, 1996, pp. 226--231.

\bibitem{julier1997ukf}
S.~J. Julier and J.~K. Uhlmann, ``New extension of the {Kalman} filter to nonlinear systems,'' in \emph{Signal Processing, Sensor Fusion, and Target Recognition VI}, vol. 3068.\hskip 1em plus 0.5em minus 0.4em\relax SPIE, 1997, pp. 182--193.

\bibitem{chen2021lidarmos}
X.~Chen \emph{et~al.}, ``Moving object segmentation in {3D} {LiDAR} data: A learning-based approach exploiting sequential data,'' \emph{IEEE Robot. Autom. Lett.}, vol.~6, no.~4, pp. 6529--6536, 2021.

\end{thebibliography}
